\documentclass[11pt]{article}

\usepackage[preprint]{acl}

\usepackage{times}
\usepackage{latexsym}

\usepackage[T1]{fontenc}

\usepackage[utf8]{inputenc}

\usepackage{microtype}

\usepackage{inconsolata}

\usepackage{graphicx}

\usepackage[utf8]{inputenc} %
\usepackage[T1]{fontenc}    %
\usepackage{hyperref}       %
\usepackage{url}            %
\usepackage{booktabs}       %
\usepackage{amsfonts}       %
\usepackage{nicefrac}       %
\usepackage{microtype}      %
\usepackage{xcolor}         %

\usepackage{amsmath}
\usepackage{amssymb}
\usepackage[ruled,vlined]{algorithm2e}
\usepackage{adjustbox}
\usepackage{caption}
\usepackage{comment}
\usepackage{enumitem}
\usepackage{float}
\usepackage{graphicx}
\usepackage{longtable}
\usepackage{linguex}
\usepackage{listings}
\usepackage{multirow}
\usepackage{soul}
\usepackage{subcaption}
\usepackage{relsize}
\usepackage{tabularray}
\usepackage{tikz}
\usepackage{todonotes}
\usepackage{tcolorbox}
\usepackage{colortbl}
\usepackage{pgf}
\usepgflibrary{fpu}  %

\definecolor{steercolor}{RGB}{255, 120, 100}
\definecolor{deactcolor}{RGB}{80, 120, 255} 

\newtcolorbox{promptbox}{
  colback=gray!5,
  colframe=black,
  boxrule=0.5pt,
  arc=2pt
}

\title{Multilingual Emotion Neurons in Large Audio-Language Models}

\author{
 \textbf{Xiutian Zhao\textsuperscript{1}},
 \textbf{Philipp Koehn\textsuperscript{1}},
 \textbf{Bj\"orn Schuller\textsuperscript{2}}, 
 \textbf{Berrak Sisman\textsuperscript{1}}
\\
 \textsuperscript{1} Center for Language and Speech Processing (CLSP), Johns Hopkins University, USA \\
 \textsuperscript{2} Group on Language, Audio \& Music (GLAM), Imperial College London, UK
}

\begin{document}
\maketitle
\begin{abstract}

Emotion is central to human communication, and its expression varies across languages. Large audio-language models (LALMs) achieve strong performance on multilingual speech tasks, yet it remains unclear whether they encode emotion through language-specific correlations or language-agnostic representations. We present the first neuron-level interpretability study of this question. We define Multilingual Emotion Neurons (MLENs) as functional units exhibiting stable emotional selectivity and aligned causal effects across languages, and introduce Consistency-Regularized Fusion (CR-Fusion) to identify them. Across four modern LALMs and 12 typologically diverse languages, emotion-sensitive neurons identified independently per language show minimal overlap, and additional monolingual identification data saturates quickly without isolating more transferable units, motivating identification from pooled cross-lingual evidence. Causal interventions demonstrate that MLENs identified by CR-Fusion provide more precise and transferable affective control than monolingual neuron sets in both zero-shot and low-resource settings. Leave-one-out ablations further reveal asymmetric transfer: individual identification languages, including low-resource ones, contribute non-redundant evidence, while several low-resource languages benefit most from the resulting cross-lingual transfer. Together, our findings provide the first causal, neuron-level account of how LALMs encode emotion across languages, and establish multilingual neuron identification as an effective mechanism for understanding cross-lingual affective behavior.

\end{abstract}

\section{Introduction}
Emotion is a core channel of human communication: beyond what we say, speech conveys how we feel through prosody, timing, intensity, and culturally shaped display rules \cite{wilce2009language, lindquist2015role}. In a multilingual world, this affective layer must be robust to linguistic diversity: people routinely infer emotion from unfamiliar languages, yet recognition is modulated by language, culture, and learned acoustic conventions \cite{russell1991culture, pell2009recognizing, doi:10.1126/science.aaw8160}. This tension between pan-cultural cues and language-specific realizations makes multilingual emotion an ideal testbed for probing the generalizability of concept representations in modern foundation models.

Decades of cross-lingual SER research show that emotion cues partially generalize to unseen languages, yet transfer remains sensitive to language and corpus mismatch, and human and model transfer patterns diverge systematically~\citep{7337399, 10889008}. It thus remains open to what extent current models rely on cross-lingually shared affective representations versus dataset-, corpus-, and language-local correlations.

Parallel interpretability work shows that individual neurons in LLMs and multimodal models align with human-interpretable concepts, including affective states \citep{huben2024sparse,bau2017network,sofroniew2026emotionconceptsfunctionlarge}. However, emotion-correlated neurons identified in one language often fail to generalize to others in multilingual encoders \citep{singh-etal-2026-lost}, and cross-lingual generalization of activation steering remains contested \citep{maraia-etal-2026-activation}. This tension between language-specific and language-agnostic emotion representations motivates our investigation (\S\ref{sec:related}).

Modern multimodal foundation models \cite{yao2024minicpmvgpt4vlevelmllm, xu2025qwen25omnitechnicalreport}, particularly LALMs \cite{kimiteam2025kimiaudiotechnicalreport, goel2025audioflamingo3advancing}, offer a new paradigm for probing this question. By integrating vast acoustic and linguistic pre-training, LALMs have achieved competitive performance on 
a broad range of spoken language understanding
tasks~\citep{10448257,sakshi2025mmau}, including speech emotion
recognition~\citep{cheng2024emotionllama,he2025meralionaudiollmbridgingaudiolanguage}. However, mechanistic accounts of how these models represent affective information remain lacking. It is unclear whether LALMs utilize language-specific functional units or possess a shared representational structure that generalizes emotion processing across broad linguistic diversity.

To the best of our knowledge, this work represents the first systematic attempt to investigate and manipulate the intrinsic multilingual emotion representations within LALMs.
Using SER as a probe task, we identify affective neurons through activations across four open-source LALMs, Audio-Flamingo-3 \cite{goel2025audioflamingo3advancing}, Qwen2.5-Omni-7B \cite{xu2025qwen25omnitechnicalreport}, MiniCPM-o-4.5 \cite{yao2024minicpmvgpt4vlevelmllm}, and Kimi-Audio \cite{kimiteam2025kimiaudiotechnicalreport}, and 12 typologically diverse languages, including high-resource languages (e.g., English, Mandarin) and low-resource languages such as Amharic, Bengali and Urdu, where emotional speech data is often sparse \cite{koehn-knowles-2017-six, 10711189}. To isolate neurons with stable cross-lingual emotional selectivity, we propose \emph{Consistency-Regularized Fusion} (CR-Fusion) that operates atop any neuron selector (e.g., Mean-activation-based \cite{bau2019identifying, 10.1609/aaai.v33i01.33016309}) to leverage multilingual evidence for identifying cross-lingually stable affective units.

Our investigation yields five principal findings. 
(1) \textbf{Emotion-sensitive neurons (ESNs) identified independently in different languages exhibit minimal set intersection} with weak to moderate rank correlation in selectivity scores, revealing that monolingual identification procedures select substantially different neurons even when underlying ranking structures are similar. 
(2) \textbf{Monolingual evidence saturates rapidly} beyond 50 instances; additional within-language data yields no further improvement in isolating transferable neurons. 
(3) We define Multilingual Emotion Neurons (MLENs) as functional units exhibiting stable emotional selectivity across languages. Causal interventions confirm that \textbf{MLENs play a functional role in emotion processing}: deactivation selectively impairs recognition of target emotions, while steering selectively enhances it. Our proposed CR-Fusion \textbf{matches or outperforms the best monolingual identification} in zero-shot and low-resource settings across models, with the exception of steering on Qwen2.5-Omni-7B, the model with the lowest language-invariant share.
(4) Causal effects are heterogeneous across emotions: anger, happiness and sadness show the largest and most stable intervention effects, while fear and neutral are weaker and more variable. This ordering tracks baseline recognition accuracy and class support rather than representational universality---per-emotion invariant shares are in fact highest for neutral and fear. 
(5) We uncover \textbf{asymmetric transfer patterns}, with individual identification languages contributing non-substitutable evidence.
These results demonstrate that \textbf{low-resource languages can contribute non-redundant evidence for multilingual neuron identification while simultaneously benefiting from improved cross-lingual transfer.}

These findings provide a mechanistic framework for understanding how multimodal foundation models process emotion across languages, with practical implications for training-free cross-lingual generalization and equitable affective computing for low-resource language communities.

\section{Related Work}
\label{sec:related}
\paragraph{Multilingual Speech Emotion Recognition.}
Research on multilingual SER has progressed from early cross-lingual transfer using traditional deep neural networks \cite{7337399, 8462162, zehra2021cross} to more scalable approaches leveraging self-supervised encoders such as wav2vec 2.0 and transformer-based architectures \cite{9747417, app12189188}. Recent benchmarks like EmoBox \cite{ma24b_interspeech} have standardized multilingual evaluation while highlighting persistent cross-lingual robustness challenges, with studies revealing systematic divergences between human and model transfer patterns \cite{singh2023decodingemotionscomprehensivemultilingual, 10889008}. Current work explores LALMs for zero-shot cross-lingual recognition, emotion captioning, and joint audio-text reasoning \cite{11209040, Xu_Chen_Yu_Huang_Wu_Zhang_Li_Luo_Gu_2024, he2025meralionaudiollmbridgingaudiolanguage}, complemented by parameter-efficient methods like LoRA for cross-lingual alignment \cite{goncalves24_interspeech}. However, there is no mechanistic study of emotion representation across languages in LALMs \cite{shou2025multimodallargelanguagemodels}.

\paragraph{Neuron-Level Concept and Emotion Interpretability in Audio-Language Models.}
Identifying functional units that respond selectively to human-interpretable concepts is a long-standing theme in interpretability, established in vision \cite{bau2017network,bau2020understanding} and LLMs \cite{huben2024sparse,voita-etal-2024-neurons,yu-ananiadou-2024-neuron,tang-etal-2024-language}. For affect specifically, recent work identifies clustered emotion neurons in LLMs \cite{lee-etal-2025-large} that are causally controllable via activation steering \cite{wang2025llmsfeelemotioncircuits}, with \citet{sofroniew2026emotionconceptsfunctionlarge} showing how affective states decompose into
modular circuits; yet \citet{singh-etal-2026-lost} find that emotion-correlated neurons identified in one language fail to generalize in multilingual encoders, and cross-lingual generalization of steering remains contested \cite{maraia-etal-2026-activation}. In LALMs, neuron-level analysis has so far targeted modality attribution and generic acoustic concepts \cite{huo-etal-2024-mmneuron,wu2024andaudionetworkdissection}, or probed phonetic and prosodic encoding layer-wise \cite{yang2025audiolenscloserlookauditory}; affective units have been identified only monolingually, for SER \cite{zhao-etal-2026-discovering} and emotional voice conversion \cite{zhao2026neuronlevelemotioncontrolspeechgenerative}. Their stability across linguistic boundaries has not been explored, which motivates our investigation (extended discussion in Appendix~\ref{app:related}).

\section{Method}
\label{sec:method}
Our pipeline comprises three stages: (1) activation logging on correctly recognized SER instances, (2) neuron scoring and selection using principled fusion strategies that integrate language-conditioned statistics into cross-lingual neuron sets, and (3) causal analysis through targeted intervention.

\subsection{Multilingual Activation Logging}
\label{subsec:logging}
Let $\mathcal{L}$ denote a set of languages and $\mathcal{E}$ the set of emotions. 
We run the unintervened model on an SER task of emotion-labeled speech and restrict logging to correctly predicted items to reduce contamination from failure modes and to obtain cleaner emotion-conditioned statistics. Let $\mathcal{D}^{(\ell,e)}$ denote the set of correctly predicted utterances of emotion $e \in \mathcal{E}$ in language $\ell \in \mathcal{L}$. For an utterance $x$ with token positions $t \in \{1,\dots,T_x\}$, let $a_{l,n,t}(x)$ denote the scalar SwiGLU gate activation \cite{shazeer2020gluvariantsimprovetransformer} of neuron $n$ in layer $l$, and let $w_{x,t} \in \{0,1\}$ mask out irrelevant positions (e.g., padding and instruction-prompt tokens). We aggregate positive-activation counts and valid token counts as
\begin{equation}
\begin{aligned}
K^{(\ell,e)}_{l,n} = \!\!\sum_{x \in \mathcal{D}^{(\ell,e)}} \sum_{t=1}^{T_x}
  w_{x,t}\, \mathbb{I}\!\left[a_{l,n,t}(x) > 0\right],
\\
T^{(\ell,e)} = \!\!\sum_{x \in \mathcal{D}^{(\ell,e)}} \sum_{t=1}^{T_x} w_{x,t},
\end{aligned}
\end{equation}
and form the activation-probability profile
$P^{(\ell,e)}_{l,n} = K^{(\ell,e)}_{l,n} \big/ T^{(\ell,e)}$.

\subsection{Monolingual Evidence and Neuron Identification Methods}
\label{subsec:selector}

We first obtain monolingual evidence by scoring neurons independently per language. Let $s^{(\ell,e)}_{l,n}$ denote a selector score derived from $P^{(\ell,e)}_{l,n}$. We use a margin-based selector, \textbf{Contrastive Activation Margin (ConAct)} \cite{zhao-etal-2026-finding}, as our default due to its consistently stronger intervention effectiveness in comparison with LAP \cite{huben2024sparse, gurnee2024universal}, LAPE \cite{tang-etal-2024-language, namazifard-poech-2025-isolating}, and MAD \cite{bau2019identifying, 10.1609/aaai.v33i01.33016309}.
ConAct assigns each neuron to its most preferred emotion and measures a margin against the second most preferred. Using $P^{(\ell,e)}_{l,n}$, we define: $P^{(1)}_{l,n}(\ell) = \max_{e \in \mathcal{E}} P^{(\ell,e)}_{l,n}$
\begin{equation*}
e^{(1)}_{l,n}(\ell) = \arg\max_{e} P^{(\ell,e)}_{l,n},
P^{(2)}_{l,n}(\ell) = \max_{e \neq e^{(1)}_{l,n}(\ell)} P^{(\ell,e)}_{l,n}
\end{equation*}
\begin{equation*}
s^{(\ell,e)}_{l,n} =
\begin{cases}
P^{(1)}_{l,n}(\ell) - P^{(2)}_{l,n}(\ell), & e = e^{(1)}_{l,n}(\ell) \\
0, & \text{otherwise}
\end{cases}.
\end{equation*}

For each $(\ell,e)$, we rank neurons by the emotion-conditioned score $s^{(\ell,e)}_{l,n}$ and select a fixed fraction $r{=}0.5\%$ to obtain monolingual ESN sets $\mathcal{I}_l^{(\ell,e)}$.

\subsection{Fusion Strategies for Multilingual Emotion Neurons}
\label{sec:fusion}
We now construct MLENs by fusing monolingual evidence $\{s^{(\ell,e)}_{l,n}\}_{\ell\in\mathcal{L}}$. The goal is to identify neurons that are not only emotion-selective, but also consistent across languages. Our designs are inspired by multi-view learning and ensembling principles (e.g., intersection/consensus, averaging, and variance-regularized objectives), adapted here to neuron selection.

\paragraph{Overlap Fusion.}
This conservative strategy retains neurons supported by \emph{all} languages. We form a fixed-budget MLEN set by re-ranking candidates with a joint score that prefers agreement (i.e. by its
weakest per-language evidence), $s^{\mathrm{overlap},(e)}_{l,n} \;=\; \min_{\ell \in L}\, \tilde{s}^{(\ell,e)}_{l,n},$
where $\tilde s$ denotes per-language normalized scores (z-score or robust z-score across neurons). We then select the top-$r$ neurons by $s^{\text{overlap}}$ subject to a same-emotion constraint (the preferred emotion agrees across languages).

\paragraph{Joint Fusion.}
This strategy treats all languages as a single unified dataset by pooling raw activation statistics before computing selector scores. We first merge the activation counts and token counts across all languages:
$K^{(\text{joint},e)}_{l,n} = \sum_{\ell\in\mathcal{L}} K^{(\ell,e)}_{l,n}, \quad T^{(\text{joint},e)} = \sum_{\ell\in\mathcal{L}} T^{(\ell,e)}.$
Then, we compute the pooled activation probability:
$P^{(\text{joint},e)}_{l,n} = \frac{K^{(\text{joint},e)}_{l,n}}{T^{(\text{joint},e)}}$.
We apply the selector directly to these pooled probabilities to obtain joint scores $s^{(\text{joint},e)}_{l,n}$, and select the top-$r$ neurons. This approach is data-efficient and treats cross-lingual evidence as natural augmentation, but may be dominated by high-resource languages if case counts are imbalanced across $\mathcal{L}$.

\paragraph{Consistency-Regularized Fusion (CR-Fusion).}
To explicitly prefer cross-lingual stability, we penalize dispersion of scores across languages. 
$\tilde{s}^{(\ell,e)}_{l,n} =
\bigl(s^{(\ell,e)}_{l,n}-\mu^{(\ell)}\bigr)/\bigl(\sigma^{(\ell)}+\epsilon_0\bigr)$,
where $\mu^{(\ell)}$ and $\sigma^{(\ell)}$ are the mean and standard deviation computed over all (layer, neuron, emotion) triplets.
\begin{equation*}
\begin{aligned}
    \mu^{\mathrm{CR}, (e)}_{l,n} &= \frac{1}{|\mathcal{L}|}\sum_{\ell\in\mathcal{L}} \tilde{s}^{(\ell,e)}_{l,n}, \\
    \sigma^{\mathrm{CR}, (e)}_{l,n} &= \sqrt{\frac{1}{|\mathcal{L}|}\sum_{\ell\in\mathcal{L}}\big(\tilde{s}^{(\ell,e)}_{l,n} - \mu^{\mathrm{CR}, (e)}_{l,n}\big)^2}.
\end{aligned}
\end{equation*}

We then score neurons by:
$s^{\mathrm{CR},(e)}_{l,n} = \mu^{\mathrm{CR},(e)}_{l,n} - \lambda \cdot \sigma^{\mathrm{CR},(e)}_{l,n},$
where $\lambda {\ge} 0$ acts as a regularization parameter that penalizes cross-lingual variance. Higher values of $\lambda$ prioritize neurons with high stability over those with high absolute monolingual scores, thereby filtering out language-specific units. Sensitivity to $\lambda$ is discussed in \S~\ref{sec:lambda}.

\subsection{The Estimand of Cross-Lingual Fusion}
\label{sec:theory}
 
\paragraph{Fusion Strategies as a Robustness Family.}
Write the per-language selector score for a fixed neuron as $s^{(\ell,e)} = \theta(e) + \delta^{(\ell,e)} + \xi^{(\ell,e)}$, 
where $\theta^{(e)}$ is a language-invariant emotion effect, $\delta^{(\ell,e)}$ is genuine language-specific deviation with $\mathbb{E}_{\ell}[\delta^{(\ell,e)}]=0$ and variance $\tau^2$, and $\xi^{(\ell,e)}$ is finite-sample noise. Under a Gaussian approximation, the CR-Fusion objective $\mu^{\mathrm{CR}} - \lambda\,\sigma^{\mathrm{CR}}$ is the $(1{-}q)$ lower confidence bound of the neuron's selectivity in a randomly drawn language, with $\lambda = \Phi^{-1}(1-q)$. The three fusion strategies of \S\ref{sec:fusion} are therefore one family indexed by robustness: mean fusion ($\lambda{=}0$), of which Joint Fusion is the token-weighted counterpart, targets expected transfer to a new language, CR-Fusion targets quantile transfer, and Overlap Fusion  targets worst-case transfer.
 
\paragraph{The Predicted Optimal Penalty.}
Our evaluation averages ESS over held-out languages, which constitutes an expected-transfer objective. If languages are exchangeable, the Bayes-optimal selection criterion for this objective is $\theta$ itself, whose minimum-variance unbiased estimator is the cross-language mean; the predicted optimum is therefore $\lambda$ near zero. Moreover, the observed dispersion estimates $\tau^2 + \bar v$, where $\bar v$ is within-language sampling variance: since $\hat P$ is a binomial proportion, $v \approx P(1-P)/T^{(\ell,e)}$, and $T$ varies by more than an order of magnitude across our conditions (Appendix~\ref{app:id_budget}). A large penalty potentially removes neurons whose statistics are estimated from sparse data rather than neurons that are meaningfully language-specific. Moreover, because selection at a fixed top-$r$ fraction operates in the upper tail, where score magnitude and dispersion are positively correlated, a large penalty drives the selected set toward uniformly weak units. Both predictions are confirmed in \S\ref{sec:lambda}.
 
\paragraph{The Recovered Estimand.}
Decompose the activation-probability tensor additively over languages and emotions,
\begin{equation}
P^{(\ell,e)}_{l,n} \;=\; m_{l,n} + u^{(\ell)}_{l,n} + b^{(e)}_{l,n}
+ \gamma^{(\ell,e)}_{l,n} + \varepsilon^{(\ell,e)}_{l,n},
\label{eq:decomp}
\end{equation}
where $m$ is the neuron's baseline activity, $u^{(\ell)}$ a language main effect (overall firing shift), $b^{(e)}$ the language-invariant emotion effect, and $\gamma^{(\ell,e)}$ the language$\times$emotion interaction.
For a complete design with uniform weights, the least-squares fit satisfies $\hat{m}+\hat{b}^{(e)} = \frac{1}{|\mathcal{L}|}\sum_{\ell} P^{(\ell,e)}$, the unweighted cross-language mean profile. Joint Fusion scores the token-weighted pooled profile $\sum_{\ell} K^{(\ell,e)} / \sum_{\ell} T^{(\ell,e)}$,
and small-$\lambda$ CR-Fusion approximates it.

Empirically, applying ConAct to $m + \hat b$ selects neuron sets with mean Jaccard similarity $0.98$ against Joint Fusion on the two models with complete language--emotion coverage, dropping to $0.41$--$0.63$ on the two models with missing cells, where the weighted fit and naive pooling legitimately diverge (Appendix~\ref{app:decomp}). Cross-lingual fusion thus succeeds not by consensus filtering but by estimating a specific, identifiable estimand: the invariant emotion effect, which no single-corpus procedure can isolate.
 
\paragraph{Magnitude of the Invariant Share.} Noise-corrected variance components of Eq.~(\ref{eq:decomp}) yield invariant shares between $0.30$ (Qwen2.5-Omni-7B) and $0.59$ (MiniCPM-o-4.5) among top-ranked emotion-selective neurons (Appendix~\ref{app:decomp}). In no model does the invariant share exceed $0.6$: much of the emotion-conditioned activation structure, in two models the majority, is specific to individual languages or corpora (see Limitations).

\subsection{Intervention}
\label{subsec:intervention}
Given selected neuron indices $\mathcal{I}_l^{(e)}$ for emotion $e$
at layer $l$ (monolingual or fused), we evaluate causal effects via two interventions: deactivation and steering. Our interventions operate on the post-activation SwiGLU gate outputs in decoder MLP modules. Let $g_{l,t} \in \mathbb{R}^{D_l}$ denote the activated gate vector at layer $l$ and token position $t$, computed as $g_{l,t} = \text{act}(\text{gate\_proj}(x_{l,t}))$, where $\text{act}(\cdot)$ is the activation function. 

\paragraph{Deactivation.}
We zero selected neurons' activations through element-wise masking:
\begin{equation*}
de^{(e)}_{l,n} =
\begin{cases}
0, & n \in \mathcal{I}^{(e)}_l \\
1, & \text{otherwise}
\end{cases},
\qquad
\tilde{g}^{\mathrm{deact}}_{l,t} {=} g_{l,t} \odot de^{(e)}_l.
\end{equation*}

\paragraph{Steering.}
We amplify selected neurons by applying a multiplicative gain factor $\alpha \geq 0$:
\begin{equation*}
st^{(e)}_{l,n}(\alpha) =
\begin{cases}
1 {+} \alpha,  n \in \mathcal{I}^{(e)}_l \\
1,  \text{otherwise}
\end{cases},
\tilde{g}^{\mathrm{steer}}_{l,t} {=} g_{l,t} \odot st^{(e)}_l(\alpha).
\end{equation*}

The modified gate activations $\tilde{g}_{l,t}$ are then used in place of $g_{l,t}$.
We apply these interventions systematically across evaluated languages to assess whether MLENs induce aligned causal changes, thereby validating their cross-lingual generalizability and functional specificity.

\section{Experiment Setup}

\paragraph{Datasets and Partition.}
Our test bed comprises 12 monolingual emotional speech datasets categorized by their role in the transferability pipeline. The \textbf{Identification Set ($\mathcal{L}_\text{id}$)} consists of languages used during activation logging to ensure cross-lingual consistency: Amharic (ASED \cite{10.1145/3529759}), Moroccan Arabic (MDER \cite{ev21-c430-24}), Bengali (SUBESCO \cite{SUBESCO}), English (MSP-Podcast \cite{11457331}), Italian (Emozionalmente \cite{Emozionalmente}), Mandarin (BIIC \cite{BIIC}), Polish (nEMO \cite{christop-2024-nemo}), and Urdu (UrduSER \cite{UrduSER}). The \textbf{Held-out Set ($\mathcal{L}_\text{held}$)} contains languages reserved strictly for zero-shot evaluation: French (CaFE \cite{10.1145/3204949.3208121}), German (EmoDB \cite{EmoDB}), Persian (ShEMO \cite{mohamad2019shemo}) and Russian (RESD \cite{aniemore_2023}). Notably, Amharic, Bengali, Urdu, and Moroccan Arabic are treated as low-resource languages within this framework \cite{koehn-knowles-2017-six}. The dataset statistics are provided in Appendix~\ref{appendix:reproducibility} Table~\ref{tab:datasets}. Evaluation partitions sample up to 150 utterances per emotion (Appendix~\ref{app:data}).

\paragraph{Models, Prompting and Decoding.}
\label{sec:decoding}
We evaluate four open-sourced LALMs: Audio-Flamingo-3 \cite{goel2025audioflamingo3advancing}, Kimi-audio \cite{kimiteam2025kimiaudiotechnicalreport}, MiniCPM-o-4.5 \cite{yao2024minicpmvgpt4vlevelmllm}, and Qwen2.5-Omni-7B \cite{xu2025qwen25omnitechnicalreport}. All selected models designedly support English and Mandarin, with varying, officially unclaimed capabilities in other languages.
All models are evaluated in a controlled multiple-choice format using a single instruction template (Appendix~\ref{appendix:prompt}). 
All inference uses deterministic decoding (greedy, temperature=0); reported
variance therefore reflects cross-language variability
(Appendix~\ref{appendix:stats}).

\paragraph{Metrics.}
We assess baseline performances using unweighted average recall (UAR) across five emotions: anger, fear, happiness, neutral, and sadness. To quantify causal effects, we define \textbf{Self-Effects (SE)} as the accuracy change on a specific emotion subset when applying the matched neuron mask, relative to the unintervened baseline: $\text{SE}(e) = \mathrm{Acc}_{e \to e} - \mathrm{Acc}_{\text{unintervened}} (e)$. 
\textbf{Average Cross-Effects (ACE)} measures, for each target emotion $e$, the mean accuracy change on $e$ under masks matched to other emotions: $\mathrm{ACE}(e) = \frac{1}{|\mathcal{E}|-1}\sum_{e' \neq e} \left(\mathrm{Acc}_{e' \to e} - \mathrm{Acc}_{\text{unintervened}} (e)\right)$.
\textbf{Emotion Selectivity Score (ESS)} quantifies specificity by comparing self-effect against cross-effect. The per-emotion ESS is defined as:
$\text{ESS}(e) = \text{SE}(e) - \text{ACE}(e).$
The global ESS, which aggregates across all valid emotions, is computed as:
$\text{ESS} = \frac{1}{|\mathcal{E}_{\text{valid}}|} \sum_{e \in \mathcal{E}_{\text{valid}}} \text{ESS}(e),$
where $\mathcal{E}_{\text{valid}}$ denotes emotions with at least one test instance. Unless otherwise stated, all reported ESS values refer to this global metric.
Negative ESS under deactivation indicates selective impairment of target emotions while preserving recognition of other emotions; positive ESS under steering indicates selective enhancement of target emotions.

\section{Results}
\label{sec:results}

\subsection{Cross-Lingual Agreement of Monolingual Emotion-Sensitive Neurons}
\begin{figure}[ht!]
    \centering
    \begin{subfigure}[b]{0.5\columnwidth}
        \centering
        \includegraphics[width=\linewidth]{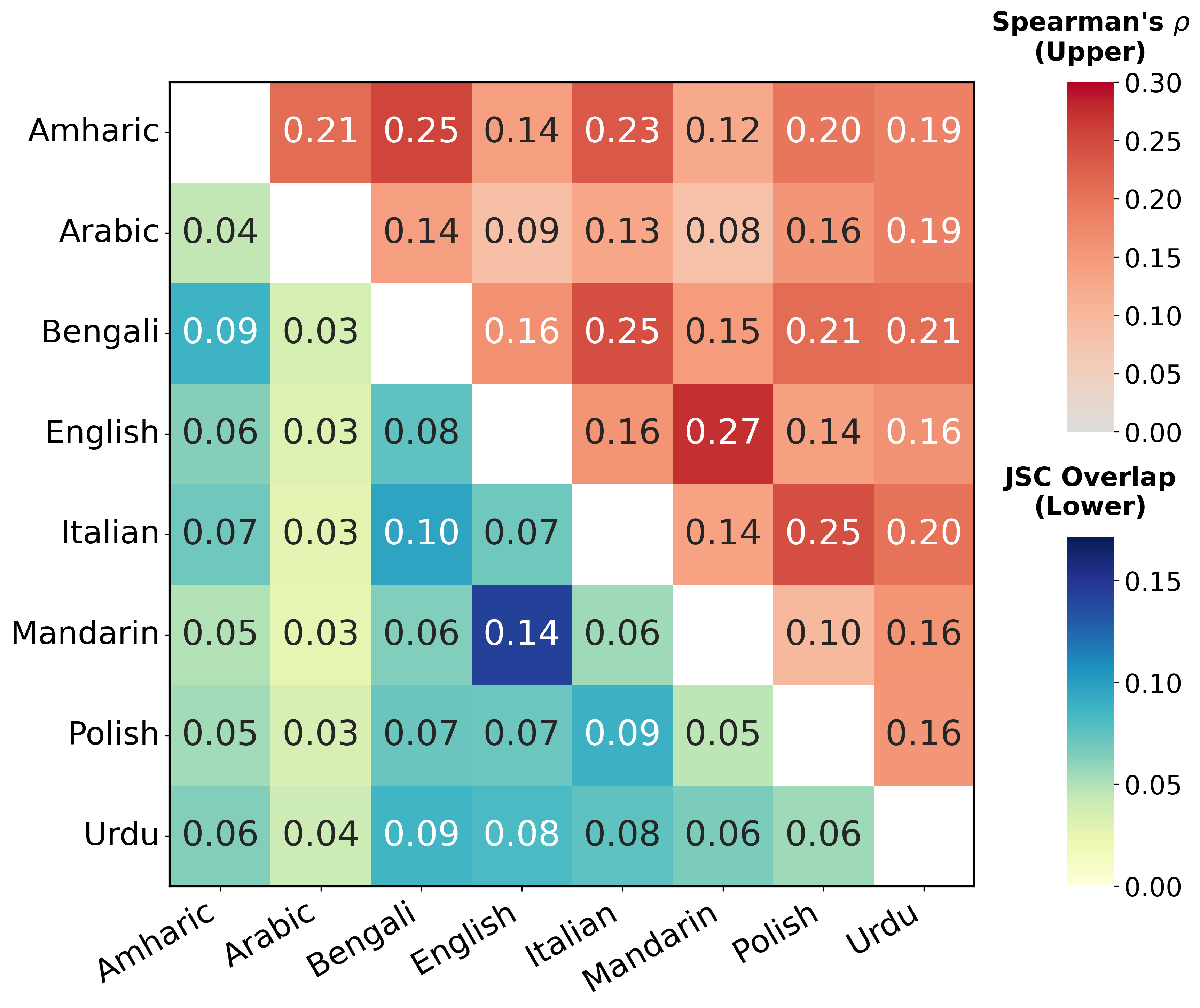}
        \caption{Audio-Flamingo-3}
    \end{subfigure}\hfill
    \begin{subfigure}[b]{0.5\columnwidth}
        \centering
        \includegraphics[width=\linewidth]{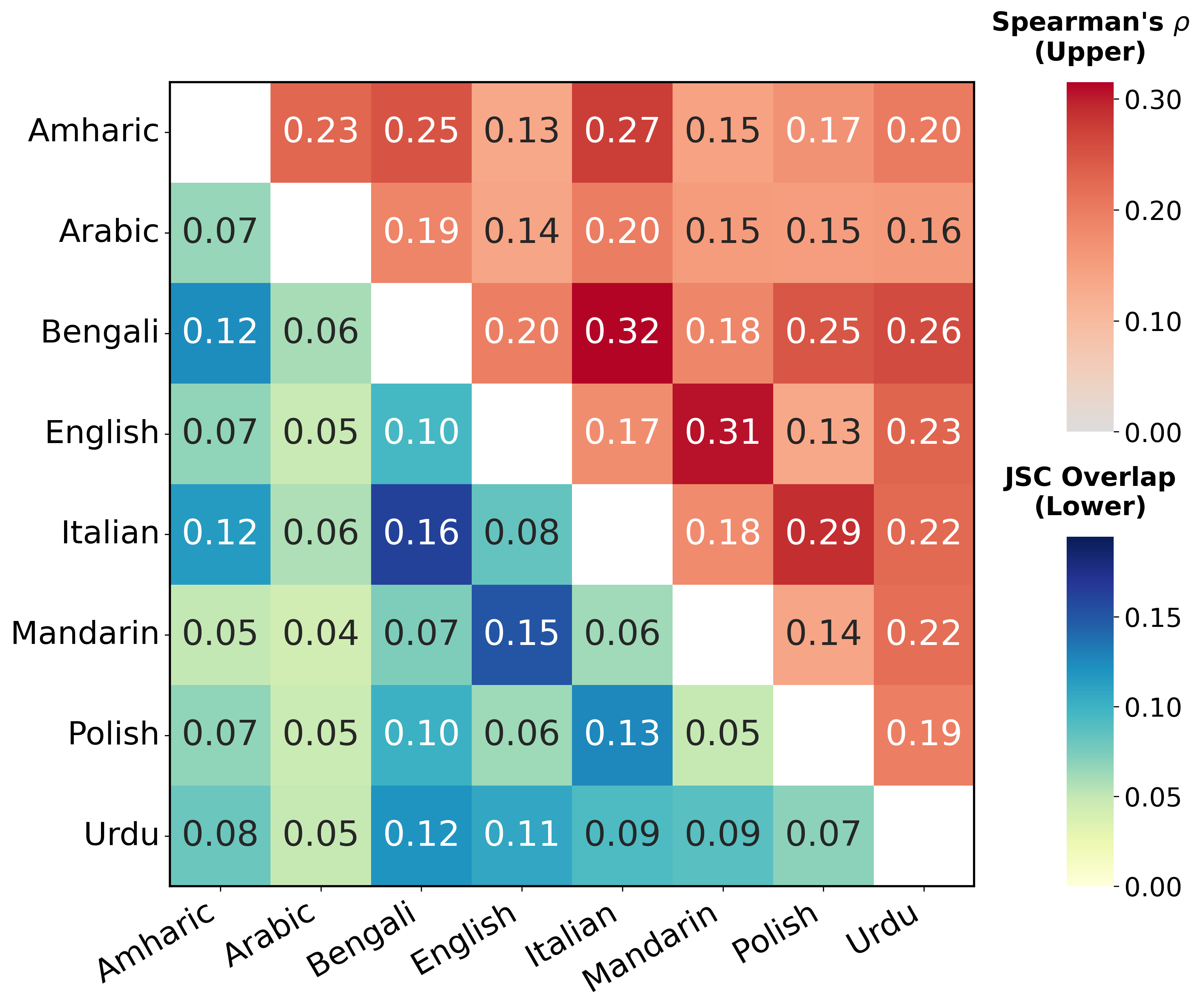}
        \caption{Kimi-Audio}
    \end{subfigure}
    
    \caption{Cross-lingual agreement of monolingually-identified ESNs across two models (see other two in Appendix~\ref{app:agreement}). Each heatmap displays JSC (lower triangle), measuring neuron set intersection, and Spearman's $\rho$ (upper triangle), measuring agreement in selectivity score rankings. All neurons selected using ConAct with $r{=}0.5\%$.}
    \label{fig:descriptive_stat}
\vspace{-4mm}
\end{figure}

We first examine whether ESNs identified independently per language exhibit cross-lingual agreement in both ranking structure and set membership. 
Figure~\ref{fig:descriptive_stat} present pairwise Jaccard Similarity Coefficient (JSC) and Spearman's rank correlation ($\rho$) for all four LALMs across eight identification languages. Across all models, we observe a consistent dissociation: the exact set intersection of top-ranked neurons (discrete identity overlap) is minimal (JSC predominantly below 0.10, with occasional pairs reaching 0.15--0.16), while the underlying selectivity scores retain a shared ranking structure ($\rho$ mostly between 0.10 and 0.25 for Audio-Flamingo-3 and Kimi-Audio, and up to $\approx$0.6 for MiniCPM-o-4.5).

\subsection{Saturation of Monolingual Evidence}
\label{sec:saturation}
\begin{figure}[ht!]
    \centering
    \begin{subfigure}[b]{0.5\columnwidth}
        \centering
        \includegraphics[width=\linewidth]{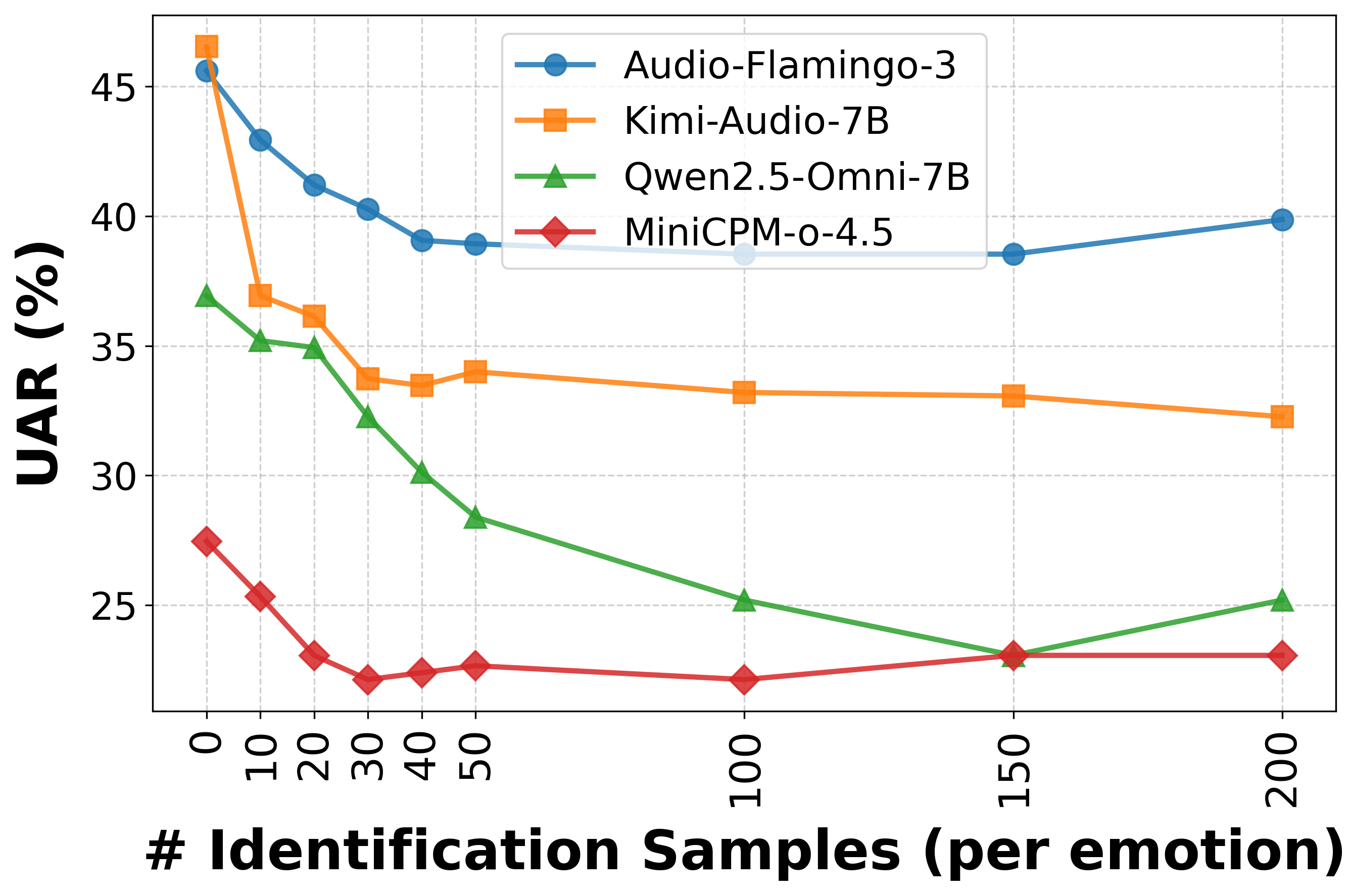}
        \caption{MSP-Podcast (Deact.)}
    \end{subfigure}\hfill
    \begin{subfigure}[b]{0.5\columnwidth}
        \centering
        \includegraphics[width=\linewidth]{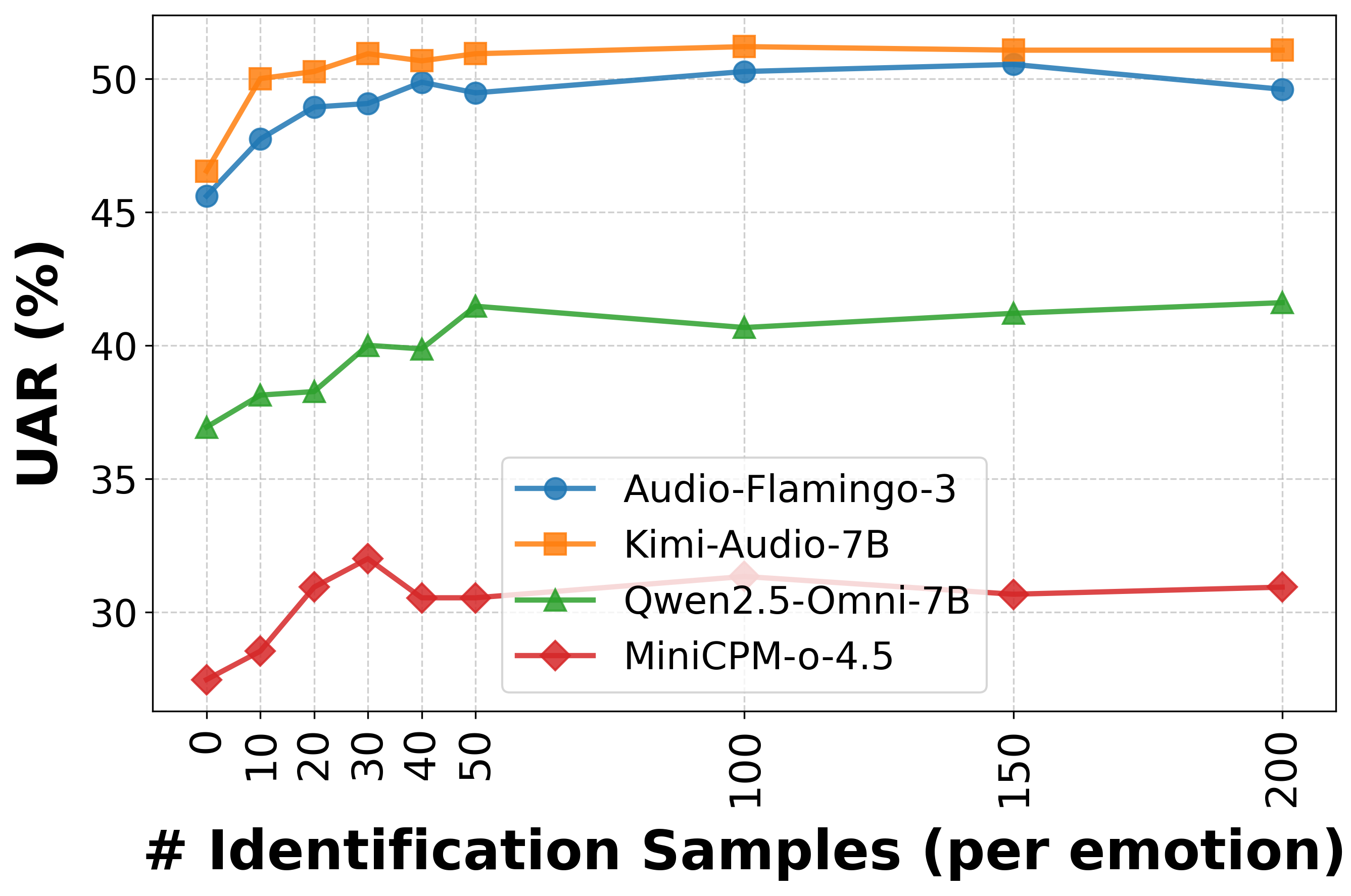}
        \caption{MSP-Podcast (Steering)}
    \end{subfigure}\hfill
    \caption{Saturation of monolingual evidence: UAR (\%) under intervention as identification sample size grows (MSP-Podcast; BIIC in Appendix~\ref{app:saturation} shows the same plateau beyond ${\sim}50$ instances).}
    \label{fig:saturation}
    \vspace{-4mm}
\end{figure}

We evaluate the causal properties of identified neurons through functional intervention. To establish monolingual baselines before introducing fusion, Figure~\ref{fig:saturation} shows the causal efficacy of ESN masks identified using varying numbers of correctly predicted instances from the Identification Set ($\mathcal{L}_\text{id}$). Across all evaluated models, intervention effects plateau rapidly beyond approximately 50 instances.

The bottleneck for isolating transferable affective units is therefore the
linguistic scope of the identification data rather than sample volume,
making cross-lingual evidence integration essential. Accordingly, all subsequent experiments use a fixed budget of $c{=}50$ instances per emotion to focus on the causal benefits of cross-lingual consistency (see Appendix~\ref{app:id_budget} for exact counts and justification).

\subsection{Causal Validation of MLENs Through Targeted Intervention}

\begin{table*}[ht!]
\centering
\resizebox{\textwidth}{!}{%
\pgfkeys{/pgf/fpu=true}
\newcommand{\gc}[1]{%
  \pgfmathsetmacro{\val}{#1}%
  \pgfmathtruncatemacro{\intensity}{max(0, min(60, (0 - \val) * 3))}%
  \ifnum\intensity>0
    \cellcolor{deactcolor!\intensity}#1%
  \else
    \pgfmathtruncatemacro{\posintensity}{min(25, \val * 25)}%
    \cellcolor{steercolor!\posintensity}#1%
  \fi
}%
\begin{tabular}{@{}llccccccccccc@{}}
\toprule
\multirow{2}{*}{LALM} & \multirow{2}{*}{Group} & \multicolumn{8}{c}{Monolingual ESN Masks} & \multicolumn{3}{c}{Fusion MLEN Masks} \\ 
\cmidrule(lr){3-10} \cmidrule(lr){11-13}
& & English & Mandarin & Amharic & Arabic & Bengali & Italian & Polish & Urdu & Overlap & Joint & CR ($\lambda{=}0.3$) \\ 
\midrule
Audio & \multicolumn{1}{l|}{$\mathcal{L}_\text{id}$} 
  & \cellcolor{deactcolor!18}-5.92 & \cellcolor{deactcolor!12}-3.93 & \cellcolor{deactcolor!12}-3.85 & \cellcolor{deactcolor!4}-1.43 & \cellcolor{deactcolor!9}-2.92 & \cellcolor{deactcolor!6}-2.03 & \cellcolor{deactcolor!14}-4.70 & \cellcolor{deactcolor!18}-6.16 
  & \cellcolor{deactcolor!30}\textbf{-10.15} & \cellcolor{deactcolor!28}-9.20 & \cellcolor{deactcolor!30}-9.92 \\
Flamingo 3 & \multicolumn{1}{l|}{$\mathcal{L}_\text{held}$} 
  & \cellcolor{deactcolor!17}-5.64 & \cellcolor{deactcolor!15}-4.84 & \cellcolor{deactcolor!10}-3.31 & \cellcolor{deactcolor!4}-1.34 & \cellcolor{deactcolor!10}-3.44 & \cellcolor{deactcolor!5}-1.64 & \cellcolor{deactcolor!13}-4.33 & \cellcolor{deactcolor!14}-4.62 
  & \cellcolor{deactcolor!27}-8.86 & \cellcolor{deactcolor!29}-9.57 & \cellcolor{deactcolor!30}\textbf{-9.93} \\ 
\cmidrule(l){2-13} 
& \multicolumn{1}{l|}{Avg.} 
  & \cellcolor{deactcolor!17}-5.83 {\scriptsize (2.05)} & \cellcolor{deactcolor!13}-4.24 {\scriptsize (1.26)} & \cellcolor{deactcolor!11}-3.67 {\scriptsize (1.16)} & \cellcolor{deactcolor!4}-1.40 {\scriptsize (1.06)} & \cellcolor{deactcolor!9}-3.09 {\scriptsize (0.95)} & \cellcolor{deactcolor!6}-1.90 {\scriptsize (0.65)} & \cellcolor{deactcolor!14}-4.58 {\scriptsize (1.13)} & \cellcolor{deactcolor!17}-5.65 {\scriptsize (2.20)} 
  & \cellcolor{deactcolor!29}-9.72 {\scriptsize (4.17)} & \cellcolor{deactcolor!28}-9.33 {\scriptsize (2.30)} & \cellcolor{deactcolor!30}\textbf{-9.92 {\scriptsize (2.57)}} \\ 
\midrule
Kimi & \multicolumn{1}{l|}{$\mathcal{L}_\text{id}$} 
  & \cellcolor{deactcolor!39}-13.00 & \cellcolor{deactcolor!36}-12.05 & \cellcolor{deactcolor!22}-7.22 & \cellcolor{deactcolor!11}-3.83 & \cellcolor{deactcolor!32}-10.81 & \cellcolor{deactcolor!38}-12.80 & \cellcolor{deactcolor!15}-4.85 & \cellcolor{deactcolor!39}-12.97 
  & \cellcolor{deactcolor!21}-6.93 & \cellcolor{deactcolor!50}-16.68 & \cellcolor{deactcolor!50}\textbf{-16.75} \\
-Audio & \multicolumn{1}{l|}{$\mathcal{L}_\text{held}$} 
  & \cellcolor{deactcolor!38}-12.80 & \cellcolor{deactcolor!38}-12.54 & \cellcolor{deactcolor!17}-5.82 & \cellcolor{deactcolor!11}-3.63 & \cellcolor{deactcolor!30}-9.98 & \cellcolor{deactcolor!41}-13.64 & \cellcolor{deactcolor!10}-3.47 & \cellcolor{deactcolor!45}-15.02 
  & \cellcolor{deactcolor!23}-7.66 & \cellcolor{deactcolor!53}-17.58 & \cellcolor{deactcolor!58}\textbf{-19.31} \\ 
\cmidrule(l){2-13} 
& \multicolumn{1}{l|}{Avg.} 
  & \cellcolor{deactcolor!39}-12.93 {\scriptsize (3.68)} & \cellcolor{deactcolor!37}-12.21 {\scriptsize (3.54)} & \cellcolor{deactcolor!20}-6.75 {\scriptsize (2.40)} & \cellcolor{deactcolor!11}-3.76 {\scriptsize (1.42)} & \cellcolor{deactcolor!32}-10.53 {\scriptsize (2.85)} & \cellcolor{deactcolor!39}-13.08 {\scriptsize (3.67)} & \cellcolor{deactcolor!13}-4.39 {\scriptsize (1.77)} & \cellcolor{deactcolor!41}-13.66 {\scriptsize (3.13)} 
  & \cellcolor{deactcolor!22}-7.18 {\scriptsize (2.03)} & \cellcolor{deactcolor!51}-16.98 {\scriptsize (4.28)} & \cellcolor{deactcolor!53}\textbf{-17.60 {\scriptsize (4.50)}} \\ 
\midrule
MiniCPM & \multicolumn{1}{l|}{$\mathcal{L}_\text{id}$} 
  & \cellcolor{deactcolor!18}-6.02 & \cellcolor{deactcolor!20}-6.72 & \cellcolor{deactcolor!6}-2.07 & \cellcolor{deactcolor!2}-0.57 & \cellcolor{deactcolor!9}-2.88 & \cellcolor{deactcolor!8}-2.61 & \cellcolor{deactcolor!8}-2.68 & \cellcolor{deactcolor!19}-6.22 
  & \cellcolor{deactcolor!13}-4.39 & \cellcolor{deactcolor!17}-5.62 & \cellcolor{deactcolor!22}\textbf{-7.29} \\
-o-4.5 & \multicolumn{1}{l|}{$\mathcal{L}_\text{held}$} 
  & \cellcolor{deactcolor!24}-7.86 & \cellcolor{deactcolor!28}-9.47 & \cellcolor{deactcolor!10}-3.41 & \cellcolor{deactcolor!2}-0.74 & \cellcolor{deactcolor!10}-3.36 & \cellcolor{deactcolor!13}-4.41 & \cellcolor{deactcolor!8}-2.83 & \cellcolor{deactcolor!25}-8.29 
  & \cellcolor{deactcolor!16}-5.35 & \cellcolor{deactcolor!23}-7.56 & \cellcolor{deactcolor!29}\textbf{-9.51} \\ 
\cmidrule(l){2-13} 
& \multicolumn{1}{l|}{Avg.} 
  & \cellcolor{deactcolor!20}-6.63 {\scriptsize (2.98)} & \cellcolor{deactcolor!23}-7.64 {\scriptsize (3.99)} & \cellcolor{deactcolor!8}-2.52 {\scriptsize (2.09)} & \cellcolor{deactcolor!2}-0.63 {\scriptsize (0.55)} & \cellcolor{deactcolor!9}-3.04 {\scriptsize (1.34)} & \cellcolor{deactcolor!10}-3.21 {\scriptsize (2.15)} & \cellcolor{deactcolor!8}-2.73 {\scriptsize (2.40)} & \cellcolor{deactcolor!21}-6.91 {\scriptsize (3.89)} 
  & \cellcolor{deactcolor!14}-4.71 {\scriptsize (2.34)} & \cellcolor{deactcolor!19}-6.26 {\scriptsize (3.49)} & \cellcolor{deactcolor!24}\textbf{-8.03 {\scriptsize (4.42)}} \\ 
\midrule
Qwen2.5 & \multicolumn{1}{l|}{$\mathcal{L}_\text{id}$} 
  & \cellcolor{deactcolor!21}\textbf{-7.09} & \cellcolor{deactcolor!19}-6.28 & \cellcolor{deactcolor!20}-6.70 & \cellcolor{deactcolor!2}-0.83 & \cellcolor{deactcolor!9}-3.10 & \cellcolor{deactcolor!7}-2.37 & \cellcolor{deactcolor!1}-0.28 & \cellcolor{deactcolor!6}-1.97 
  & \cellcolor{steercolor!17}0.68 & \cellcolor{deactcolor!18}-5.91 & \cellcolor{deactcolor!20}-6.68 \\
-Omni-7B & \multicolumn{1}{l|}{$\mathcal{L}_\text{held}$} 
  & \cellcolor{deactcolor!20}-6.77 & \cellcolor{deactcolor!18}-5.91 & \cellcolor{deactcolor!18}-6.02 & \cellcolor{deactcolor!2}-0.57 & \cellcolor{deactcolor!10}-3.38 & \cellcolor{deactcolor!11}-3.82 & \cellcolor{steercolor!7}0.28 & \cellcolor{deactcolor!12}-3.84 
  & \cellcolor{steercolor!10}0.40 & \cellcolor{deactcolor!19}-6.34 & \cellcolor{deactcolor!24}\textbf{-7.86} \\ 
\cmidrule(l){2-13} 
& \multicolumn{1}{l|}{Avg.} 
  & \cellcolor{deactcolor!21}-6.99 {\scriptsize (2.28)} & \cellcolor{deactcolor!18}-6.16 {\scriptsize (1.82)} & \cellcolor{deactcolor!19}-6.48 {\scriptsize (1.17)} & \cellcolor{deactcolor!2}-0.74 {\scriptsize (1.54)} & \cellcolor{deactcolor!10}-3.19 {\scriptsize (1.95)} & \cellcolor{deactcolor!9}-2.85 {\scriptsize (1.31)} & \cellcolor{deactcolor!0}-0.09 {\scriptsize (1.16)} & \cellcolor{deactcolor!8}-2.59 {\scriptsize (1.90)} 
  & \cellcolor{steercolor!15}0.58 {\scriptsize (1.64)} & \cellcolor{deactcolor!18}-6.05 {\scriptsize (0.78)} & \cellcolor{deactcolor!21}\textbf{-7.07 {\scriptsize (2.16)}} \\ 
\bottomrule
\end{tabular}%
}
\caption{Deactivation effects on ESS. ``Avg.'' rows denote means across all languages regardless of group; standard deviation in parentheses. Negative values indicate reduced specificity, and color intensity indicates effect magnitude. Per-language results in Appendix~\ref{appendix:per_emotion} Table~\ref{tab:deact_results_full}.}
\label{tab:deact_results}
\end{table*}

\begin{table*}[ht!]
\centering
\resizebox{\textwidth}{!}{%
\begin{tabular}{@{}llccccccccccc@{}}
\toprule
\multirow{2}{*}{LALM} & \multirow{2}{*}{Group} & \multicolumn{8}{c}{Monolingual ESN Masks} & \multicolumn{3}{c}{Fusion MLEN Masks} \\ 
\cmidrule(lr){3-10} \cmidrule(lr){11-13}
& & English & Mandarin & Amharic & Arabic & Bengali & Italian & Polish & Urdu & Overlap & Joint & CR ($\lambda{=}0.3$) \\ 
\midrule
Audio & \multicolumn{1}{l|}{$\mathcal{L}_\text{id}$} 
  & \cellcolor{steercolor!30}3.72 & \cellcolor{steercolor!17}2.11 & \cellcolor{steercolor!15}1.89 & \cellcolor{steercolor!4}0.45 & \cellcolor{steercolor!12}1.51 & \cellcolor{steercolor!7}0.91 & \cellcolor{steercolor!23}2.82 & \cellcolor{steercolor!29}3.60 
  & \cellcolor{steercolor!18}2.22 & \cellcolor{steercolor!36}4.50 & \cellcolor{steercolor!37}\textbf{4.68} \\
Flamingo 3 & \multicolumn{1}{l|}{$\mathcal{L}_\text{held}$} 
  & \cellcolor{steercolor!22}2.72 & \cellcolor{steercolor!16}2.04 & \cellcolor{steercolor!13}1.66 & \cellcolor{steercolor!2}0.28 & \cellcolor{steercolor!11}1.40 & \cellcolor{steercolor!10}1.19 & \cellcolor{steercolor!14}1.73 & \cellcolor{steercolor!18}2.25 
  & \cellcolor{steercolor!12}1.51 & \cellcolor{steercolor!30}3.80 & \cellcolor{steercolor!32}\textbf{4.00} \\ 
\cmidrule(l){2-13} 
& \multicolumn{1}{l|}{Avg.} 
  & \cellcolor{steercolor!27}3.39 {\scriptsize (1.11)} & \cellcolor{steercolor!17}2.09 {\scriptsize (0.63)} & \cellcolor{steercolor!14}1.81 {\scriptsize (0.90)} & \cellcolor{steercolor!3}0.39 {\scriptsize (0.41)} & \cellcolor{steercolor!12}1.48 {\scriptsize (0.59)} & \cellcolor{steercolor!8}1.00 {\scriptsize (0.54)} & \cellcolor{steercolor!20}2.46 {\scriptsize (1.08)} & \cellcolor{steercolor!25}3.15 {\scriptsize (1.20)} 
  & \cellcolor{steercolor!16}1.98 {\scriptsize (0.88)} & \cellcolor{steercolor!34}4.27 {\scriptsize (1.21)} & \cellcolor{steercolor!36}\textbf{4.45 {\scriptsize (1.26)}} \\ 
\midrule
Kimi & \multicolumn{1}{l|}{$\mathcal{L}_\text{id}$} 
  & \cellcolor{steercolor!43}5.36 & \cellcolor{steercolor!42}5.19 & \cellcolor{steercolor!26}3.22 & \cellcolor{steercolor!17}2.12 & \cellcolor{steercolor!38}4.80 & \cellcolor{steercolor!41}5.08 & \cellcolor{steercolor!18}2.19 & \cellcolor{steercolor!45}5.58 
  & \cellcolor{steercolor!26}3.28 & \cellcolor{steercolor!58}\textbf{7.25} & \cellcolor{steercolor!58}\textbf{7.25} \\
-Audio & \multicolumn{1}{l|}{$\mathcal{L}_\text{held}$} 
  & \cellcolor{steercolor!38}4.70 & \cellcolor{steercolor!36}4.53 & \cellcolor{steercolor!19}2.43 & \cellcolor{steercolor!16}1.96 & \cellcolor{steercolor!33}4.15 & \cellcolor{steercolor!32}4.05 & \cellcolor{steercolor!14}1.70 & \cellcolor{steercolor!41}5.15 
  & \cellcolor{steercolor!27}3.39 & \cellcolor{steercolor!50}6.29 & \cellcolor{steercolor!51}\textbf{6.34} \\ 
\cmidrule(l){2-13} 
& \multicolumn{1}{l|}{Avg.} 
  & \cellcolor{steercolor!41}5.14 {\scriptsize (1.58)} & \cellcolor{steercolor!40}4.97 {\scriptsize (1.57)} & \cellcolor{steercolor!24}2.96 {\scriptsize (1.30)} & \cellcolor{steercolor!17}2.07 {\scriptsize (1.05)} & \cellcolor{steercolor!37}4.58 {\scriptsize (1.72)} & \cellcolor{steercolor!38}4.74 {\scriptsize (1.46)} & \cellcolor{steercolor!16}2.03 {\scriptsize (1.09)} & \cellcolor{steercolor!44}5.44 {\scriptsize (1.45)} 
  & \cellcolor{steercolor!27}3.32 {\scriptsize (1.04)} & \cellcolor{steercolor!55}6.93 {\scriptsize (2.02)} & \cellcolor{steercolor!56}\textbf{6.95 {\scriptsize (1.74)}} \\ 
\midrule
MiniCPM & \multicolumn{1}{l|}{$\mathcal{L}_\text{id}$} 
  & \cellcolor{steercolor!33}4.18 & \cellcolor{steercolor!41}5.15 & \cellcolor{steercolor!16}1.98 & \cellcolor{steercolor!2}0.26 & \cellcolor{steercolor!17}2.10 & \cellcolor{steercolor!18}2.21 & \cellcolor{steercolor!22}2.81 & \cellcolor{steercolor!41}5.17 
  & \cellcolor{steercolor!21}2.59 & \cellcolor{steercolor!34}4.29 & \cellcolor{steercolor!48}\textbf{5.97} \\
-o-4.5 & \multicolumn{1}{l|}{$\mathcal{L}_\text{held}$} 
  & \cellcolor{steercolor!38}4.70 & \cellcolor{steercolor!44}5.48 & \cellcolor{steercolor!21}2.65 & \cellcolor{steercolor!2}-0.09 & \cellcolor{steercolor!15}1.82 & \cellcolor{steercolor!21}2.66 & \cellcolor{steercolor!15}1.92 & \cellcolor{steercolor!42}5.24 
  & \cellcolor{steercolor!22}2.74 & \cellcolor{steercolor!31}3.88 & \cellcolor{steercolor!52}\textbf{6.48} \\ 
\cmidrule(l){2-13} 
& \multicolumn{1}{l|}{Avg.} 
  & \cellcolor{steercolor!35}4.35 {\scriptsize (1.58)} & \cellcolor{steercolor!42}5.26 {\scriptsize (1.88)} & \cellcolor{steercolor!18}2.20 {\scriptsize (1.39)} & \cellcolor{steercolor!1}0.15 {\scriptsize (0.43)} & \cellcolor{steercolor!16}2.01 {\scriptsize (0.67)} & \cellcolor{steercolor!19}2.36 {\scriptsize (1.15)} & \cellcolor{steercolor!20}2.51 {\scriptsize (1.26)} & \cellcolor{steercolor!42}5.20 {\scriptsize (2.07)} 
  & \cellcolor{steercolor!21}2.64 {\scriptsize (1.00)} & \cellcolor{steercolor!33}4.15 {\scriptsize (1.24)} & \cellcolor{steercolor!49}\textbf{6.14 {\scriptsize (2.48)}} \\ 
\midrule
Qwen2.5 & \multicolumn{1}{l|}{$\mathcal{L}_\text{id}$} 
  & \cellcolor{steercolor!32}\textbf{3.94} & \cellcolor{steercolor!29}3.68 & \cellcolor{steercolor!27}3.35 & \cellcolor{steercolor!2}0.26 & \cellcolor{steercolor!10}1.28 & \cellcolor{steercolor!9}1.13 & \cellcolor{steercolor!2}0.24 & \cellcolor{steercolor!12}1.45 
  & \cellcolor{steercolor!4}-0.22 & \cellcolor{steercolor!31}3.87 & \cellcolor{steercolor!27}3.39 \\
-Omni-7B & \multicolumn{1}{l|}{$\mathcal{L}_\text{held}$} 
  & \cellcolor{steercolor!25}3.14 & \cellcolor{steercolor!32}\textbf{3.97} & \cellcolor{steercolor!24}3.01 & \cellcolor{steercolor!1}0.15 & \cellcolor{steercolor!2}0.23 & \cellcolor{steercolor!15}1.91 & \cellcolor{steercolor!0}-0.01 & \cellcolor{steercolor!16}2.03 
  & \cellcolor{steercolor!2}0.29 & \cellcolor{steercolor!29}3.58 & \cellcolor{steercolor!29}3.64 \\ 
\cmidrule(l){2-13} 
& \multicolumn{1}{l|}{Avg.} 
  & \cellcolor{steercolor!29}3.68 {\scriptsize (0.93)} & \cellcolor{steercolor!30}\textbf{3.78 {\scriptsize (0.81)}} & \cellcolor{steercolor!26}3.24 {\scriptsize (0.77)} & \cellcolor{steercolor!2}0.23 {\scriptsize (0.69)} & \cellcolor{steercolor!7}0.93 {\scriptsize (1.35)} & \cellcolor{steercolor!11}1.39 {\scriptsize (0.60)} & \cellcolor{steercolor!1}0.16 {\scriptsize (0.94)} & \cellcolor{steercolor!13}1.65 {\scriptsize (0.96)} 
  & \cellcolor{steercolor!1}-0.05 {\scriptsize (1.01)} & \cellcolor{steercolor!30}3.77 {\scriptsize (0.95)} & \cellcolor{steercolor!28}3.47 {\scriptsize (1.14)} \\ 
\bottomrule
\end{tabular}%
}
\caption{Steering effects on ESS with $\alpha{=}0.5$. ``Avg.'' rows denote means across all languages regardless of group; standard deviation in parentheses. Positive values indicate enhanced specificity, and color intensity indicates effect magnitude. Per-language results in Appendix~\ref{appendix:per_emotion} Table~\ref{tab:steer_results_full}.}
\label{tab:steer_results}
\vspace{-4mm}
\end{table*}

We evaluate the functional necessity and influence of the identified neurons through two targeted interventions: deactivation (to assess necessity for recognition) and activation steering (to assess the neurons' capacity to modulate model behavior).

\textbf{Deactivation.} To assess whether identified neurons contribute causally to emotion recognition, we measure the reduction in ESS under deactivation, which are summarized in Table~\ref{tab:deact_results} across two language groups. CR-Fusion ($\lambda{=}0.3$) exceeds the best monolingual mask for three models in both groups. Joint Fusion does so on Audio-Flamingo-3 and Kimi-Audio, but falls short of the strongest monolingual mask on MiniCPM-o-4.5 ($-6.26$ vs.\ $-7.64$) and Qwen2.5-Omni-7B ($-6.05$ vs.\ $-6.99$), consistent with its token-weighted pooling being dominated by high-resource conditions. Overlap Fusion is the weakest fusion strategy on three of four models and reverses sign on Qwen2.5-Omni-7B ($+0.58$); Audio-Flamingo-3 is the exception, where the consensus set remains strongly causal.

\textbf{Steering.}
Table~\ref{tab:steer_results} shows steering effects across two language groups. 
CR-Fusion ($\lambda$=0.3) yields superior steering efficacy in $\mathcal{L}_\text{held}$ for three of the four LALMs. For MiniCPM-o-4.5, CR-Fusion achieves $+6.48$ pp in held-out languages, substantially outperforming English ($+4.70$ pp) and Mandarin ($+5.48$ pp) baselines. The exception is Qwen2.5-Omni-7B, where the Mandarin mask remains marginally stronger ($+3.97$ vs.\ $+3.64$ pp); this is consistent with Qwen having the lowest language-invariant share (\S3.4, Appendix~\ref{app:decomp}), so there is less invariant structure for fusion to recover. Together, these results provide causal evidence that CR-Fusion identifies functionally important, language-shared affective units.

\section{Discussion}
\label{sec:discussion}
\subsection{Emotion-Level Heterogeneity in Cross-Lingual Generalization}
\label{sec:heterogeneity}

Causal interventions reveal substantial heterogeneity across emotions. Anger, happiness, and sadness exhibit the largest ESS magnitudes across all evaluated LALMs under both deactivation and steering, and this ordering is preserved across the identification and held-out groups (Appendix~\ref{app:emotion_detail}). Fear and neutral show weaker and more variable effects; for fear, this plausibly reflects acoustic overlap with high-arousal anger and a floor effect from near-zero baseline accuracy in several model--language pairs (Appendix~\ref{app:baseline}), which leaves few correctly predicted instances available for identification (Appendix~\ref{app:id_budget}) rather than a shortage of fear utterances in the corpora themselves (Table~\ref{tab:decomp_emotion}). Strong causal potency does not, however, imply more language-invariant encoding: per-emotion invariant shares are highest for neutral and fear, so the causal ordering is better explained by baseline accuracy and class support than by representational universality.

\begin{figure}[ht!]
    \centering
    \begin{subfigure}[b]{0.5\columnwidth}
        \centering
        \includegraphics[width=\linewidth]{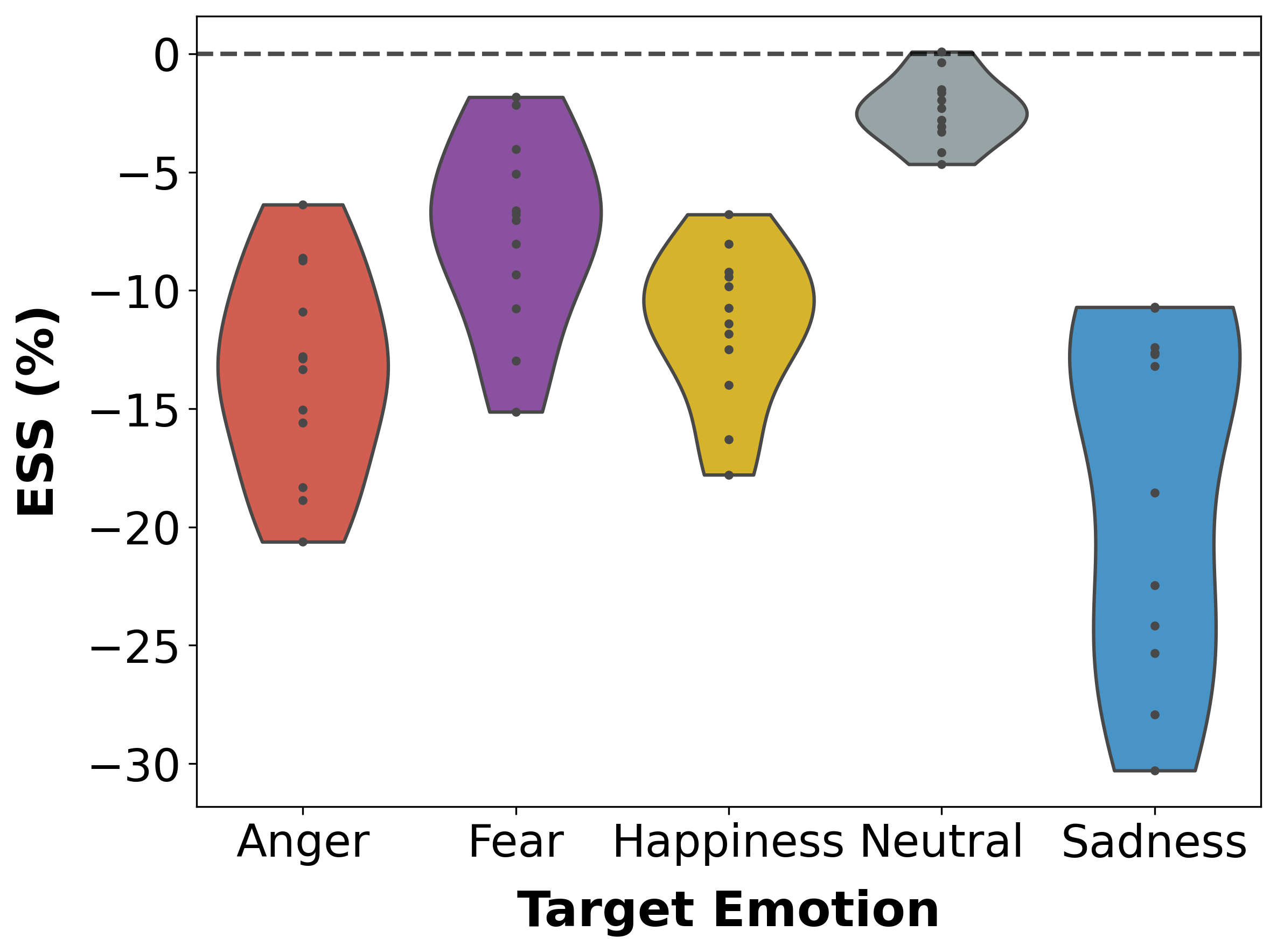}
        \caption{Deactivation}
    \end{subfigure}\hfill
    \begin{subfigure}[b]{0.5\columnwidth}
        \centering
        \includegraphics[width=\linewidth]{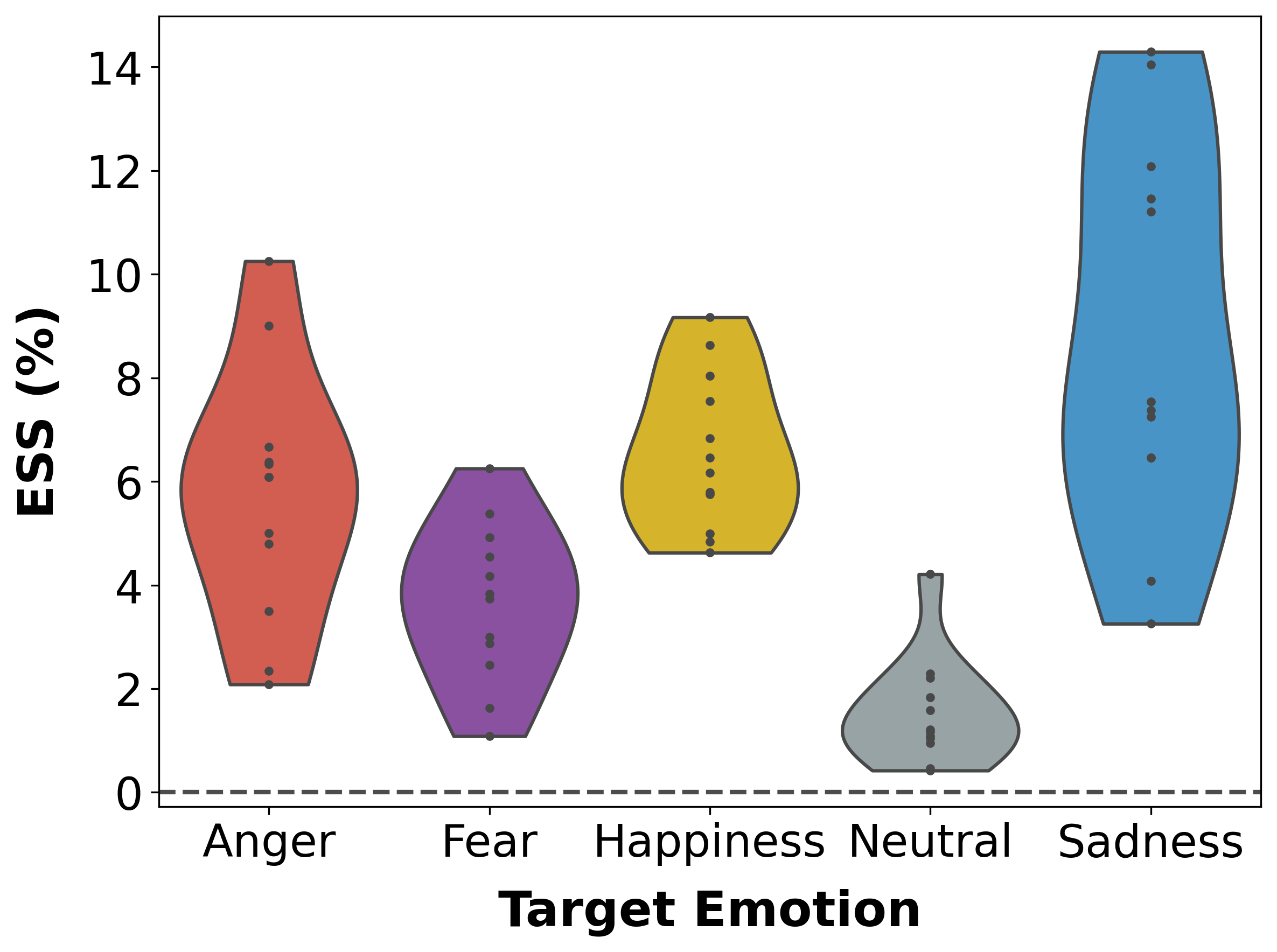}
        \caption{Steering, $\alpha{=}0.5$}
    \end{subfigure}
    \caption{ESS distributions under (a) deactivation and (b) steering,
    comparing CR-Fusion ($\lambda{=}0.3$) against monolingual baselines
    across all evaluated emotions. Per-emotion magnitudes are provided in Appendix~\ref{app:emotion_detail}.}
    \label{fig:emotion_generalization}
    \vspace{-4mm}
\end{figure}

Independently of these differences, Figure~\ref{fig:emotion_generalization} shows that CR-Fusion ($\lambda{=}0.3$) matches or exceeds monolingual baselines for every emotion in both groups. Consistent with \S\ref{sec:theory}, this advantage does not come from suppressing cross-lingual variance (the penalty is near zero at the operating point) but from pooling multiple corpora to estimate the language-invariant component.

\subsection{The Impact of Consistency Penalty Strength}
\label{sec:lambda}
    \vspace{-4mm}

\begin{figure}[ht!]
    \centering
    \begin{subfigure}[b]{0.5\columnwidth}
        \centering
        \includegraphics[width=\linewidth]{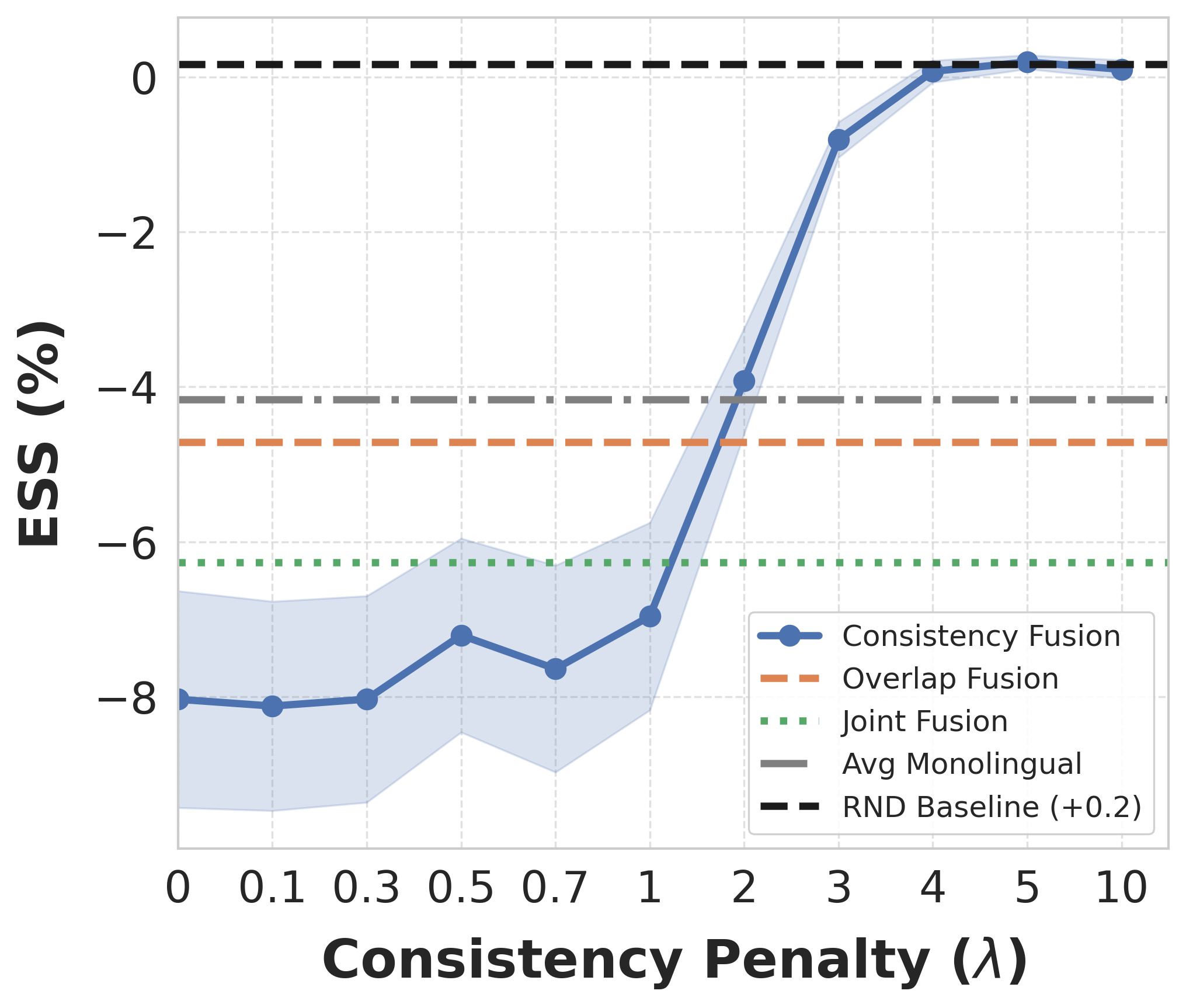}
        \caption{MiniCPM (Deact.)}
    \end{subfigure}\hfill
    \begin{subfigure}[b]{0.5\columnwidth}
        \centering
        \includegraphics[width=\linewidth]{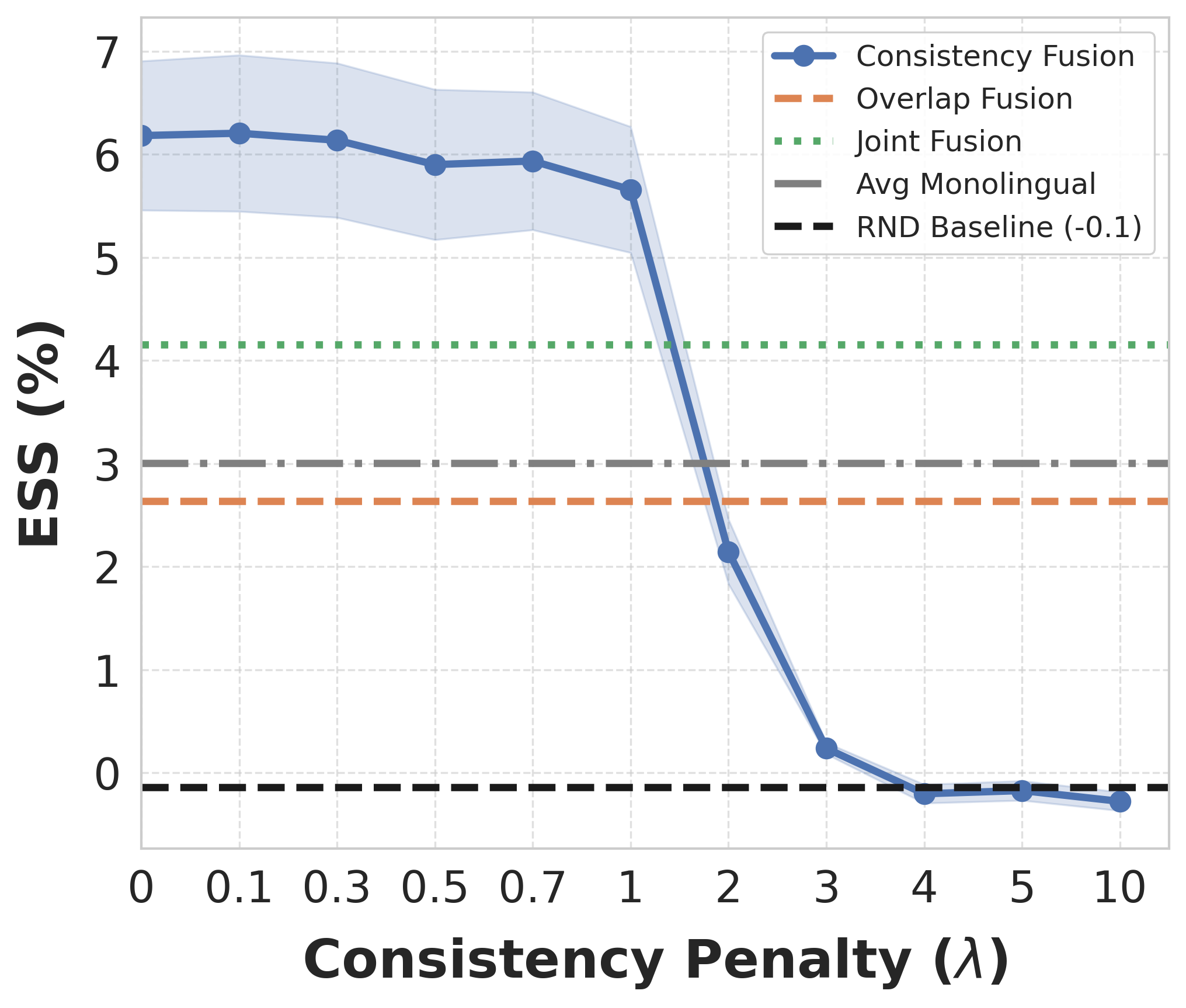}
        \caption{MiniCPM (Steering)}
    \end{subfigure}\hfill

    \caption{Sensitivity of intervention effects to consistency penalty $\lambda$. Average ESS ($\pm1$ standard error of the mean across 12 languages, shown as the shaded band) under (a) deactivation and (b) steering for MiniCPM. Results for other LALMs are provided in Appendix~\ref{appendix:supplementary}.}
    \label{fig:lambda}
    \vspace{-2mm}
\end{figure}

We analyze the sensitivity of causal effects to the consistency penalty $\lambda$, which trades aggregate selector strength ($\mu$) against cross-lingual stability (low $\sigma$). Figure~\ref{fig:lambda} and
Appendix~\ref{app:lambda} Figure~\ref{fig:lambda_more} show that intervention effects are strongest at small $\lambda$ and degrade sharply beyond $\lambda \approx 1$, approaching the random-selection
baseline by $\lambda \approx 3$ for all four LALMs; for Qwen2.5-Omni-7B under deactivation the effect reverses sign, indicating that at large $\lambda$ the selected units are no longer emotion-selective at all. We therefore adopt $\lambda{=}0.3$ as the operating point, where the penalty is active but weak.

This profile favors both predictions of \S\ref{sec:theory}: the optimal penalty is near zero under the expected-transfer objective, and large $\lambda$ preferentially discards low-resource conditions while shifting selection toward uniformly weak units. Overlap Fusion is correspondingly the weakest fusion strategy in Table~\ref{tab:steer_results} for all four models and in Table~\ref{tab:deact_results} for three.

\subsection{Language Contributions and Asymmetric Transfer}
\label{sec:loo}

\begin{figure}[ht!]
    \centering
    \begin{subfigure}[b]{0.5\columnwidth}
        \centering
        \includegraphics[width=\linewidth]{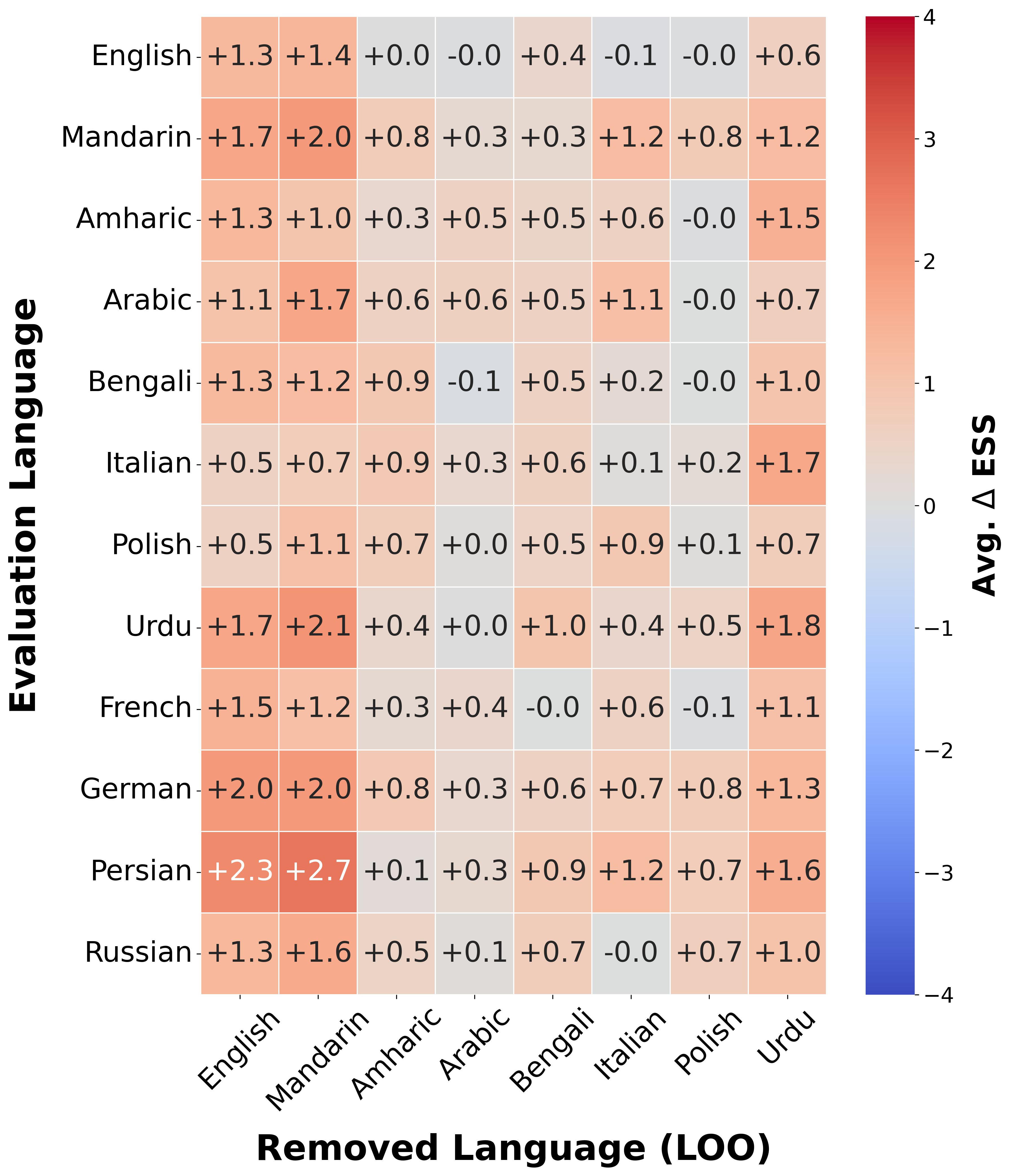}
        \caption{Deactivation}
    \end{subfigure}\hfill
    \begin{subfigure}[b]{0.5\columnwidth}
        \centering
        \includegraphics[width=\linewidth]{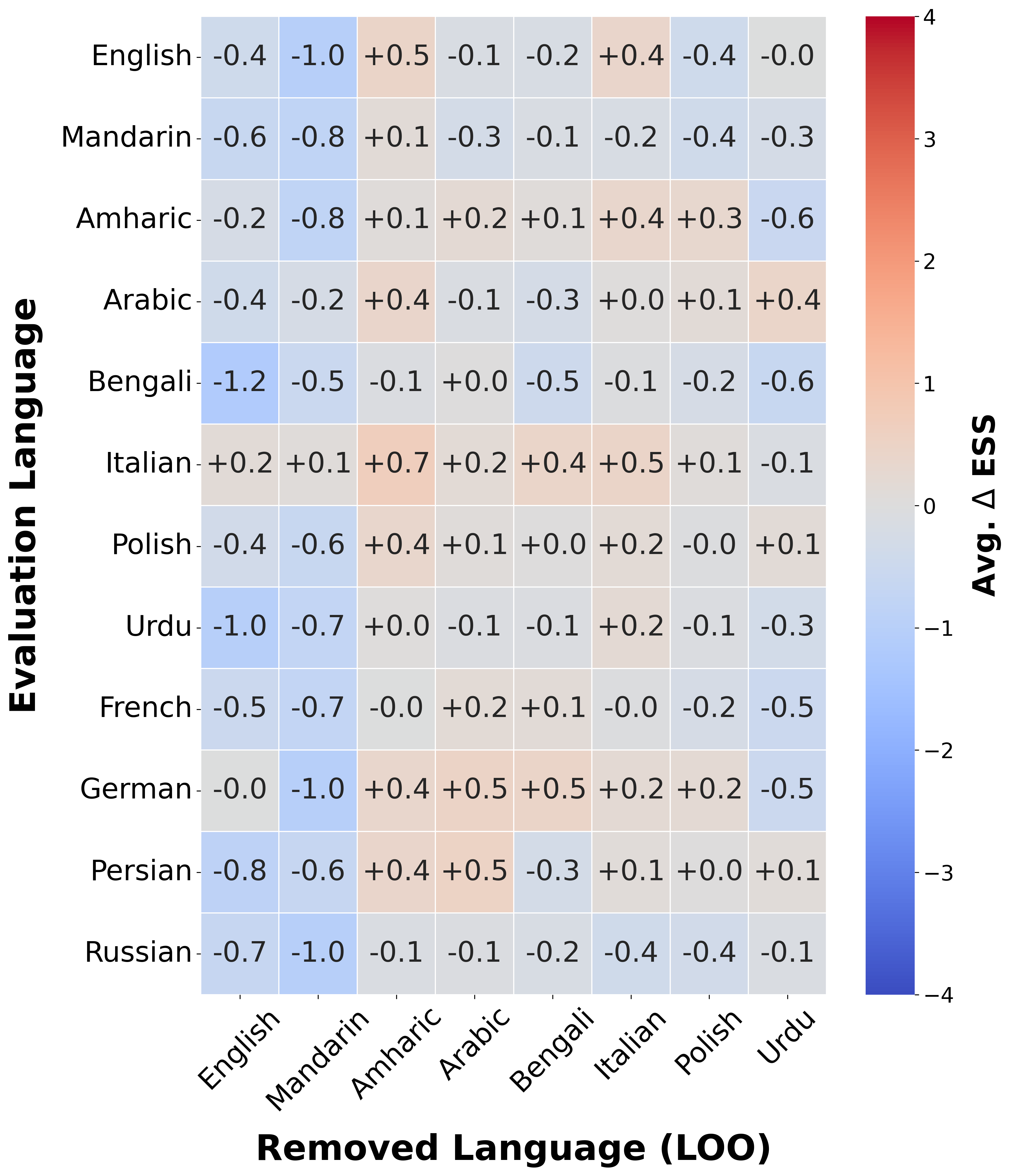}
        \caption{Steering, $\alpha{=}0.5$}
    \end{subfigure}
    \caption{Leave-one-out $\Delta$ESS heatmaps averaged over four models: effect of excluding each identification language (columns) on each evaluation language (rows).}
    \label{fig:ess_dis_loo}
    \vspace{-4mm}
\end{figure}

We use leave-one-out to test whether the multilingual mask is driven by a few dominant identification languages: for each identification language, we remove it from the fusion pool, re-identify the CR-Fusion mask, and measure the change in ESS on each evaluation language (Figure~\ref{fig:ess_dis_loo}). Under deactivation, no single removal explains the gains of CR-Fusion; removals instead produce consistent but non-uniform ESS reductions across evaluation languages, including when low-resource languages are removed. The fused mask thus aggregates partially complementary evidence, with low-resource corpora supplying non-redundant signal for selecting transferable units. Steering shows the same qualitative pattern with smaller and more localized changes. Finally, because each language is represented by a single corpus, language identity and recording condition (elicitation style, channel, speakers) are confounded; we therefore read it as evidence of non-redundant cross-corpus contributions rather than a typological transfer hierarchy.

\section{Conclusion}
\label{sec:conclusoin}
This work presents the first systematic mechanistic investigation of multilingual emotion representations in LALMs. Through causal interventions across 12 typologically diverse languages and four modern LALMs, we showed that monolingually identified emotion neurons share little set-level overlap and that monolingual evidence saturates rapidly, motivating multilingual identification; that CR-Fusion recovers a cross-lingually shared emotion component inaccessible to any single-corpus procedure, yielding more transferable causal control in zero-shot and low-resource settings; and that leave-one-out ablation reveals asymmetric, non-substitutable language contributions, with low-resource corpora both supplying non-redundant identification evidence and being among those that benefit most from fusion.

By demonstrating shared affective representations that generalize across diverse spoken languages, this work provides a mechanistic foundation for equitable affective computing, enabling speech emotion recognition systems that can serve low-resource language communities without requiring extensive language-specific training data.

\newpage

\section*{Limitations}
\label{sec:limitations}
Our language stratification reflects functional roles in the experimental pipeline rather than strict claims about model exposure, as LALMs may have encountered these languages implicitly during web-crawled pre-training. Additionally, our analysis addresses discrete emotion categories; extending to dimensional affect models or fine-grained emotional states remains for future work.

We also admit that each language in our study is represented by a single corpus, so language identity and recording condition (elicitation style, channel, speaker population) are confounded by construction. A post-hoc analysis of the language$\times$emotion interaction component of Eq.~\eqref{eq:decomp} indicates that similarity in language-conditioned activation structure tracks corpus elicitation type (acted vs.\ naturalistic) at least as strongly as typological relatedness; the asymmetric transfer patterns of \S\ref{sec:loo} should therefore be read as corpus-level rather than purely linguistic effects. Disentangling the two requires at least one language represented by corpora of different elicitation types, which we leave to future work. Third, all fusion strategies studied here recover only the language-invariant component of the emotion code; the language-specific component, which accounts for $41$--$70\%$ of emotion-conditioned activation variance across models, is discarded, and exploiting it for target-aware adaptation remains open.

\section*{Ethical Considerations and Broader Societal Impact}
\label{appendix:ethics}
While our work aims to advance mechanistic understanding of emotion processing in multimodal foundation models, we acknowledge potential risks associated with neuron-level emotion manipulation techniques. The intervention methods demonstrated in this study could be misused to artificially induce or suppress emotional signals in generated speech, enabling more persuasive synthetic media or deceptive voice-based systems. Furthermore, fine-grained understanding of ESNs could facilitate targeted attacks on emotion recognition systems or be exploited to bypass affective content moderation. We emphasize that our interventions are evaluated only in controlled research settings on pre-trained models and are not packaged or evaluated as a deployment system; nevertheless, the underlying techniques could be employed in undesirable ways. We hence encourage the research community to develop appropriate safeguards, such as detection mechanisms for artificially manipulated emotional content, alongside continued interpretability research.

\section*{Acknowledgments}
This work was supported by the National Science Foundation (NSF) under CAREER Award IIS-2533652, and in part by the Singapore Ministry of Digital Development and Information under its AIVP Programme (Award Number: AIVP-2026-010).

\bibliography{custom, preprint}

@ARTICLE{11457331,
author={Busso, Carlos and Lotfian, Reza and Sridhar, Kusha and Salman, Ali N. and Lin, Wei-Cheng and Goncalves, Lucas and Parthasarathy, Srinivas and Naini, Abinay Reddy and Leem, Seong-Gyun and Martinez-Lucas, Luz and Chou, Huang-Cheng and Mote, Pravin},
journal={ IEEE Transactions on Affective Computing },
title={{ The MSP-Podcast Corpus }},
year={2026},
volume={1},
ISSN={1949-3045},
pages={1-19},
doi={10.1109/TAFFC.2026.3678489},
url = {https://doi.ieeecomputersociety.org/10.1109/TAFFC.2026.3678489},
publisher={IEEE Computer Society},
address={Los Alamitos, CA, USA},
month=mar}

@inproceedings{wu2024andaudionetworkdissection,
author = {Wu, Tung-Yu and Lin, Yu-Xiang and Weng, Tsui-Wei},
title = {AND: audio network dissection for interpreting deep acoustic models},
year = {2024},
publisher = {JMLR.org},
booktitle = {Proceedings of the 41st International Conference on Machine Learning},
articleno = {2199},
numpages = {25},
location = {Vienna, Austria},
series = {ICML'24}
}

@inproceedings{xu2025deciphering,
  title={Deciphering Functions of Neurons in Vision-Language Models},
  author={Xu, Jiaqi and Lan, Cuiling and Lu, Yan},
  booktitle={Proceedings of the 33rd ACM International Conference on Multimedia},
  pages={3173--3181},
  year={2025}
}

@INPROCEEDINGS {bau2017network,
author = {Bau, David and Zhou, Bolei and Khosla, Aditya and Oliva, Aude and Torralba, Antonio },
booktitle = { 2017 IEEE Conference on Computer Vision and Pattern Recognition (CVPR) },
title = {{ Network Dissection: Quantifying Interpretability of Deep Visual Representations }},
year = {2017},
volume = {},
ISSN = {1063-6919},
pages = {3319-3327},
doi = {10.1109/CVPR.2017.354},
url = {https://doi.ieeecomputersociety.org/10.1109/CVPR.2017.354},
publisher = {IEEE Computer Society},
address = {Los Alamitos, CA, USA},
month =Jul}

@article{bau2020understanding,
   title={Understanding the role of individual units in a deep neural network},
   volume={117},
   ISSN={1091-6490},
   url={http://dx.doi.org/10.1073/pnas.1907375117},
   DOI={10.1073/pnas.1907375117},
   number={48},
   journal={Proceedings of the National Academy of Sciences},
   publisher={Proceedings of the National Academy of Sciences},
   author={Bau, David and Zhu, Jun-Yan and Strobelt, Hendrik and Lapedriza, Agata and Zhou, Bolei and Torralba, Antonio},
   year={2020},
   month=sep, pages={30071–30078}
}

@inproceedings{tang-etal-2024-language,
    title = "Language-Specific Neurons: The Key to Multilingual Capabilities in Large Language Models",
    author = "Tang, Tianyi  and
      Luo, Wenyang  and
      Huang, Haoyang  and
      Zhang, Dongdong  and
      Wang, Xiaolei  and
      Zhao, Xin  and
      Wei, Furu  and
      Wen, Ji-Rong",
    editor = "Ku, Lun-Wei  and
      Martins, Andre  and
      Srikumar, Vivek",
    booktitle = "Proceedings of the 62nd Annual Meeting of the Association for Computational Linguistics (Volume 1: Long Papers)",
    month = aug,
    year = "2024",
    address = "Bangkok, Thailand",
    publisher = "Association for Computational Linguistics",
    url = "https://aclanthology.org/2024.acl-long.309/",
    doi = "10.18653/v1/2024.acl-long.309",
    pages = "5701--5715"
}

@inproceedings{10.1609/aaai.v33i01.33016309,
author = {Dalvi, Fahim and Durrani, Nadir and Sajjad, Hassan and Belinkov, Yonatan and Bau, Anthony and Glass, James},
title = {What is one grain of sand in the desert? analyzing individual neurons in deep NLP models},
year = {2019},
isbn = {978-1-57735-809-1},
publisher = {AAAI Press},
url = {https://doi.org/10.1609/aaai.v33i01.33016309},
doi = {10.1609/aaai.v33i01.33016309},
booktitle = {Proceedings of the Thirty-Third AAAI Conference on Artificial Intelligence and Thirty-First Innovative Applications of Artificial Intelligence Conference and Ninth AAAI Symposium on Educational Advances in Artificial Intelligence},
articleno = {774},
numpages = {9},
location = {Honolulu, Hawaii, USA},
series = {AAAI'19/IAAI'19/EAAI'19}
}

@inproceedings{bau2019identifying,
title={Identifying and Controlling Important Neurons in Neural Machine Translation},
author={Anthony Bau and Yonatan Belinkov and Hassan Sajjad and Nadir Durrani and Fahim Dalvi and James Glass},
booktitle={International Conference on Learning Representations},
year={2019},
url={https://openreview.net/forum?id=H1z-PsR5KX},
}

@inproceedings{voita-etal-2024-neurons,
    title = "Neurons in Large Language Models: Dead, N-gram, Positional",
    author = "Voita, Elena  and
      Ferrando, Javier  and
      Nalmpantis, Christoforos",
    editor = "Ku, Lun-Wei  and
      Martins, Andre  and
      Srikumar, Vivek",
    booktitle = "Findings of the Association for Computational Linguistics: ACL 2024",
    month = aug,
    year = "2024",
    address = "Bangkok, Thailand",
    publisher = "Association for Computational Linguistics",
    url = "https://aclanthology.org/2024.findings-acl.75/",
    doi = "10.18653/v1/2024.findings-acl.75",
    pages = "1288--1301"
}

@article{gurnee2024universal,
title={Universal Neurons in {GPT}2 Language Models},
author={Wes Gurnee and Theo Horsley and Zifan Carl Guo and Tara Rezaei Kheirkhah and Qinyi Sun and Will Hathaway and Neel Nanda and Dimitris Bertsimas},
journal={Transactions on Machine Learning Research},
issn={2835-8856},
year={2024},
url={https://openreview.net/forum?id=ZeI104QZ8I},
note={}
}

@inproceedings{huo-etal-2024-mmneuron,
    title = "{MMN}euron: Discovering Neuron-Level Domain-Specific Interpretation in Multimodal Large Language Model",
    author = "Huo, Jiahao  and
      Yan, Yibo  and
      Hu, Boren  and
      Yue, Yutao  and
      Hu, Xuming",
    editor = "Al-Onaizan, Yaser  and
      Bansal, Mohit  and
      Chen, Yun-Nung",
    booktitle = "Proceedings of the 2024 Conference on Empirical Methods in Natural Language Processing",
    month = nov,
    year = "2024",
    address = "Miami, Florida, USA",
    publisher = "Association for Computational Linguistics",
    url = "https://aclanthology.org/2024.emnlp-main.387/",
    doi = "10.18653/v1/2024.emnlp-main.387",
    pages = "6801--6816"
}

@inproceedings{fang2024towards,
author = {Fang, Junfeng and Bi, Zongze and Wang, Ruipeng and Jiang, Houcheng and Gao, Yuan and Wang, Kun and Zhang, An and Shi, Jie and Wang, Xiang and Chua, Tat-Seng},
title = {Towards neuron attributions in multimodal large language models},
year = {2024},
isbn = {9798331314385},
publisher = {Curran Associates Inc.},
address = {Red Hook, NY, USA},
booktitle = {Proceedings of the 38th International Conference on Neural Information Processing Systems},
articleno = {3904},
numpages = {24},
location = {Vancouver, BC, Canada},
series = {NIPS '24}
}

@inproceedings{yu-ananiadou-2024-neuron,
    title = "Neuron-Level Knowledge Attribution in Large Language Models",
    author = "Yu, Zeping  and
      Ananiadou, Sophia",
    editor = "Al-Onaizan, Yaser  and
      Bansal, Mohit  and
      Chen, Yun-Nung",
    booktitle = "Proceedings of the 2024 Conference on Empirical Methods in Natural Language Processing",
    month = nov,
    year = "2024",
    address = "Miami, Florida, USA",
    publisher = "Association for Computational Linguistics",
    url = "https://aclanthology.org/2024.emnlp-main.191/",
    doi = "10.18653/v1/2024.emnlp-main.191",
    pages = "3267--3280"
}

@inproceedings{zhao-etal-2026-finding,
    title = "Finding Culture-Sensitive Neurons in Vision-Language Models",
    author = "Zhao, Xiutian  and
      Choenni, Rochelle  and
      Saxena, Rohit  and
      Titov, Ivan",
    editor = "Demberg, Vera  and
      Inui, Kentaro  and
      Marquez, Llu{\'i}s",
    booktitle = "Proceedings of the 19th Conference of the {E}uropean Chapter of the {A}ssociation for {C}omputational {L}inguistics (Volume 1: Long Papers)",
    month = mar,
    year = "2026",
    address = "Rabat, Morocco",
    publisher = "Association for Computational Linguistics",
    url = "https://aclanthology.org/2026.eacl-long.155/",
    pages = "3366--3381",
    ISBN = "979-8-89176-380-7"
}

@inproceedings{lee-etal-2025-large,
    title = "Do Large Language Models Have ``Emotion Neurons''? Investigating the Existence and Role",
    author = "Lee, Jaewook  and
      Lee, Woojin  and
      Kwon, Oh-Woog  and
      Kim, Harksoo",
    editor = "Che, Wanxiang  and
      Nabende, Joyce  and
      Shutova, Ekaterina  and
      Pilehvar, Mohammad Taher",
    booktitle = "Findings of the Association for Computational Linguistics: ACL 2025",
    month = jul,
    year = "2025",
    address = "Vienna, Austria",
    publisher = "Association for Computational Linguistics",
    url = "https://aclanthology.org/2025.findings-acl.806/",
    doi = "10.18653/v1/2025.findings-acl.806",
    pages = "15617--15639",
    ISBN = "979-8-89176-256-5"
}

@inproceedings{namazifard-poech-2025-isolating,
    title = "Isolating Culture Neurons in Multilingual Large Language Models",
    author = "Namazifard, Danial  and
      Poech, Lukas Galke",
    editor = "Inui, Kentaro  and
      Sakti, Sakriani  and
      Wang, Haofen  and
      Wong, Derek F.  and
      Bhattacharyya, Pushpak  and
      Banerjee, Biplab  and
      Ekbal, Asif  and
      Chakraborty, Tanmoy  and
      Singh, Dhirendra Pratap",
    booktitle = "Proceedings of the 14th International Joint Conference on Natural Language Processing and the 4th Conference of the Asia-Pacific Chapter of the Association for Computational Linguistics",
    month = dec,
    year = "2025",
    address = "Mumbai, India",
    publisher = "The Asian Federation of Natural Language Processing and The Association for Computational Linguistics",
    url = "https://aclanthology.org/2025.findings-ijcnlp.45/",
    pages = "768--785",
    ISBN = "979-8-89176-303-6"
}

@ARTICLE{Akman2025Improving,
  author={Akman, Alican and Sun, Qiyang and Schuller, Björn W.},
  journal={IEEE Signal Processing Letters}, 
  title={Improving Audio Explanations Using Audio Language Models}, 
  year={2025},
  volume={32},
  number={},
  pages={741-745},
  doi={10.1109/LSP.2025.3532218}}

@inproceedings{singla2022audio,
  title={What do audio transformers hear? probing their representations for language delivery \& structure},
  author={Singla, Yaman Kumar and Shah, Jui and Chen, Changyou and Shah, Rajiv Ratn},
  booktitle={2022 IEEE International Conference on Data Mining Workshops (ICDMW)},
  pages={910--925},
  year={2022},
  organization={IEEE}
}

@inproceedings{skerry2018towards,
  title={Towards end-to-end prosody transfer for expressive speech synthesis with tacotron},
  author={Skerry-Ryan, RJ and Battenberg, Eric and Xiao, Ying and Wang, Yuxuan and Stanton, Daisy and Shor, Joel and Weiss, Ron and Clark, Rob and Saurous, Rif A},
  booktitle={international conference on machine learning},
  pages={4693--4702},
  year={2018},
  organization={PMLR}
}

@inproceedings{wang2018style,
  title={Style tokens: Unsupervised style modeling, control and transfer in end-to-end speech synthesis},
  author={Wang, Yuxuan and Stanton, Daisy and Zhang, Yu and Ryan, RJ-Skerry and Battenberg, Eric and Shor, Joel and Xiao, Ying and Jia, Ye and Ren, Fei and Saurous, Rif A},
  booktitle={International conference on machine learning},
  pages={5180--5189},
  year={2018},
  organization={PMLR}
}

@inproceedings{wu2019end,
  title={End-to-end emotional speech synthesis using style tokens and semi-supervised training},
  author={Wu, Pengfei and Ling, Zhenhua and Liu, Lijuan and Jiang, Yuan and Wu, Hongchuan and Dai, Lirong},
  booktitle={2019 Asia-Pacific Signal and Information Processing Association Annual Summit and Conference (APSIPA ASC)},
  pages={623--627},
  year={2019},
  organization={IEEE}
}

@article{liu2021expressive,
  title={Expressive TTS Training With Frame and Style Reconstruction Loss},
  author={Liu, Rui and Sisman, Berrak and Gao, Guanglai and Li, Haizhou},
  journal={IEEE/ACM Transactions on Audio, Speech, and Language Processing},
  volume={29},
  pages={1806--1818},
  year={2021},
  publisher={IEEE}
}

@inproceedings{lei2021fine,
  title={Fine-grained emotion strength transfer, control and prediction for emotional speech synthesis},
  author={Lei, Yi and Yang, Shan and Xie, Lei},
  booktitle={2021 IEEE Spoken Language Technology Workshop (SLT)},
  pages={423--430},
  year={2021},
  organization={IEEE}
}

@article{lorenzo2018investigating,
  title={Investigating different representations for modeling and controlling multiple emotions in DNN-based speech synthesis},
  author={Lorenzo-Trueba, Jaime and Henter, Gustav Eje and Takaki, Shinji and Yamagishi, Junichi and Morino, Yosuke and Ochiai, Yuta},
  journal={Speech Communication},
  volume={99},
  pages={135--143},
  year={2018},
  publisher={Elsevier}
}

@article{10.1145/3529759,
author = {Retta, Ephrem Afele and Almekhlafi, Eiad and Sutcliffe, Richard and Mhamed, Mustafa and Ali, Haider and Feng, Jun},
title = {A New Amharic Speech Emotion Dataset and Classification Benchmark},
year = {2023},
issue_date = {January 2023},
publisher = {Association for Computing Machinery},
address = {New York, NY, USA},
volume = {22},
number = {1},
issn = {2375-4699},
url = {https://doi.org/10.1145/3529759},
doi = {10.1145/3529759},
journal = {ACM Trans. Asian Low-Resour. Lang. Inf. Process.},
month = may,
articleno = {20},
numpages = {22}
}

@article{SUBESCO,
    doi = {10.1371/journal.pone.0250173},
    author = {Sultana, Sadia AND Rahman, M. Shahidur AND Selim, M. Reza AND Iqbal, M. Zafar},
    journal = {PLOS ONE},
    publisher = {Public Library of Science},
    title = {SUST Bangla Emotional Speech Corpus (SUBESCO): An audio-only emotional speech corpus for Bangla},
    year = {2021},
    month = {04},
    volume = {16},
    url = {https://doi.org/10.1371/journal.pone.0250173},
    pages = {1-27},
    number = {4},

}

@article{UrduSER,
title = {UrduSER: A comprehensive dataset for speech emotion recognition in Urdu language},
journal = {Data in Brief},
volume = {60},
pages = {111627},
year = {2025},
issn = {2352-3409},
doi = {https://doi.org/10.1016/j.dib.2025.111627},
url = {https://www.sciencedirect.com/science/article/pii/S2352340925003580},
author = {Muhammad Zaheer Akhtar and Rashid Jahangir and QuratUl Ain and Muhammad Asif Nauman and Mueen Uddin and Syed Sajid Ullah}
}

@INPROCEEDINGS{BIIC,
  author={Upadhyay, Shreya G. and Chien, Woan-Shiuan and Su, Bo-Hao and Goncalves, Lucas and Wu, Ya-Tse and Salman, Ali N. and Busso, Carlos and Lee, Chi-Chun},
  booktitle={2023 11th International Conference on Affective Computing and Intelligent Interaction (ACII)}, 
  title={An Intelligent Infrastructure Toward Large Scale Naturalistic Affective Speech Corpora Collection}, 
  year={2023},
  volume={},
  number={},
  pages={1-8},
  doi={10.1109/ACII59096.2023.10388175}}

@ARTICLE{Emozionalmente,
  author={Catania, Fabio and Wilke, Jordan W. and Garzotto, Franca},
  journal={IEEE Transactions on Audio, Speech and Language Processing}, 
  title={Emozionalmente: A Crowdsourced Corpus of Simulated Emotional Speech in Italian}, 
  year={2025},
  volume={33},
  number={},
  pages={1142-1155},
  doi={10.1109/TASLPRO.2025.3540662}}

@inproceedings{christop-2024-nemo,
    title = "n{EMO}: Dataset of Emotional Speech in {P}olish",
    author = "Christop, Iwona",
    editor = "Calzolari, Nicoletta  and
      Kan, Min-Yen  and
      Hoste, Veronique  and
      Lenci, Alessandro  and
      Sakti, Sakriani  and
      Xue, Nianwen",
    booktitle = "Proceedings of the 2024 Joint International Conference on Computational Linguistics, Language Resources and Evaluation (LREC-COLING 2024)",
    month = may,
    year = "2024",
    address = "Torino, Italia",
    publisher = "ELRA and ICCL",
    url = "https://aclanthology.org/2024.lrec-main.1059/",
    pages = "12111--12116"
}

@article{mohamad2019shemo,
  title={ShEMO: a large-scale validated database for Persian speech emotion detection},
  author={Mohamad Nezami, Omid and Jamshid Lou, Paria and Karami, Mansoureh},
  journal={Language Resources and Evaluation},
  volume={53},
  number={1},
  pages={1--16},
  year={2019},
  publisher={Springer}
}

@misc{ev21-c430-24,
doi = {10.21227/ev21-c430},
url = {https://dx.doi.org/10.21227/ev21-c430},
author = {Mhamed-amine Soumiaa},
publisher = {IEEE Dataport},
title = {Moroccan Dialect Emotion Recognition Dataset},
year = {2024} }

@misc{aniemore_2023,
	author       = {Amentes, Artem and Davidchuk, Nikita and Lubenets, Ilya},
	title        = { RESD (Revision 75ed61a) },
	year         = 2023,
	url          = { https://huggingface.co/datasets/Aniemore/resd },
	doi          = { 10.57967/hf/1273 },
	publisher    = { Hugging Face }
}

@inproceedings{EmoDB,
  title     = {{EmoDB 2.0: A Database of Emotional Speech in a World that is not Black or White but Grey}},
  author    = {Felix Burkhardt and Oliver Schrüfer and Uwe Reichel and Hagen Wierstorf and Anna Derington and Florian Eyben and Björn W. Schuller},
  year      = {2025},
  booktitle = {{Interspeech 2025}},
  pages     = {4488--4492},
  doi       = {10.21437/Interspeech.2025-1951},
  issn      = {2958-1796},
}

@inproceedings{10.1145/3204949.3208121,
author = {Gournay, Philippe and Lahaie, Olivier and Lefebvre, Roch},
title = {A canadian french emotional speech dataset},
year = {2018},
isbn = {9781450351928},
publisher = {Association for Computing Machinery},
address = {New York, NY, USA},
url = {https://doi.org/10.1145/3204949.3208121},
doi = {10.1145/3204949.3208121},
booktitle = {Proceedings of the 9th ACM Multimedia Systems Conference},
pages = {399–402},
numpages = {4},
location = {Amsterdam, Netherlands},
series = {MMSys '18}
}

@inproceedings{koehn-knowles-2017-six,
    title = "Six Challenges for Neural Machine Translation",
    author = "Koehn, Philipp  and
      Knowles, Rebecca",
    editor = "Luong, Thang  and
      Birch, Alexandra  and
      Neubig, Graham  and
      Finch, Andrew",
    booktitle = "Proceedings of the First Workshop on Neural Machine Translation",
    month = aug,
    year = "2017",
    address = "Vancouver",
    publisher = "Association for Computational Linguistics",
    url = "https://aclanthology.org/W17-3204/",
    doi = "10.18653/v1/W17-3204",
    pages = "28--39"
}

@ARTICLE{10711189,
  author={Mohmad Dar, G. H. and Delhibabu, Radhakrishnan},
  journal={IEEE Access}, 
  title={Speech Databases, Speech Features, and Classifiers in Speech Emotion Recognition: A Review}, 
  year={2024},
  volume={12},
  number={},
  pages={151122-151152},
  doi={10.1109/ACCESS.2024.3476960}}

@inproceedings{maraia-etal-2026-activation,
    title = "Can Activation Steering Generalize Across Languages? A Study on Syllogistic Reasoning in Language Models",
    author = "Maraia, Gabriele  and
      Ranaldi, Leonardo  and
      Valentino, Marco  and
      Zanzotto, Fabio Massimo",
    editor = "Demberg, Vera  and
      Inui, Kentaro  and
      Marquez, Llu{\'i}s",
    booktitle = "Proceedings of the 19th Conference of the {E}uropean Chapter of the {A}ssociation for {C}omputational {L}inguistics (Volume 1: Long Papers)",
    month = mar,
    year = "2026",
    address = "Rabat, Morocco",
    publisher = "Association for Computational Linguistics",
    url = "https://aclanthology.org/2026.eacl-long.125/",
    doi = "10.18653/v1/2026.eacl-long.125",
    pages = "2739--2753",
    ISBN = "979-8-89176-380-7"
}

@inproceedings{singh-etal-2026-lost,
    title = "Lost in Activations: A Neuron-level Analysis of Encoders for Cross-Lingual Emotion Detection",
    author = "Singh, Pranaydeep  and
      De Clercq, Orphee  and
      Lefever, Els",
    editor = "Demberg, Vera  and
      Inui, Kentaro  and
      Marquez, Llu{\'i}s",
    booktitle = "Proceedings of the 19th Conference of the {E}uropean Chapter of the {A}ssociation for {C}omputational {L}inguistics (Volume 2: Short Papers)",
    month = mar,
    year = "2026",
    address = "Rabat, Morocco",
    publisher = "Association for Computational Linguistics",
    url = "https://aclanthology.org/2026.eacl-short.9/",
    doi = "10.18653/v1/2026.eacl-short.9",
    pages = "154--159",
    ISBN = "979-8-89176-381-4"
}

@inproceedings{
huben2024sparse,
title={Sparse Autoencoders Find Highly Interpretable Features in Language Models},
author={Robert Huben and Hoagy Cunningham and Logan Riggs Smith and Aidan Ewart and Lee Sharkey},
booktitle={The Twelfth International Conference on Learning Representations},
year={2024},
url={https://openreview.net/forum?id=F76bwRSLeK}
}

@inproceedings{ma24b_interspeech,
  title     = {{EmoBox: Multilingual Multi-corpus Speech Emotion Recognition Toolkit and Benchmark}},
  author    = {Ziyang Ma and Mingjie Chen and Hezhao Zhang and Zhisheng Zheng and Wenxi Chen and Xiquan Li and Jiaxin Ye and Xie Chen and Thomas Hain},
  year      = {2024},
  booktitle = {{Interspeech 2024}},
  pages     = {1580--1584},
  doi       = {10.21437/Interspeech.2024-788},
  issn      = {2958-1796},
}

@inproceedings{goncalves24_interspeech,
  title     = {{Bridging Emotions Across Languages: Low Rank Adaptation for Multilingual Speech Emotion Recognition}},
  author    = {Lucas Goncalves and Donita Robinson and Elizabeth Richerson and Carlos Busso},
  year      = {2024},
  booktitle = {{Interspeech 2024}},
  pages     = {4688--4692},
  doi       = {10.21437/Interspeech.2024-1226},
  issn      = {2958-1796},
}

@Article{app12189188,
AUTHOR = {Al-onazi, Badriyya B. and Nauman, Muhammad Asif and Jahangir, Rashid and Malik, Muhmmad Mohsin and Alkhammash, Eman H. and Elshewey, Ahmed M.},
TITLE = {Transformer-Based Multilingual Speech Emotion Recognition Using Data Augmentation and Feature Fusion},
JOURNAL = {Applied Sciences},
VOLUME = {12},
YEAR = {2022},
NUMBER = {18},
ARTICLE-NUMBER = {9188},
URL = {https://www.mdpi.com/2076-3417/12/18/9188},
ISSN = {2076-3417},
DOI = {10.3390/app12189188}
}

@INPROCEEDINGS{10889008,
  author={Han, Zhichen and Geng, Tianqi and Feng, Hui and Yuan, Jiahong and Richmond, Korin and Li, Yuanchao},
  booktitle={ICASSP 2025 - 2025 IEEE International Conference on Acoustics, Speech and Signal Processing (ICASSP)}, 
  title={Cross-Lingual Speech Emotion Recognition: Humans vs. Self-Supervised Models}, 
  year={2025},
  volume={},
  number={},
  pages={1-5},
  doi={10.1109/ICASSP49660.2025.10889008}}

@INPROCEEDINGS{8462162,
  author={Neumann, Michael and Thang Vu, N goc},
  booktitle={2018 IEEE International Conference on Acoustics, Speech and Signal Processing (ICASSP)}, 
  title={CRoss-lingual and Multilingual Speech Emotion Recognition on English and French}, 
  year={2018},
  volume={},
  number={},
  pages={5769-5773},
  doi={10.1109/ICASSP.2018.8462162}}

@INPROCEEDINGS{11209040,
  author={Zou, Heqing and Lv, Fengmao and Zheng, Desheng and Chng, Eng Siong and Rajan, Deepu},
  booktitle={2025 IEEE International Conference on Multimedia and Expo (ICME)}, 
  title={Large Language Models Meet Contrastive Learning: Zero-Shot Emotion Recognition Across Languages}, 
  year={2025},
  volume={},
  number={},
  pages={1-6},
  doi={10.1109/ICME59968.2025.11209040}}

@INPROCEEDINGS{9747417,
  author={Sharma, Mayank},
  booktitle={ICASSP 2022 - 2022 IEEE International Conference on Acoustics, Speech and Signal Processing (ICASSP)}, 
  title={Multi-Lingual Multi-Task Speech Emotion Recognition Using wav2vec 2.0}, 
  year={2022},
  volume={},
  number={},
  pages={6907-6911},
  doi={10.1109/ICASSP43922.2022.9747417}}

@article{Xu_Chen_Yu_Huang_Wu_Zhang_Li_Luo_Gu_2024, title={SECap: Speech Emotion Captioning with Large Language Model}, volume={38}, url={https://ojs.aaai.org/index.php/AAAI/article/view/29902}, DOI={10.1609/aaai.v38i17.29902}, abstractNote={Speech emotions are crucial in human communication and are extensively used in fields like speech synthesis and natural language understanding. Most prior studies, such as speech emotion recognition, have categorized speech emotions into a fixed set of classes. Yet, emotions expressed in human speech are often complex, and categorizing them into predefined groups can be insufficient to adequately represent speech emotions. On the contrary, describing speech emotions directly by means of natural language may be a more effective approach. Regrettably, there are not many studies available that have focused on this direction. Therefore, this paper proposes a speech emotion captioning framework named SECap, aiming at effectively describing speech emotions using natural language. Owing to the impressive capabilities of large language models in language comprehension and text generation, SECap employs LLaMA as the text decoder to allow the production of coherent speech emotion captions. In addition, SECap leverages HuBERT as the audio encoder to extract general speech features and Q-Former as the Bridge-Net to provide LLaMA with emotion-related speech features. To accomplish this, Q-Former utilizes mutual information learning to disentangle emotion-related speech features and speech contents, while implementing contrastive learning to extract more emotion-related speech features. The results of objective and subjective evaluations demonstrate that: 1) the SECap framework outperforms the HTSAT-BART baseline in all objective evaluations; 2) SECap can generate high-quality speech emotion captions that attain performance on par with human annotators in subjective mean opinion score tests.}, number={17}, journal={Proceedings of the AAAI Conference on Artificial Intelligence}, author={Xu, Yaoxun and Chen, Hangting and Yu, Jianwei and Huang, Qiaochu and Wu, Zhiyong and Zhang, Shi-Xiong and Li, Guangzhi and Luo, Yi and Gu, Rongzhi}, year={2024}, month={Mar.}, pages={19323-19331} }

@ARTICLE{7337399,
  author={Albornoz, E.M. and Milone, D.H.},
  journal={IEEE Transactions on Affective Computing}, 
  title={Emotion recognition in never-seen languages using a novel ensemble method with emotion profiles}, 
  year={2017},
  volume={8},
  number={1},
  pages={43-53},
  doi={10.1109/TAFFC.2015.2503757}}

@article{zehra2021cross,
  title={Cross corpus multi-lingual speech emotion recognition using ensemble learning},
  author={Zehra, Wisha and Javed, Abdul Rehman and Jalil, Zunera and Khan, Habib Ullah and Gadekallu, Thippa Reddy},
  journal={Complex \& Intelligent Systems},
  volume={7},
  number={4},
  pages={1845--1854},
  year={2021},
  publisher={Springer}
}

@book{wilce2009language, 
place={Cambridge}, 
series={Studies in the Social and Cultural Foundations of Language}, 
title={Language and Emotion}, 
publisher={Cambridge University Press}, 
author={Wilce, James M.}, 
year={2009}, 
collection={Studies in the Social and Cultural Foundations of Language}
}

@article{lindquist2015role,
  title={The role of language in emotion: Predictions from psychological constructionism},
  author={Lindquist, Kristen A and MacCormack, Jennifer K and Shablack, Holly},
  journal={Frontiers in psychology},
  volume={6},
  pages={121301},
  year={2015},
  publisher={Frontiers}
}

@article{pell2009recognizing,
  title={Recognizing emotions in a foreign language},
  author={Pell, Marc D and Monetta, Laura and Paulmann, Silke and Kotz, Sonja A},
  journal={Journal of Nonverbal Behavior},
  volume={33},
  number={2},
  pages={107--120},
  year={2009},
  publisher={Springer}
}

@article{russell1991culture,
  title={Culture and the categorization of emotions.},
  author={Russell, James A},
  journal={Psychological bulletin},
  volume={110},
  number={3},
  pages={426},
  year={1991},
  publisher={American Psychological Association}
}

@article{
doi:10.1126/science.aaw8160,
author = {Joshua Conrad Jackson  and Joseph Watts  and Teague R. Henry  and Johann-Mattis List  and Robert Forkel  and Peter J. Mucha  and Simon J. Greenhill  and Russell D. Gray  and Kristen A. Lindquist },
title = {Emotion semantics show both cultural variation and universal structure},
journal = {Science},
volume = {366},
number = {6472},
pages = {1517-1522},
year = {2019},
doi = {10.1126/science.aaw8160},
URL = {https://www.science.org/doi/abs/10.1126/science.aaw8160},
eprint = {https://www.science.org/doi/pdf/10.1126/science.aaw8160}}

@inproceedings{
sakshi2025mmau,
title={{MMAU}: A Massive Multi-Task Audio Understanding and Reasoning Benchmark},
author={S Sakshi and Utkarsh Tyagi and Sonal Kumar and Ashish Seth and Ramaneswaran Selvakumar and Oriol Nieto and Ramani Duraiswami and Sreyan Ghosh and Dinesh Manocha},
booktitle={The Thirteenth International Conference on Learning Representations},
year={2025},
url={https://openreview.net/forum?id=TeVAZXr3yv}
}

@INPROCEEDINGS{10448257,
  author={Huang, Chien-Yu and Lu, Ke-Han and Wang, Shih-Heng and Hsiao, Chi-Yuan and Kuan, Chun-Yi and Wu, Haibin and Arora, Siddhant and Chang, Kai-Wei and Shi, Jiatong and Peng, Yifan and Sharma, Roshan and Watanabe, Shinji and Ramakrishnan, Bhiksha and Shehata, Shady and Lee, Hung-Yi},
  booktitle={ICASSP 2024 - 2024 IEEE International Conference on Acoustics, Speech and Signal Processing (ICASSP)}, 
  title={Dynamic-Superb: Towards a Dynamic, Collaborative, and Comprehensive Instruction-Tuning Benchmark For Speech}, 
  year={2024},
  volume={},
  number={},
  pages={12136-12140},
  doi={10.1109/ICASSP48485.2024.10448257}}

@inproceedings{
cheng2024emotionllama,
title={Emotion-{LL}a{MA}: Multimodal Emotion Recognition and Reasoning with Instruction Tuning},
author={Zebang Cheng and Zhi-Qi Cheng and Jun-Yan He and Kai Wang and Yuxiang Lin and Zheng Lian and Xiaojiang Peng and Alexander G Hauptmann},
booktitle={The Thirty-eighth Annual Conference on Neural Information Processing Systems},
year={2024},
url={https://openreview.net/forum?id=qXZVSy9LFR}
}

@inproceedings{zhao-etal-2026-discovering,
    title = "Discovering and Causally Validating Emotion-Sensitive Neurons in Large Audio-Language Models",
    author = {Zhao, Xiutian  and
      Schuller, Bj{\"o}rn  and
      Sisman, Berrak},
    editor = "Liakata, Maria  and
      Moreira, Viviane P.  and
      Zhang, Jiajun  and
      Jurgens, David",
    booktitle = "Proceedings of the 64th Annual Meeting of the {A}ssociation for {C}omputational {L}inguistics (Volume 1: Long Papers)",
    month = jul,
    year = "2026",
    address = "San Diego, California, United States",
    publisher = "Association for Computational Linguistics",
    url = "https://aclanthology.org/2026.acl-long.687/",
    doi = "10.18653/v1/2026.acl-long.687",
    pages = "15056--15071",
    ISBN = "979-8-89176-390-6"
}

@INPROCEEDINGS{10447060,
  author={Ulgen, Ismail Rasim and Du, Zongyang and Busso, Carlos and Sisman, Berrak},
  booktitle={ICASSP 2024 - 2024 IEEE International Conference on Acoustics, Speech and Signal Processing (ICASSP)}, 
  title={Revealing Emotional Clusters in Speaker Embeddings: A Contrastive Learning Strategy for Speech Emotion Recognition}, 
  year={2024},
  volume={},
  number={},
  pages={12081-12085},
  doi={10.1109/ICASSP48485.2024.10447060}}

@misc{xu2025qwen25omnitechnicalreport,
      title={Qwen2.5-Omni Technical Report}, 
      author={Jin Xu and Zhifang Guo and Jinzheng He and Hangrui Hu and Ting He and Shuai Bai and Keqin Chen and Jialin Wang and Yang Fan and Kai Dang and Bin Zhang and Xiong Wang and Yunfei Chu and Junyang Lin},
      year={2025},
      eprint={2503.20215},
      archivePrefix={arXiv},
      primaryClass={cs.CL},
      url={https://arxiv.org/abs/2503.20215}, 
}

@misc{yao2024minicpmvgpt4vlevelmllm,
      title={MiniCPM-V: A GPT-4V Level MLLM on Your Phone}, 
      author={Yuan Yao and Tianyu Yu and Ao Zhang and Chongyi Wang and Junbo Cui and Hongji Zhu and Tianchi Cai and Haoyu Li and Weilin Zhao and Zhihui He and Qianyu Chen and Huarong Zhou and Zhensheng Zou and Haoye Zhang and Shengding Hu and Zhi Zheng and Jie Zhou and Jie Cai and Xu Han and Guoyang Zeng and Dahai Li and Zhiyuan Liu and Maosong Sun},
      year={2024},
      eprint={2408.01800},
      archivePrefix={arXiv},
      primaryClass={cs.CV},
      url={https://arxiv.org/abs/2408.01800}, 
}

@misc{kimiteam2025kimiaudiotechnicalreport,
      title={Kimi-Audio Technical Report}, 
      author={KimiTeam},
      year={2025},
      eprint={2504.18425},
      archivePrefix={arXiv},
      primaryClass={eess.AS},
      url={https://arxiv.org/abs/2504.18425}, 
}

@misc{goel2025audioflamingo3advancing,
      title={Audio Flamingo 3: Advancing Audio Intelligence with Fully Open Large Audio Language Models}, 
      author={Arushi Goel and Sreyan Ghosh and Jaehyeon Kim and Sonal Kumar and Zhifeng Kong and Sang-gil Lee and Chao-Han Huck Yang and Ramani Duraiswami and Dinesh Manocha and Rafael Valle and Bryan Catanzaro},
      year={2025},
      eprint={2507.08128},
      archivePrefix={arXiv},
      primaryClass={cs.SD},
      url={https://arxiv.org/abs/2507.08128}, 
}

@misc{shazeer2020gluvariantsimprovetransformer,
      title={GLU Variants Improve Transformer}, 
      author={Noam Shazeer},
      year={2020},
      eprint={2002.05202},
      archivePrefix={arXiv},
      primaryClass={cs.LG},
      url={https://arxiv.org/abs/2002.05202}, 
}

@misc{singh2023decodingemotionscomprehensivemultilingual,
      title={Decoding Emotions: A comprehensive Multilingual Study of Speech Models for Speech Emotion Recognition}, 
      author={Anant Singh and Akshat Gupta},
      year={2023},
      eprint={2308.08713},
      archivePrefix={arXiv},
      primaryClass={cs.CL},
      url={https://arxiv.org/abs/2308.08713}, 
}

@misc{huang2024minerminingunderlyingpattern,
      title={MINER: Mining the Underlying Pattern of Modality-Specific Neurons in Multimodal Large Language Models}, 
      author={Kaichen Huang and Jiahao Huo and Yibo Yan and Kun Wang and Yutao Yue and Xuming Hu},
      year={2024},
      eprint={2410.04819},
      archivePrefix={arXiv},
      primaryClass={cs.CL},
      url={https://arxiv.org/abs/2410.04819}, 
}

@misc{he2025meralionaudiollmbridgingaudiolanguage,
      title={MERaLiON-AudioLLM: Bridging Audio and Language with Large Language Models}, 
      author={Yingxu He and Zhuohan Liu and Shuo Sun and Bin Wang and Wenyu Zhang and Xunlong Zou and Nancy F. Chen and Ai Ti Aw},
      year={2025},
      eprint={2412.09818},
      archivePrefix={arXiv},
      primaryClass={cs.CL},
      url={https://arxiv.org/abs/2412.09818}, 
}

@misc{yang2025audiolenscloserlookauditory,
      title={AudioLens: A Closer Look at Auditory Attribute Perception of Large Audio-Language Models}, 
      author={Chih-Kai Yang and Neo Ho and Yi-Jyun Lee and Hung-yi Lee},
      year={2025},
      eprint={2506.05140},
      archivePrefix={arXiv},
      primaryClass={cs.CL},
      url={https://arxiv.org/abs/2506.05140}, 
}

@misc{xie2025emosteerttsfinegrainedtrainingfreeemotioncontrollable,
      title={EmoSteer-TTS: Fine-Grained and Training-Free Emotion-Controllable Text-to-Speech via Activation Steering}, 
      author={Tianxin Xie and Shan Yang and Chenxing Li and Dong Yu and Li Liu},
      year={2025},
      eprint={2508.03543},
      archivePrefix={arXiv},
      primaryClass={cs.SD},
      url={https://arxiv.org/abs/2508.03543}, 
}

@misc{wang2025llmsfeelemotioncircuits,
      title={Do LLMs "Feel"? Emotion Circuits Discovery and Control}, 
      author={Chenxi Wang and Yixuan Zhang and Ruiji Yu and Yufei Zheng and Lang Gao and Zirui Song and Zixiang Xu and Gus Xia and Huishuai Zhang and Dongyan Zhao and Xiuying Chen},
      year={2025},
      eprint={2510.11328},
      archivePrefix={arXiv},
      primaryClass={cs.CL},
      url={https://arxiv.org/abs/2510.11328}, 
}

@misc{shou2025multimodallargelanguagemodels,
      title={Multimodal Large Language Models Meet Multimodal Emotion Recognition and Reasoning: A Survey}, 
      author={Yuntao Shou and Tao Meng and Wei Ai and Keqin Li},
      year={2025},
      eprint={2509.24322},
      archivePrefix={arXiv},
      primaryClass={cs.CL},
      url={https://arxiv.org/abs/2509.24322}, 
}

@misc{zhao2026neuronlevelemotioncontrolspeechgenerative,
      title={Neuron-Level Emotion Control in Speech-Generative Large Audio-Language Models}, 
      author={Xiutian Zhao and Ismail Rasim Ulgen and Philipp Koehn and Björn Schuller and Berrak Sisman},
      year={2026},
      eprint={2603.17231},
      archivePrefix={arXiv},
      primaryClass={cs.CL},
      url={https://arxiv.org/abs/2603.17231}, 
}

@misc{sofroniew2026emotionconceptsfunctionlarge,
      title={Emotion Concepts and their Function in a Large Language Model}, 
      author={Nicholas Sofroniew and Isaac Kauvar and William Saunders and Runjin Chen and Tom Henighan and Sasha Hydrie and Craig Citro and Adam Pearce and Julius Tarng and Wes Gurnee and Joshua Batson and Sam Zimmerman and Kelley Rivoire and Kyle Fish and Chris Olah and Jack Lindsey},
      year={2026},
      eprint={2604.07729},
      archivePrefix={arXiv},
      primaryClass={cs.AI},
      url={https://arxiv.org/abs/2604.07729}, 
}

\appendix

\section{Reproducibility}
\label{appendix:reproducibility}

\subsection{Computational Resources}
\label{appendix:computation}
All experiments were conducted using NVIDIA A100 and H100 GPUs from a shared cluster budget with 8 GPUs of each type. The total GPU consumption, including failed runs and debugging, was approximately 312 A100-hours and 687 H100-hours. The majority of compute time was dedicated to intervention experiments across 12 languages, four LALMs and multiple parameter settings, with activation logging  requiring comparatively fewer resources. Each LALM required between 40-80 GB of GPU memory depending on model size and batch configuration. We estimate that reproducing only the final reported experiments would require approximately 400 A100-hours or equivalent.

\subsection{Datasets and Models}
\label{app:data}

\begin{table*}[h]
\resizebox{\textwidth}{!}{%
\begin{tabular}{@{}lcccccccccccc@{}}
\toprule
Group    & \multicolumn{8}{c}{$\mathcal{L}_\text{id}$}                                                                                         & \multicolumn{4}{c}{$\mathcal{L}_\text{held}$} \\ \cmidrule(l){2-9} \cmidrule(l){10-13}
Language & Amharic & Arabic & Bengali & English     & Italian                                                    & Mandarin & Polish & Urdu    & French    & German    & Persian   & Russian   \\ \cmidrule(l){2-13}
Datasets & ASED    & MDER   & SUBESCO & MSP-Podcast & \begin{tabular}[c]{@{}c@{}}Emozio\\ -nalmente\end{tabular} & BIIC     & nEMO   & UrduSER & CaFE      & EmoDB     & ShEMO     & RESD      \\ \midrule
Angry    & 486     & 400    & 1000    & 36260       & 986                                                        & 6543     & 749    & 500     & 144       & 137       & 1059      & 219       \\
Fear     & 510     & 400    & 1000    & 1943        & 986                                                        & 898      & 736    & 500     & 144       & 122       & 38        & 223       \\
Happy    & 486     & 400    & 1000    & 58684       & 986                                                        & 18859    & 749    & 500     & 144       & 115       & 201       & 218       \\
Neutral  & 522     & 400    & 1000    & 79117       & 986                                                        & 21269    & 809    & 500     & 72        & 104       & 1028      & 191       \\
Sad      & 470     & 400    & 1000    & 24629       & 986                                                        & 5679     & 769    & 500     & 144       & 121       & 449       & 162       \\ \bottomrule
\end{tabular}%
}
\caption{Dataset statistics showing utterance counts per emotion across 12 multilingual speech emotion languages/datasets, grouped by experimental role: Identification and Held. 
}
\label{tab:datasets}
\end{table*}

For each dataset, we sample up to 150 utterances per emotion for the evaluation partition using a fixed random seed to ensure reproducibility; when a dataset contains fewer than 150 utterances for an emotion (e.g., CaFE, EmoDB, ShEMO; Table~\ref{tab:datasets}), all available utterances are used.

\begin{table}[H]
\centering
\resizebox{\columnwidth}{!}{%
\begin{tabular}{@{}lll@{}}
\toprule
\textbf{Models} & \textbf{Hugging Face Identifier}                                            & \textbf{License}                \\ \midrule
Audio-Flamingo-3& \url{https://huggingface.co/nvidia/audio-flamingo-3}           & 
NVIDIA OneWay Noncommercial License\\ 
Kimi-Audio      & \url{https://huggingface.co/moonshotai/Kimi-Audio-7B-Instruct} & Apache License 2.0, MIT License \\ 
MiniCPM-o-4.5   & \url{https://huggingface.co/openbmb/MiniCPM-o-4\_5}            & Apache License 2.0              \\
Qwen2.5-Omni-7B & \url{https://huggingface.co/Qwen/Qwen2.5-Omni-7B}              & Apache License 2.0              \\ \bottomrule
\end{tabular}%
}
\caption{
Sources and licenses for the four evaluated LALMs.
}
\label{tab:models}
\end{table}

\subsection{SER Prompt Template}
\label{appendix:prompt}
To reduce intrinsic positional and label preference biases within LALMs, we randomize the index$\leftrightarrow$emotion mapping per instance: for each speech clip instance, each emotion is randomly assigned to an option letter (e.g., ``A''). We implement the following prompt template for SER.

\vspace{4mm}
\resizebox{0.8\columnwidth}{!}{%
\begin{promptbox}
Based on the provided speech clip, identify the emotion expressed in the speech.\\
Choose the option that best matches the perceived emotion from the audio:\\

A: {emotion 1}\\
B: {emotion 2}\\
C: {emotion 3}\\
D: {emotion 4}\\
E: {emotion 5}\\

Output exactly one option letter and no other text.
\end{promptbox}
}
\subsection{Decoding and Statistical Reporting}
\label{appendix:stats}
We decode deterministically (greedy; temperature 0) with a 20-token generation limit and apply lightweight post-processing to extract the option letter from model outputs.
Deterministic decoding ensures that model outputs are as reproducible as possible given fixed inputs and model weights. Consequently, repeated runs on identical data produce identical results, and variance in reported metrics reflects cross-language or cross-condition variability rather than stochastic sampling noise. In Tables~\ref{tab:deact_results} and \ref{tab:steer_results}, we report mean and standard deviation (in parentheses) computed across evaluation languages within each group, quantifying the consistency of intervention effects across typologically diverse linguistic contexts. For sensitivity analyses (Figure~\ref{fig:lambda}), shaded bands represent $\pm1$ SEM computed over all 12 evaluation languages ($n{=}12$). Data partitioning uses fixed random seeds (specified in our released code) to ensure exact reproducibility of all reported results.

We do not report seed-based confidence intervals, as deterministic decoding renders repeated runs identical and seed variance is exactly zero. Bootstrap resampling over test utterances would quantify sampling variability of the evaluation sets; we instead report cross-language dispersion, which is the variability most relevant to our claims about multilingual generalization.

All standard deviations and standard errors are computed using closed-form formulas implemented via NumPy \footnote{https://numpy.org/} library functions.

\subsection{Identification Instance Budget and Data Availability Constraints}
\label{app:id_budget}

Our neuron identification procedure targets $c{=}50$ correctly predicted instances per emotion per language. However, certain model-language-emotion triples exhibit low baseline accuracy, yielding fewer correct predictions available for activation logging. In such cases, we use all available correctly predicted instances rather than discarding the emotion or language entirely. 

We retain low-count conditions for two reasons. First, excluding underperforming conditions would systematically bias our analysis toward high-resource, well-recognized settings, undermining our goal of investigating multilingual emotion representations. Second, our saturation analysis (\S\ref{sec:saturation}) demonstrates that intervention effects plateau beyond approximately 50 instances, indicating that reduced counts can still provide informative activation statistics. 

For severely underrepresented cases such as MiniCPM-o-4.5 Fear (36 total instances, with zero for Arabic and Polish), we acknowledge reduced reliability. However, this primarily reflects MiniCPM-o-4.5's strong bias toward Neutral predictions (98.83\% baseline accuracy). The emotion-level analysis (\S\ref{sec:heterogeneity}) independently confirms that Fear exhibits weaker causal effects, consistent with data sparsity. Our primary claims rest on emotions with robust identification data across all models, namely anger, happiness, and sadness.

\subsection{Per-Emotion Detail}
\label{app:emotion_detail}

 \begin{figure}[ht!]
    \centering
    \begin{subfigure}[b]{0.5\columnwidth}
        \centering
        \includegraphics[width=\linewidth]{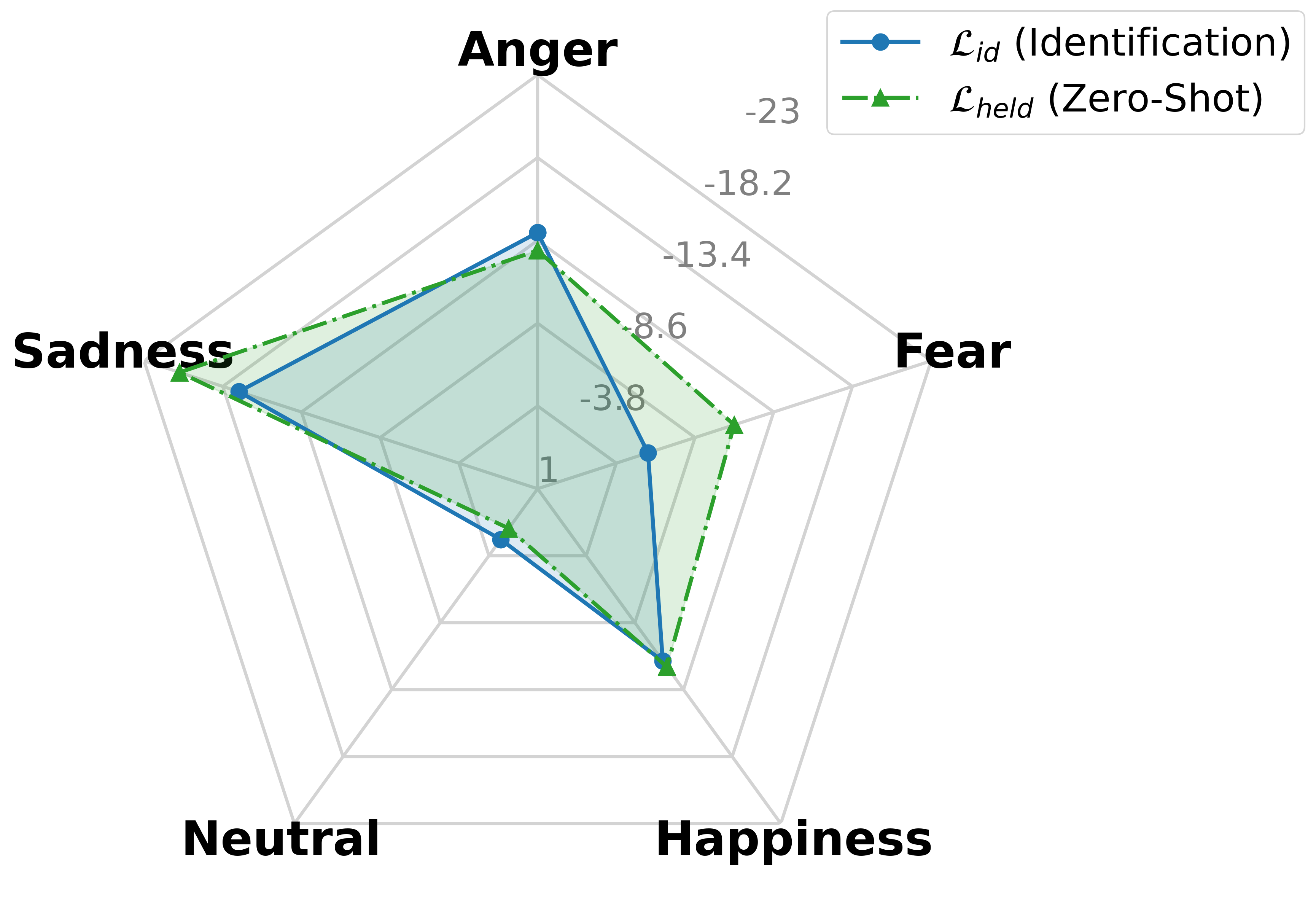}
        \caption{Deactivation}
    \end{subfigure}\hfill
    \begin{subfigure}[b]{0.5\columnwidth}
        \centering
        \includegraphics[width=\linewidth]{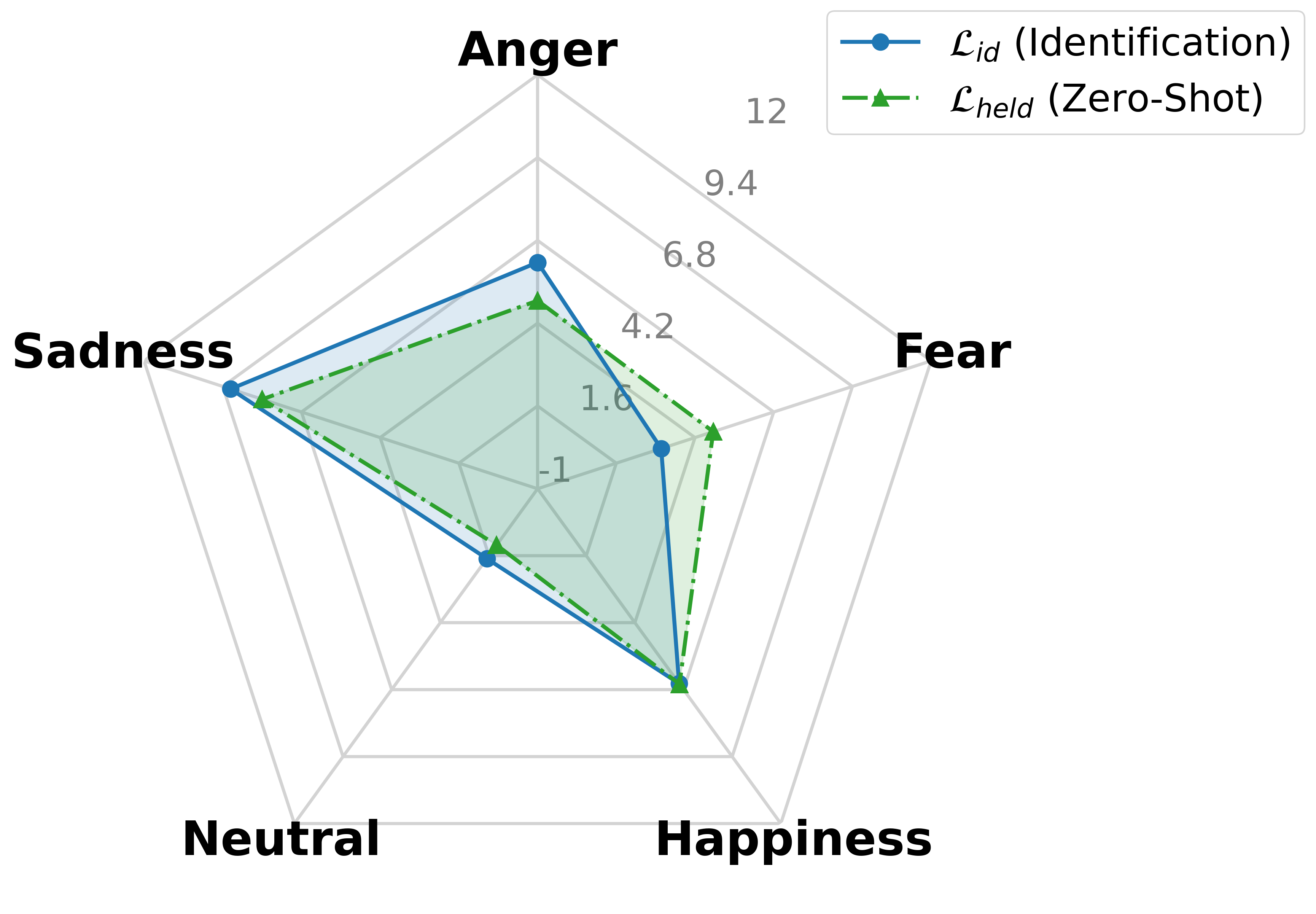}
        \caption{Steering, $\alpha{=}0.5$}
    \end{subfigure}
    \caption{Per-emotion ESS magnitudes under (a) deactivation and (b) steering, comparing CR-Fusion ($\lambda$=0.3) against monolingual baselines, reported separately for $\mathcal{L}_\text{id}$ and $\mathcal{L}_\text{held}$. Radial axes are not shared across panels.}
    \label{fig:emotion_radar}
\end{figure}

\paragraph{Group consistency.}
Figure~\ref{fig:emotion_radar} reports per-emotion ESS magnitudes separately for $\mathcal{L}_\text{id}$ and $\mathcal{L}_\text{held}$. The relative ordering of emotions is preserved across groups for all models, under both deactivation and steering: the emotions with the strongest effects on identification languages are also the strongest on languages never seen during identification. Radial axes are not shared across panels.

\paragraph{Causal potency versus representational invariance.} It is tempting to interprete the ordering in \S\ref{sec:heterogeneity} as evidence that anger, happiness, and sadness are encoded in more language-invariant form than fear or neutral. However, the variance decomposition of \S\ref{sec:theory} does not support that interpretation: per-emotion invariant shares (Table~\ref{tab:decomp_emotion}) are not highest for the three emotions with the strongest causal effects in any model, and averaged across models the ordering is inverted: neutral ($0.52$) and fear ($0.50$) exceed anger ($0.43$), sadness ($0.43$), and happiness ($0.41$).
 
The two quantities measure different things, and the dissociation is informative. The invariant share is a property of the representation: how much of a neuron's emotion tuning is shared across languages. ESS is a property of model behavior under intervention, bounded by baseline recognition accuracy, per-emotion test support, and confusability within the response set. Fear is the least or second-least accurately recognized emotion in every model, and neutral, although nominally the most accurate in three of four models, is inflated by a strong neutral response bias (Appendix~\ref{app:baseline}): its high accuracy reflects a default prediction rather than discriminative use of emotion-sensitive units, so suppressing neutral-selective units cannot remove it. Anger and sadness are recognized well above the fear floor in every model. Baseline accuracy, response bias and per-emotion test support therefore suffice to produce the observed causal ordering without any difference in representational invariance.

\section{Variance Decomposition of Multilingual Emotion Coding}
\label{app:decomp}
 
\paragraph{Estimation.}
We fit Eq.~\eqref{eq:decomp} per neuron by alternating weighted centering over the language and emotion axes of the $|L| \times |E|$ activation-probability table, after dividing each language's tensor by its global standard deviation to remove differences in dynamic range. For complete tables the fit converges in one iteration and coincides with the two-way ANOVA solution; for tables with missing cells (emotions for which a model produced too few correct predictions in a given language, Appendix~\ref{app:id_budget}) it converges to the weighted least-squares additive fit. All quantities below are computed over the top-$r{=}0.5\%$ of neurons ranked by the ConAct margin of the pooled profile, i.e.\ the population from which masks are drawn.

\paragraph{Noise correction.}
Since $\hat P$ is a binomial proportion, each cell carries sampling variance $v = P(1-P)/T^{(\ell,e)}$, which inflates the apparent interaction variance far more than the invariant variance: noise enters the residual $\hat \gamma$ with weight $(|L|{-}1)(|E|{-}1)/(|L||E|)$ but enters $\hat b$ with weight only $(|E|{-}1)/(|L||E|)$, a factor $|L|{-}1$ smaller, because averaging over languages suppresses it. We therefore report shares based on $\hat\sigma_\gamma^2 = \mathrm{ms}(\hat \gamma)/f_\gamma - \bar v$ and $\hat\sigma_b^2 = \big(\mathrm{ms}(\hat b) - \bar v f_b/|L|\big)/f_b$, where $\mathrm{ms}(\cdot)$ is the mean square, $f_\gamma = (|L|{-}1)(|E|{-}1)/(|L||E|)$ and $f_b = (|E|{-}1)/|E|$ are residual projection factors, and $\bar v$ is the mean estimated binomial variance. Uncorrected shares are within $0.03$ of corrected values for every model, and estimated noise accounts for only $13$--$22\%$ of the raw interaction energy, so the language-specific component is not a sampling artifact.
 
\begin{table}[t]
\centering
\small
\setlength{\tabcolsep}{4pt}
\begin{tabular}{lccccc}
\toprule
LALM & $\hat\sigma_b^2$ & $\hat\sigma_\gamma^2$ & Share & Share & JSC \\
     & $(\times 10^{3})$ & $(\times 10^{3})$ & (raw) & (corr.) & \\
\midrule
Audio-Flamingo-3 & 3.57 & 4.70 & 0.44 & 0.43 & 0.98 \\
Kimi-Audio       & 2.47 & 2.27 & 0.50 & 0.52 & 0.98 \\
MiniCPM-o-4.5    & 2.10 & 1.45 & 0.59 & 0.59 & 0.41 \\
Qwen2.5-Omni-7B  & 1.14 & 2.61 & 0.31 & 0.30 & 0.63 \\
\bottomrule
\end{tabular}
\caption{Variance components of Eq.~\eqref{eq:decomp} among top-ranked emotion-selective neurons ($r{=}0.5\%$, ConAct, $c{=}50$). ``Share'' is the invariant fraction $\sigma_b^2/(\sigma_b^2+\sigma_\gamma^2)$, raw and noise-corrected. ``JSC'' is the mean Jaccard similarity between masks selected on the fitted invariant component $m+\hat b$ and Joint Fusion masks: near-identity ($0.98$) for the two models with complete language--emotion tables, as predicted by the algebraic equivalence of \S\ref{sec:theory}, with divergence appearing only where missing cells make the weighted fit differ from naive pooling.}
\label{tab:decomp_model}
\end{table}
 
\begin{table}[t]
\centering
\small
\setlength{\tabcolsep}{4.5pt}
\begin{tabular}{lccccc}
\toprule
LALM & Ang. & Fear & Hap. & Neu. & Sad. \\
\midrule
Audio-Flamingo-3 & 0.39 & 0.42 & 0.42 & 0.50 & 0.42 \\
Kimi-Audio       & 0.54 & 0.50 & 0.44 & 0.61 & 0.48 \\
MiniCPM-o-4.5    & 0.57 & 0.72 & 0.49 & 0.60 & 0.52 \\
Qwen2.5-Omni-7B  & 0.23 & 0.35 & 0.30 & 0.34 & 0.29 \\
\midrule
Mean             & 0.43 & 0.50 & 0.41 & 0.52 & 0.43 \\
\bottomrule
\end{tabular}
\caption{Per-emotion noise-corrected invariant share. The share is \emph{not} highest for anger, happiness, and sadness in any model; averaged across models, neutral and fear show the highest invariant shares. Representational universality therefore does not explain the causal-potency ordering of \S\ref{sec:heterogeneity} (Appendix~\ref{app:emotion_detail}).} 
\label{tab:decomp_emotion}
\end{table}
 
\paragraph{Interpretation.}
From Table~\ref{tab:decomp_model} and \ref{tab:decomp_emotion}, we observe that: (1) the invariant share is below $0.6$ across models; in two of four it is below one half. The language-invariant emotion code that fusion isolates is real and causally potent (\S\ref{sec:results}), but it is a minority-to-half share of the emotion-conditioned activation structure; (2) The masks selected on the fitted invariant component coincide with Joint Fusion masks wherever the design is complete (JSC $0.98$), confirming empirically that Joint Fusion and small-$\lambda$ CR-Fusion estimate the invariant component rather than performing consensus filtering; (3) With only four models we do not treat the invariant share as a predictor of fusion gain, and the two orderings are not monotonically related: MiniCPM-o-4.5 has the highest invariant share ($0.59$) but one of the smallest deactivation margins over the best monolingual mask ($0.39$~pp), whereas Kimi-Audio and Audio-Flamingo-3 have lower shares and larger margins. The one systematic observation we draw is that Qwen2.5-Omni-7B, with the lowest share ($0.30$), is the only model for which a monolingual mask (Mandarin) matches fusion under steering. We report this as a consistency check on the decomposition, not as evidence of a dose--response relation between invariant share and fusion benefit.

\section{Extended Related Work}
\label{app:related}
\paragraph{Controllable Affective Speech Synthesis.} A complementary line of research controls paralinguistic affect in speech generation through learned style representations \cite{skerry2018towards,wang2018style,wu2019end,liu2021expressive} and continuous control variables \cite{lorenzo2018investigating,lei2021fine,xie2025emosteerttsfinegrainedtrainingfreeemotioncontrollable}. These methods control affect via architectural conditioning learned during training, whereas we intervene training-free on neurons of pre-trained LALMs.

\paragraph{Neuron Attribution in Multimodal Models.} Beyond the audio modality, neuron-level dissection in multimodal foundation models has focused on modality-specific functional attributions \cite{huang2024minerminingunderlyingpattern,fang2024towards,xu2025deciphering,huo-etal-2024-mmneuron}, and audio interpretability studies rely on layer-wise probing of phonetic, speaker, and prosodic cues \cite{singla2022audio,10447060,Akman2025Improving}.

\section{Supplementary Results}
\label{appendix:supplementary}

\subsection{Cross-Lingual Agreement of Monolingual Emotion-Sensitive Neurons}
\label{app:agreement}
\begin{figure}[H]
    \centering
    \begin{subfigure}[b]{0.5\columnwidth}
        \centering
        \includegraphics[width=\linewidth]{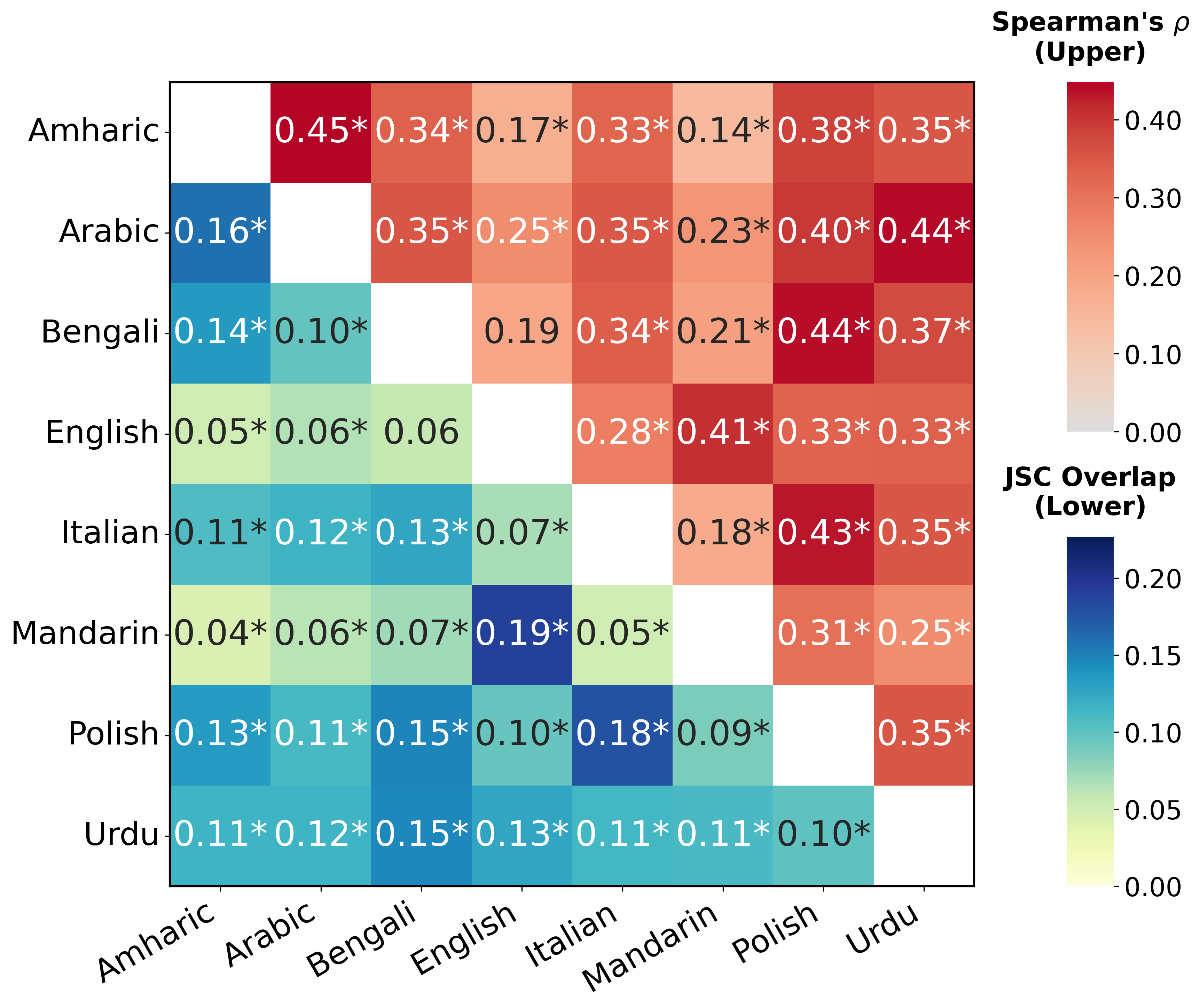}
        \caption{MiniCPM-o-4.5}
    \end{subfigure}\hfill
    \begin{subfigure}[b]{0.5\columnwidth}
        \centering
        \includegraphics[width=\linewidth]{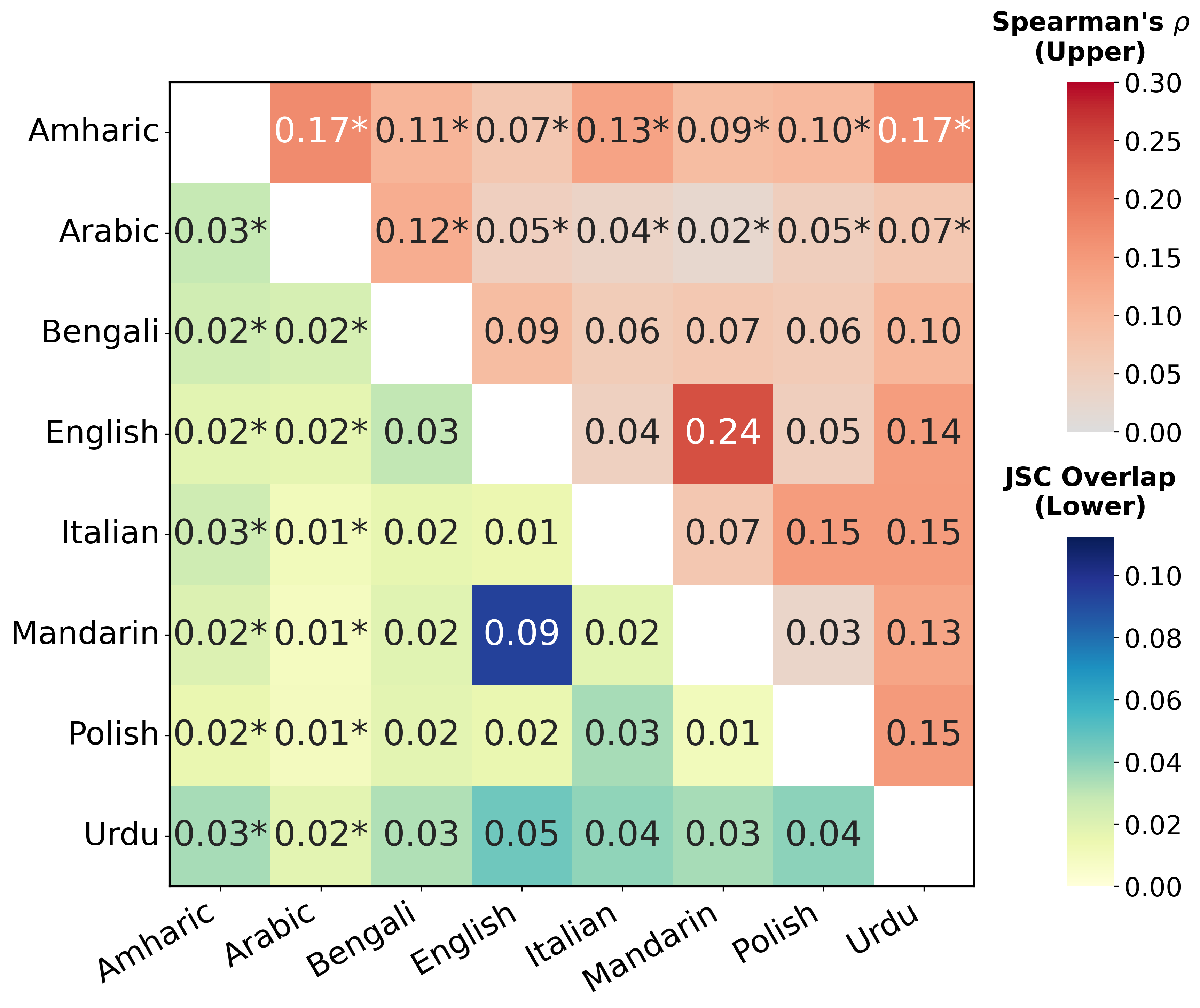}
        \caption{Qwen2.5-Omni-7B}
    \end{subfigure}\hfill
    \caption{Cross-lingual agreement of monolingually-identified ESNs for {MiniCPM-o-4.5 and Qwen2.5-Omni-7B.}}
    \label{fig:descriptive_stat_spplementary}
\end{figure}

\subsection{Saturation of Monolingual Evidence (BIIC)}
\label{app:saturation}
\begin{figure}[H]
    \centering
    \begin{subfigure}[b]{0.5\columnwidth}
        \centering
        \includegraphics[width=\linewidth]{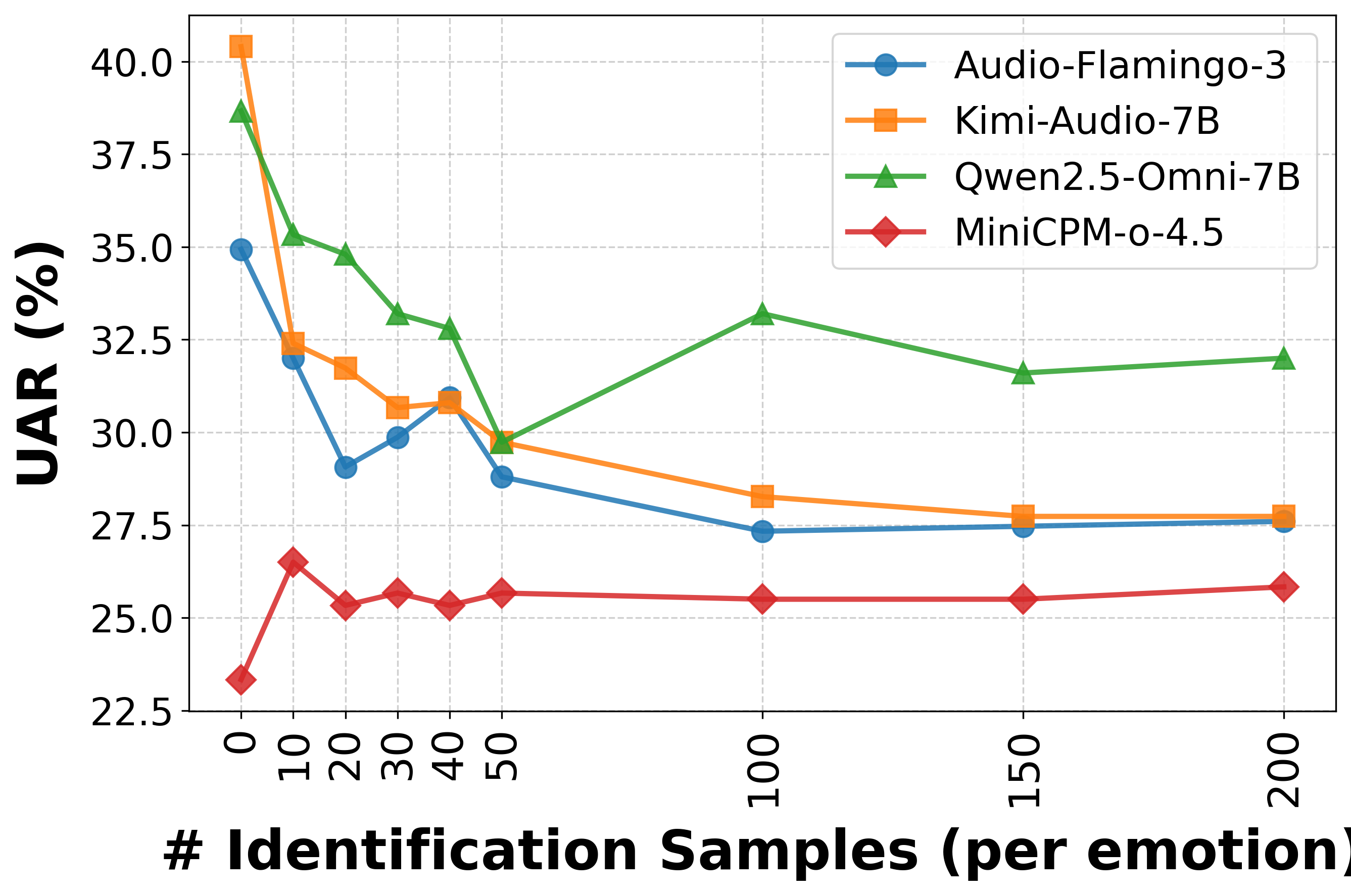}
        \caption{BIIC (Deact.)}
        \end{subfigure}\hfill
    \begin{subfigure}[b]{0.5\columnwidth}
        \centering
        \includegraphics[width=\linewidth]{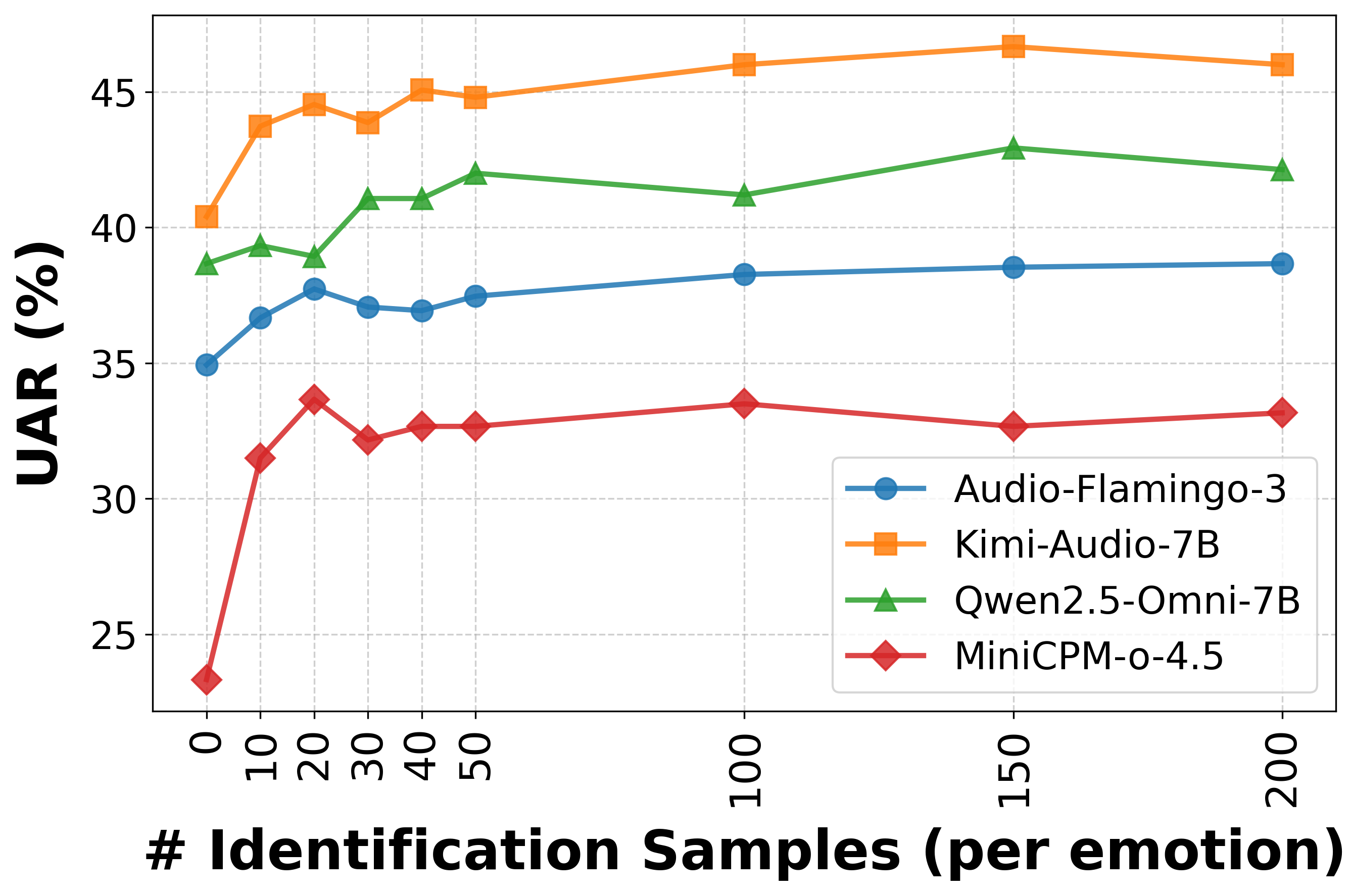}
        \caption{BIIC (Steering)}
    \end{subfigure}
    \caption{Saturation of monolingual evidence: UAR (\%) under intervention as identification sample size grows (BIIC).}
\end{figure}

\subsection{Sensitivity Analysis for Consistency Penalty}
\label{app:lambda}
\begin{figure}[H]
    \centering
    \begin{subfigure}[b]{0.5\columnwidth}
        \centering
        \includegraphics[width=\linewidth]{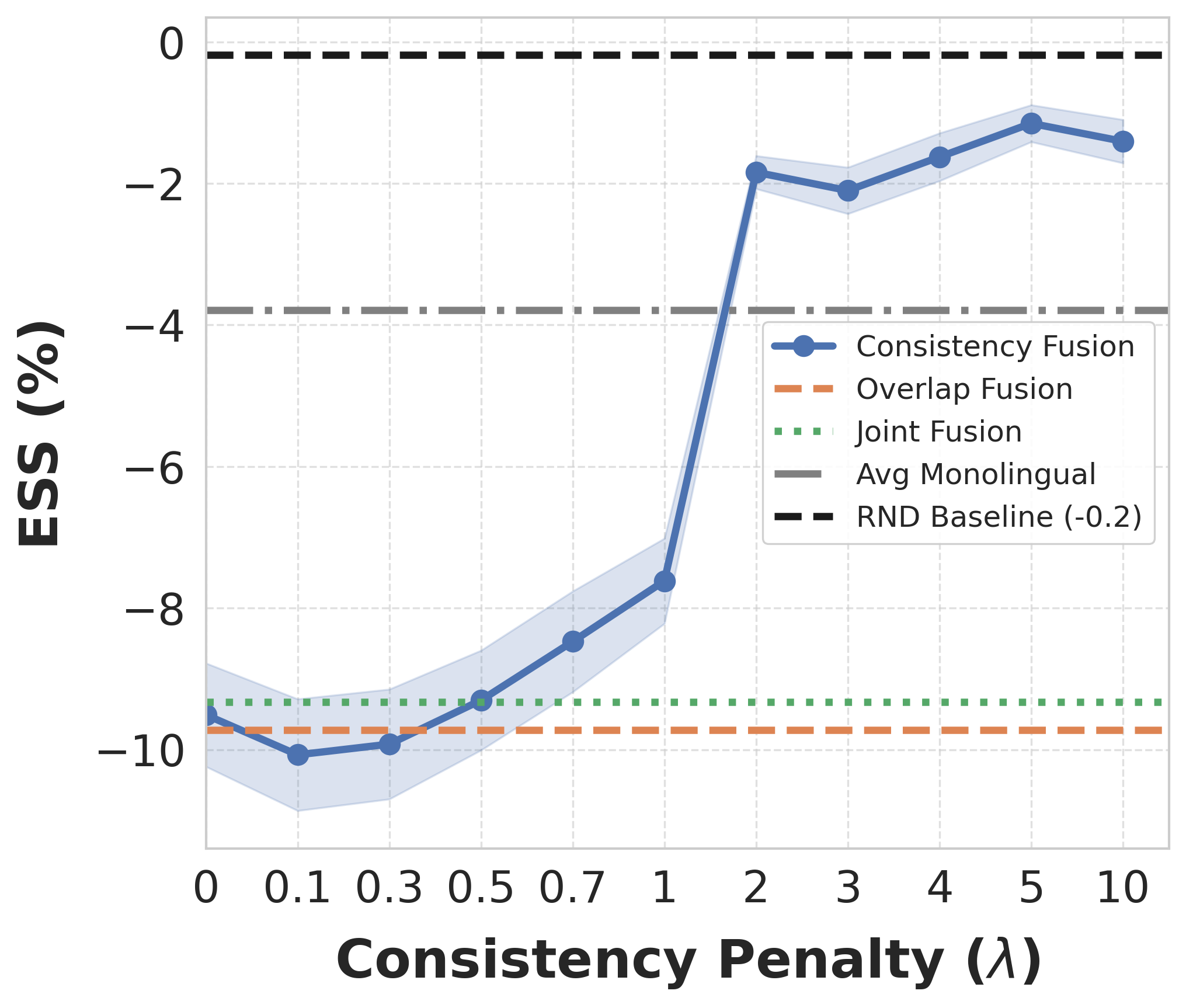}
        \caption{Audio-Flamingo-3 (Deact.)}
    \end{subfigure}\hfill
    \begin{subfigure}[b]{0.5\columnwidth}
        \centering
        \includegraphics[width=\linewidth]{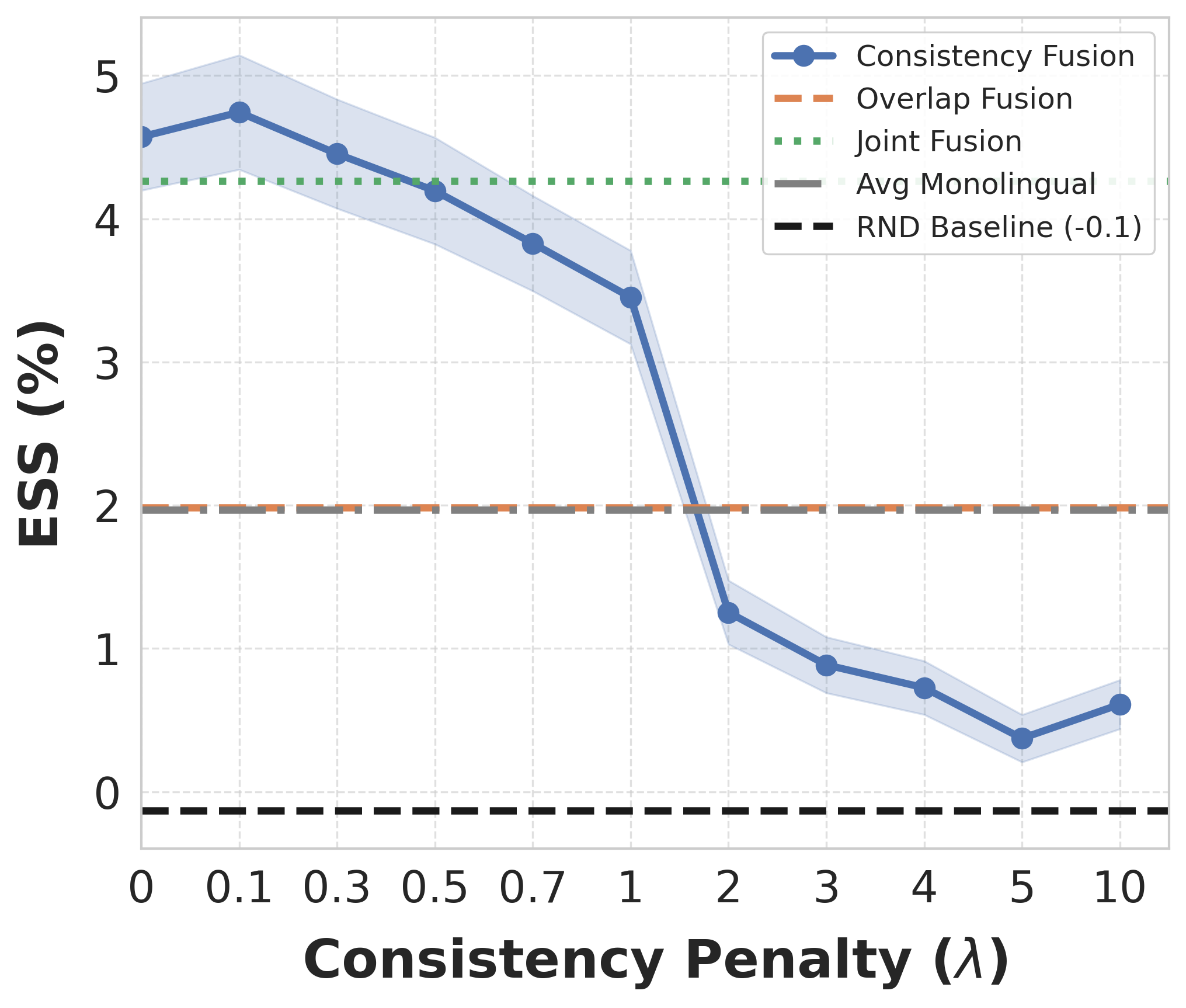}
        \caption{Audio-Flamingo-3 (Steer.)}
    \end{subfigure}

    \vspace{0.5em}

    \begin{subfigure}[b]{0.5\columnwidth}
        \centering
        \includegraphics[width=\linewidth]{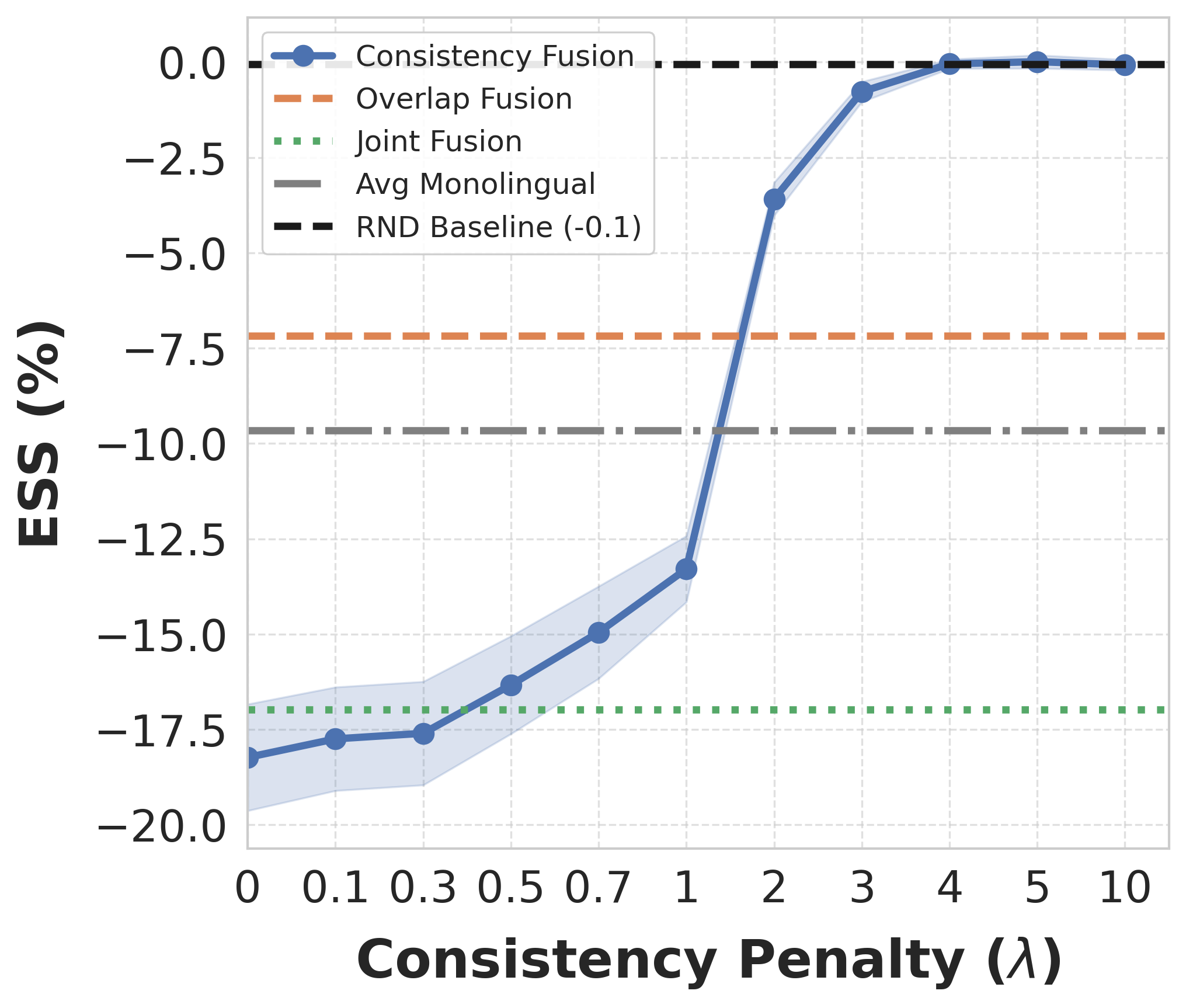}
        \caption{Kimi-Audio (Deact.)}
    \end{subfigure}\hfill
    \begin{subfigure}[b]{0.5\columnwidth}
        \centering
        \includegraphics[width=\linewidth]{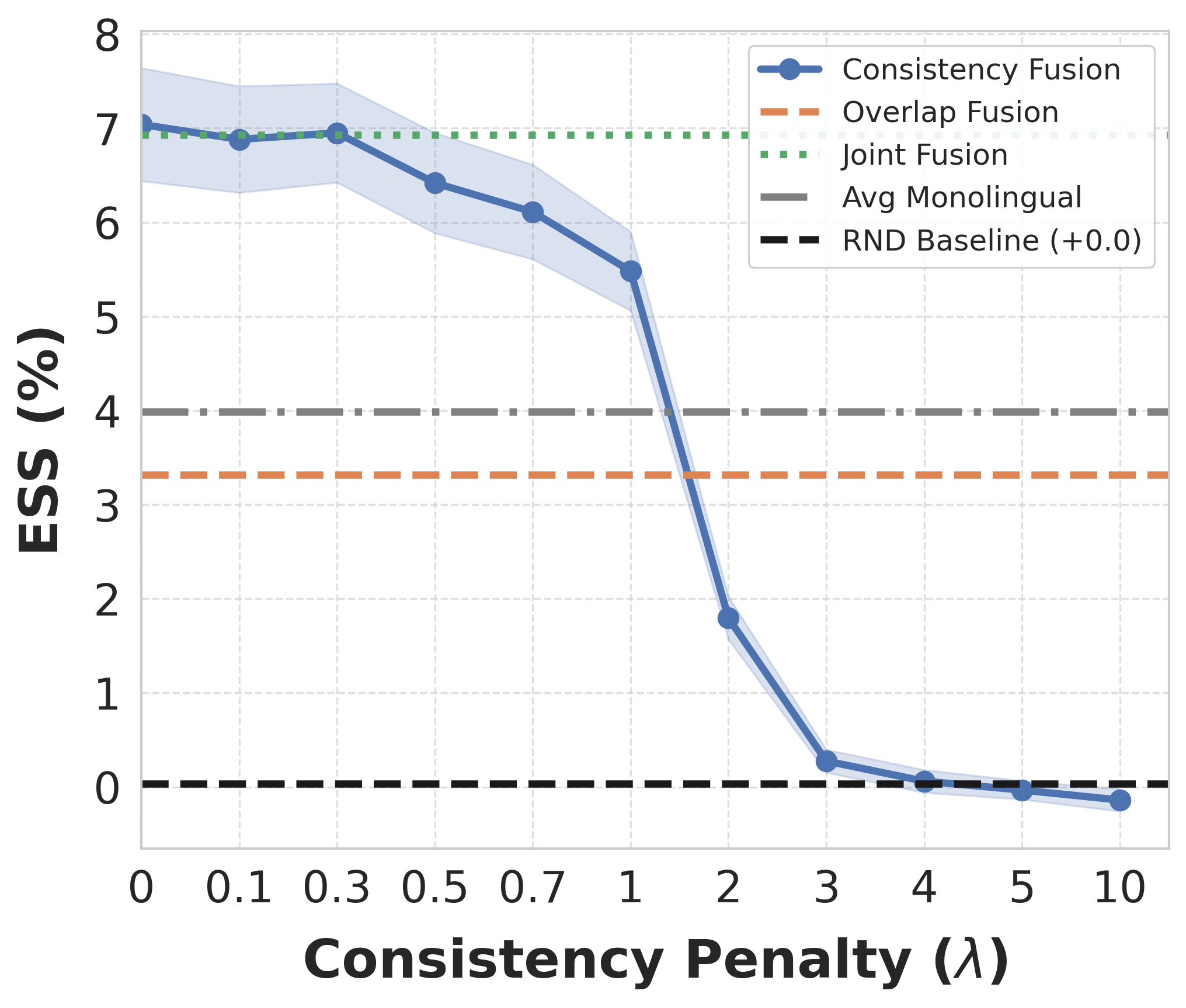}
        \caption{Kimi-Audio (Steering)}
    \end{subfigure}

    \vspace{0.5em}

    \begin{subfigure}[b]{0.5\columnwidth}
        \centering
        \includegraphics[width=\linewidth]{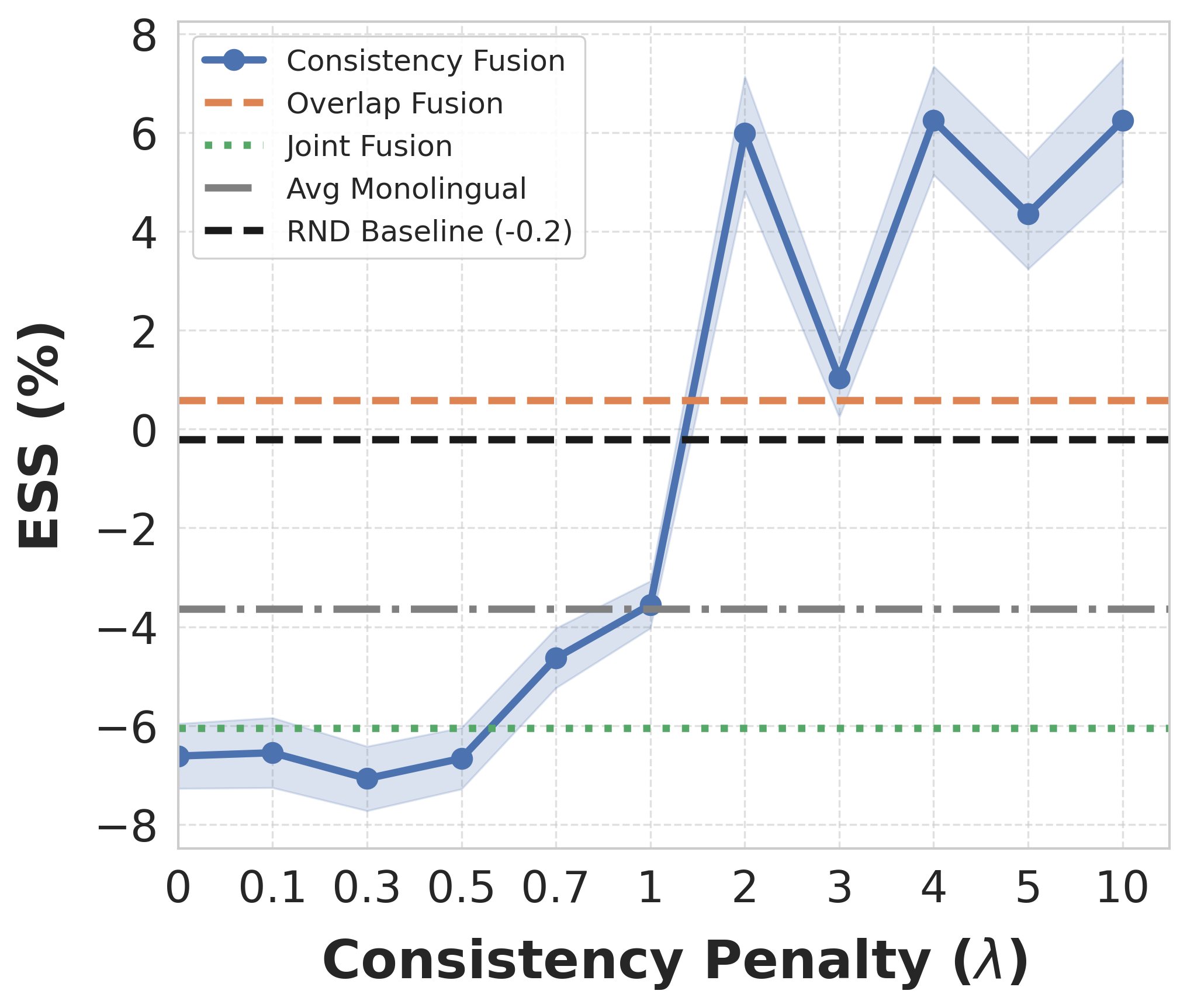}
        \caption{Qwen2.5-Omni (Deact.)}
    \end{subfigure}\hfill
    \begin{subfigure}[b]{0.5\columnwidth}
        \centering
        \includegraphics[width=\linewidth]{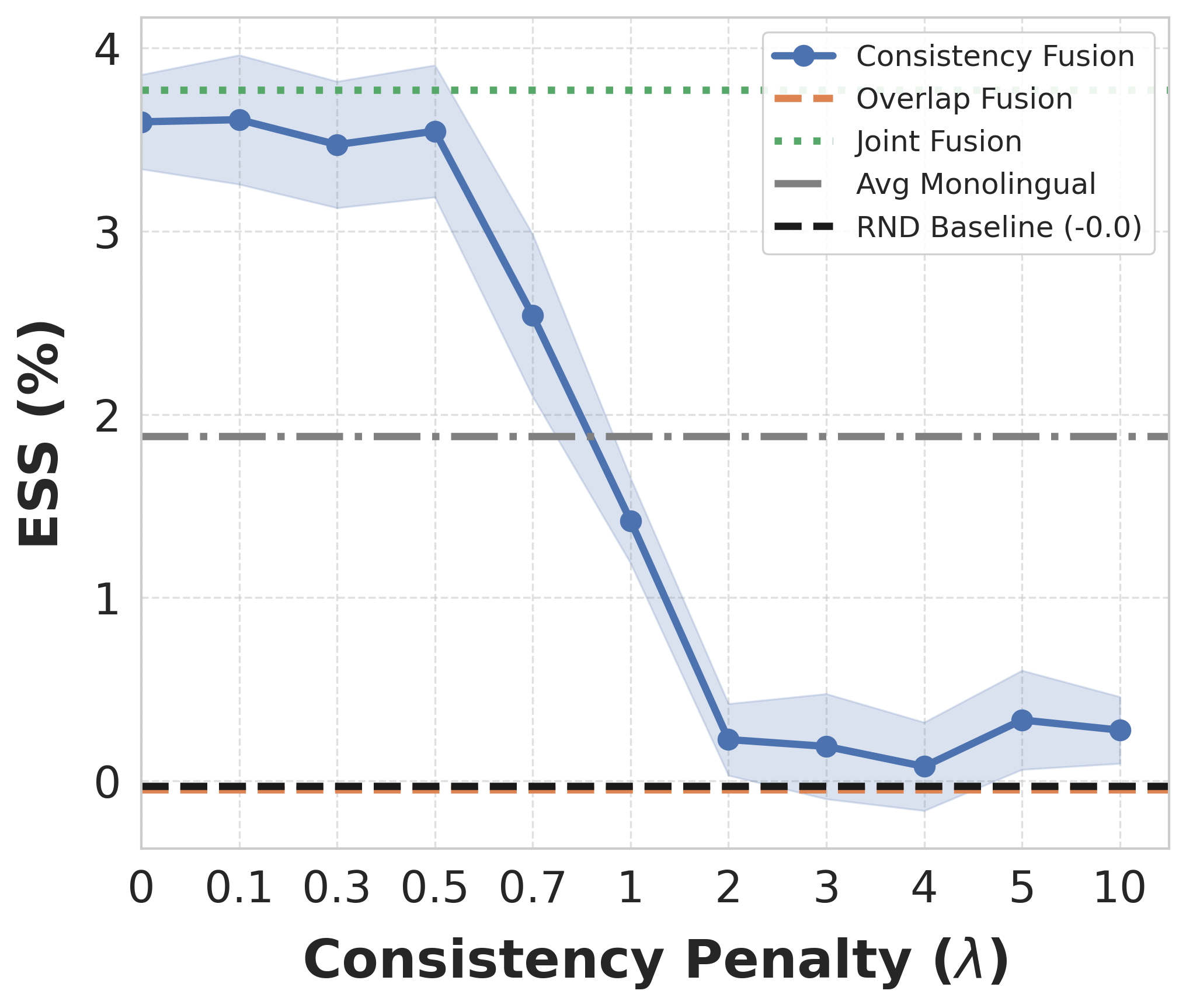}
        \caption{Qwen2.5-Omni (Steering)}
    \end{subfigure}
    
    \caption{Sensitivity to consistency penalty $\lambda$ under deactivation (left) and steering ($\alpha$=0.5, right), complementing Figure~\ref{fig:lambda}: (a,b) Audio-Flamingo-3, (c,d) Kimi-Audio, (e,f) Qwen2.5-Omni-7B. Shaded bands show $\pm 1$ SEM across 12 languages.}
    \label{fig:lambda_more}
\end{figure}

\subsection{Unintervened Baseline Results}
\label{app:baseline}

\begin{table*}[h]
\centering
\resizebox{\textwidth}{!}{%
\begin{tabular}{@{}llrrrrrrrrrrrr|r@{}}
\toprule
LALM       & Emotion   & English & Mandarin & Amharic & Arabic & Bengali & Italian & Polish & Urdu  & French & German & Persian & Russian & Avg.  \\ \midrule
           & Anger     & 70.00   & 57.33    & 82.00   & 36.67  & 90.00   & 35.33   & 10.67  & 86.00 & 93.75  & 100.00 & 95.33   & 75.33   & 69.37 \\
           & Fear      & 11.33   & 16.67    & 24.00   & 8.67   & 16.67   & 22.67   & 3.33   & 3.33  & 47.92  & 57.38  & 44.74   & 4.00    & 21.72 \\
Audio      & Happiness & 35.33   & 7.33     & 11.33   & 5.33   & 14.00   & 16.00   & 5.33   & 11.33 & 11.11  & 70.43  & 58.67   & 8.67    & 21.24 \\
Flamingo 3 & Neutral   & 82.00   & 78.67    & 68.67   & 89.33  & 78.67   & 98.00   & 96.67  & 54.00 & 79.17  & 99.04  & 84.00   & 76.67   & 82.07 \\
           & Sadness   & 29.33   & 14.67    & 43.33   & 13.33  & 41.33   & 27.33   & 8.00   & 35.33 & 56.94  & 90.08  & 69.33   & 22.67   & 37.64 \\ \cmidrule(l){2-15} 
           & UAR       & 45.60   & 34.93    & 45.87   & 30.67  & 48.13   & 39.87   & 24.80  & 38.00 & 57.78  & 83.39  & 70.41   & 37.47   & 46.41 \\ \midrule
           & Anger     & 60.00   & 38.67    & 78.00   & 50.00  & 87.33   & 57.33   & 11.33  & 91.33 & 90.28  & 94.16  & 82.67   & 66.67   & 67.31 \\
           & Fear      & 8.67    & 14.00    & 19.33   & 16.67  & 40.00   & 33.33   & 3.33   & 6.00  & 53.47  & 35.25  & 55.26   & 11.33   & 24.72 \\
Kimi-Audio & Happiness & 49.33   & 35.33    & 53.33   & 46.67  & 65.33   & 66.00   & 28.00  & 29.33 & 31.94  & 48.70  & 62.67   & 40.00   & 46.39 \\
           & Neutral   & 80.67   & 84.00    & 94.00   & 58.00  & 68.00   & 90.67   & 100.00 & 76.67 & 83.33  & 99.04  & 74.67   & 83.33   & 82.70 \\
           & Sadness   & 34.00   & 30.00    & 70.67   & 59.33  & 76.67   & 62.00   & 30.67  & 70.67 & 78.47  & 80.17  & 75.33   & 46.00   & 59.50 \\ \cmidrule(l){2-15} 
           & UAR       & 46.53   & 40.40    & 63.07   & 46.13  & 67.47   & 61.87   & 34.67  & 54.80 & 67.50  & 71.46  & 70.12   & 49.47   & 56.12 \\ \midrule
           & Anger     & 14.00   & 5.33     & 10.67   & 5.33   & 58.00   & 8.67    & 0.67   & 34.00 & 38.19  & 21.17  & 22.67   & 25.33   & 20.34 \\
           & Fear      & 1.33    & 1.33     & 1.33    & 0.00   & 1.33    & 0.00    & 0.00   & 0.00  & 2.78   & 0.00   & 0.00    & 1.33    & 0.79  \\
MiniCPM    & Happiness & 20.67   & 8.67     & 52.67   & 4.67   & 8.00    & 28.00   & 5.33   & 8.00  & 25.69  & 15.65  & 20.67   & 14.00   & 17.67 \\
-o-4.5     & Neutral   & 98.67   & 99.33    & 100.00  & 98.67  & 93.33   & 100.00  & 99.33  & 99.33 & 100.00 & 100.00 & 99.33   & 98.00   & 98.83 \\
           & Sadness   & 2.67    & 2.00     & 22.67   & 0.00   & 36.00   & 5.33    & 0.00   & 23.33 & 30.56  & 2.48   & 30.67   & 8.67    & 13.70 \\ \cmidrule(l){2-15} 
           & UAR       & 27.47   & 23.33    & 37.47   & 21.73  & 39.33   & 28.40   & 21.07  & 32.93 & 39.44  & 27.86  & 34.67   & 29.47   & 30.26 \\ \midrule
           & Anger     & 54.00   & 52.67    & 50.67   & 48.67  & 46.67   & 8.67    & 32.67  & 64.00 & 40.97  & 32.85  & 55.33   & 44.00   & 44.26 \\
           & Fear      & 13.33   & 24.67    & 2.00    & 2.67   & 2.67    & 2.00    & 2.00   & 5.33  & 4.86   & 10.66  & 7.89    & 8.00    & 7.17  \\
Qwen2.5-   & Happiness & 52.67   & 46.00    & 17.33   & 24.00  & 26.00   & 24.67   & 21.33  & 34.67 & 17.36  & 26.09  & 52.67   & 42.67   & 32.12 \\
Omni-7B    & Neutral   & 24.67   & 38.00    & 19.33   & 15.33  & 18.00   & 28.00   & 16.00  & 12.00 & 22.22  & 21.15  & 10.00   & 28.67   & 21.11 \\
           & Sadness   & 40.00   & 32.00    & 47.33   & 36.67  & 44.00   & 44.67   & 40.00  & 48.67 & 35.42  & 39.67  & 53.33   & 36.67   & 41.53 \\ \cmidrule(l){2-15} 
           & UAR       & 36.93   & 38.67    & 27.33   & 25.47  & 27.47   & 21.60   & 22.40  & 32.93 & 24.17  & 26.08  & 35.85   & 32.00   & 29.24 \\ \bottomrule
\end{tabular}%
}
\caption{Baseline SER accuracy (\%) across 12 languages with deterministic decoding. ``UAR'' rows report unweighted average recall computed over available emotions; ``Avg.'' column reports UAR across languages.}
\label{tab:baseline_full}
\end{table*}

Table~\ref{tab:baseline_full} shows the baseline SER accuracies for unintervened models. Test set class distributions vary across datasets, with some emotions (e.g., fear in Persian) having few or zero instances. Global ESS is computed only over emotions with available test cases; per-emotion breakdowns in Appendix~\ref{appendix:per_emotion} assess robustness across class frequencies.
We note that deactivation effects are bounded by baseline accuracy, creating potential floor effects for low-accuracy emotions. Fear's weak deactivation effect in MiniCPM-o-4.5 (baseline $\approx1\%$) should be interpreted cautiously given this constraint.

\subsection{Per-Language Detailed Results}
\label{appendix:per_emotion}

\begin{figure}[ht!]
    \centering
    \begin{subfigure}[b]{\columnwidth}
        \centering
        \includegraphics[width=\linewidth]{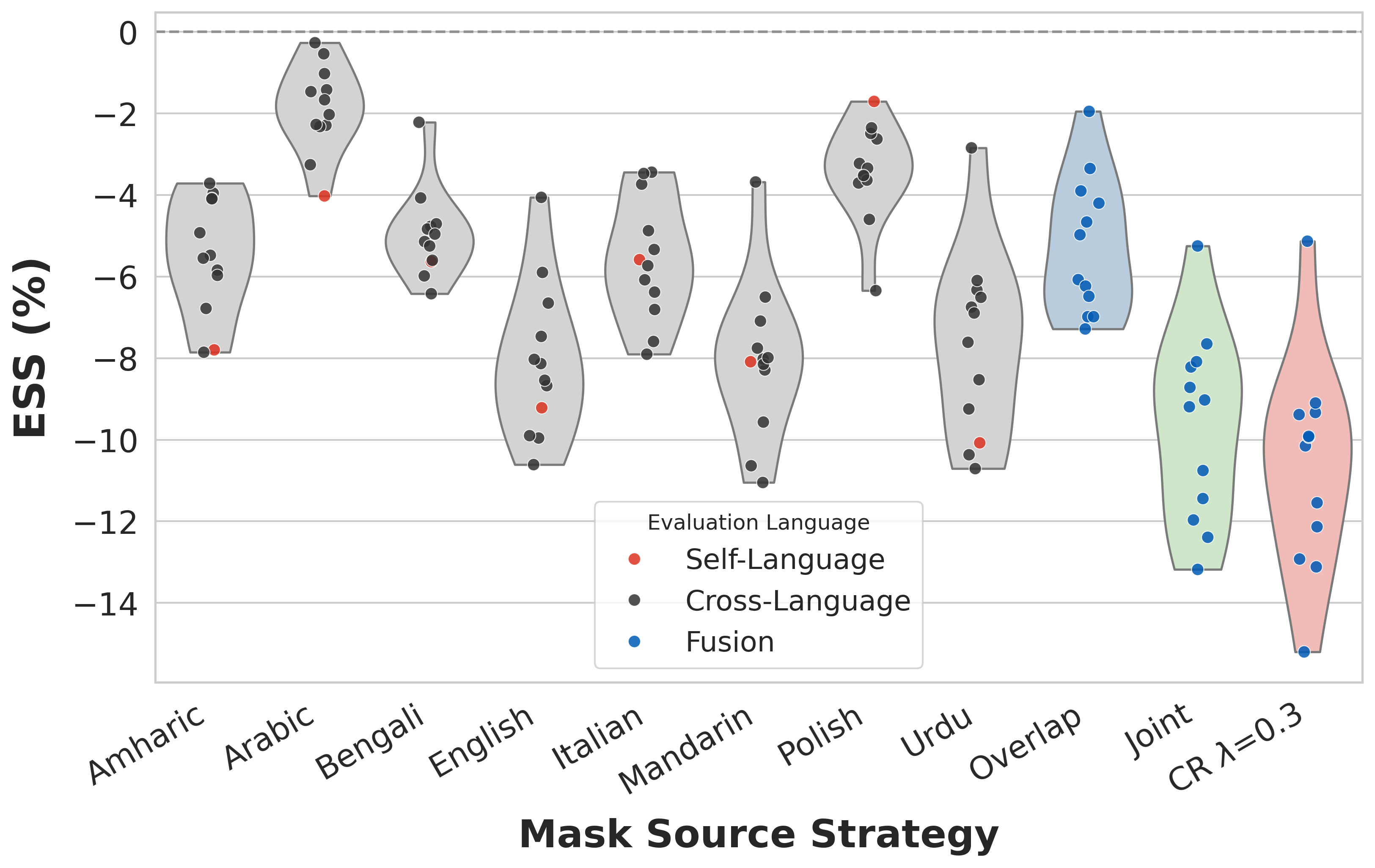}
        \caption{ESS Distribution, Deactivation}
    \end{subfigure}

    \vspace{0.5em}
    
    \begin{subfigure}[b]{\columnwidth}
        \centering
        \includegraphics[width=\linewidth]{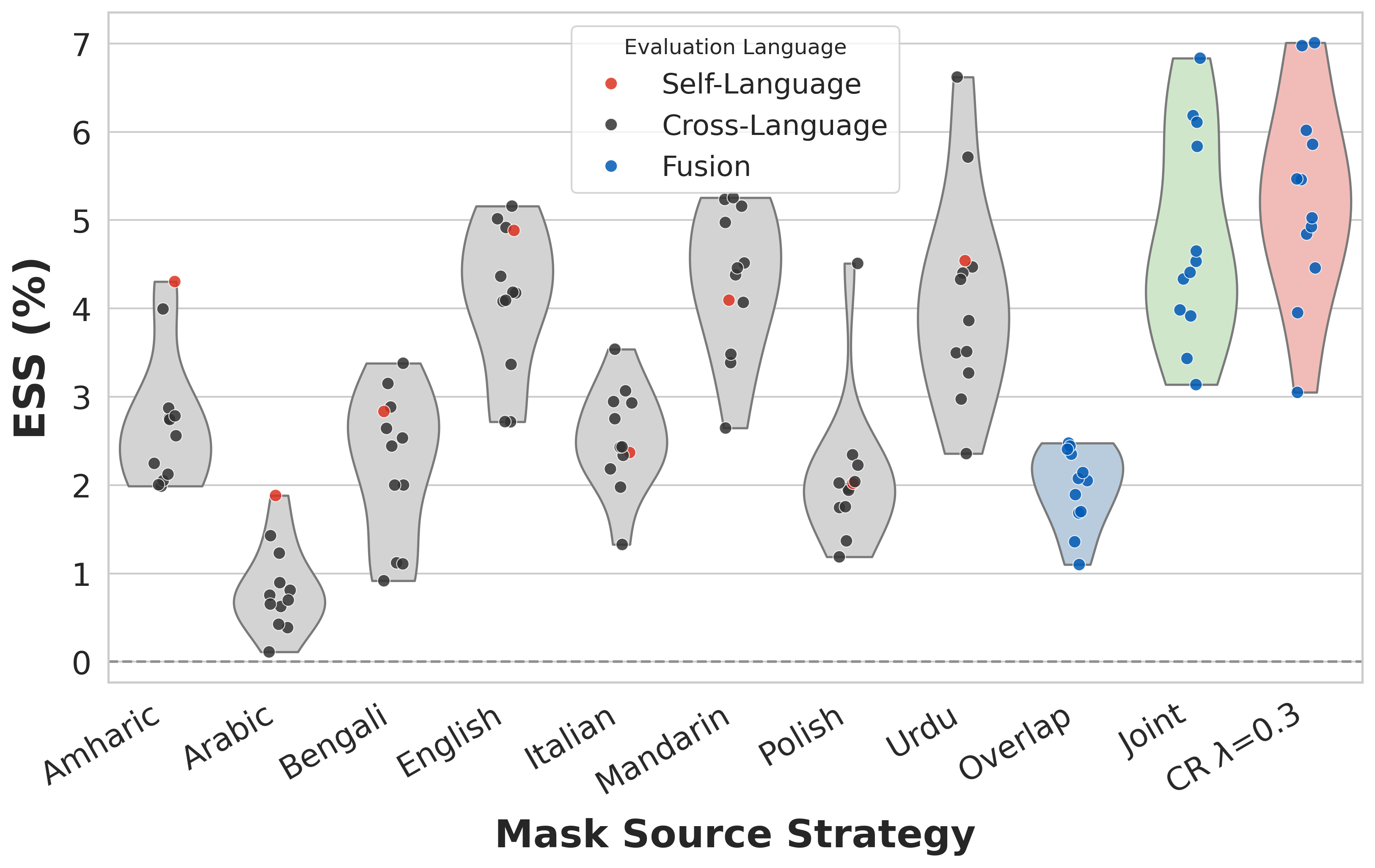}
    \end{subfigure}
    \caption{ESS distributions across monolingual and fusion masks under (a) deactivation and (b) steering.}
    \label{fig:ess_dis}
\end{figure}

Table~\ref{tab:deact_results_full} and \ref{tab:steer_results_full} provide per-language intervention results. Figure~\ref{fig:ess_dis} displays ESS distributions across mask sources and evaluation settings under deactivation, averaged over four LALMs. Monolingual masks show substantial variability, with high-resource anchors (English, Mandarin) and certain low-resource languages (Bengali, Urdu) achieving stronger effects than others (Amharic, Arabic). Consistency fusion ($\lambda$=0.3) produces a distribution shifted toward the strongest effects, surpassing the best monolingual mask.

\begin{table*}[h]
\centering
\resizebox{\textwidth}{!}{%
\begin{tabular}{@{}lllrrrrrrrrrrr@{}}
\toprule
\multirow{2}{*}{LALM} & \multirow{2}{*}{Group}                 &          & \multicolumn{8}{c}{Monolingual ESN Masks}                                   & \multicolumn{3}{c}{Fusion MLEN Masks} \\ \cmidrule(lr){4-11} \cmidrule(lr){12-14}
                      &                                       &          & English & Mandarin & Amharic & Arabic & Bengali & Italian & Polish & Urdu   & Overlap   & Joint    & CR ($\lambda{=}0.3$)   \\ \midrule
                      & \multirow{8}{*}{$\mathcal{L}_\text{id}$}   & English  & -7.43   & -5.17    & -3.30   & -0.63  & -3.77   & -1.67   & -4.70  & -4.67  & -8.03     & -7.60    & -9.27          \\
                      &                                       & Mandarin & -8.97   & -5.97    & -4.33   & -0.17  & -3.87   & -2.30   & -6.03  & -8.70  & -14.23    & -11.87   & -12.53         \\
                      &                                       & Amharic  & -3.90   & -2.40    & -4.50   & -2.27  & -2.83   & -1.80   & -4.60  & -5.33  & -12.30    & -8.90    & -9.70          \\
                      &                                       & Arabic   & -6.30   & -2.97    & -3.53   & -3.43  & -2.90   & -1.23   & -5.33  & -5.63  & -7.37     & -9.97    & -10.03         \\
                      &                                       & Bengali  & -6.70   & -4.60    & -5.67   & -1.63  & -3.67   & -3.23   & -5.57  & -8.13  & -14.33    & -11.17   & -11.97         \\
Audio                 &                                       & Italian  & -4.37   & -4.20    & -2.83   & -2.63  & -2.33   & -2.70   & -4.23  & -4.63  & -8.07     & -8.40    & -8.40          \\
Flamingo 3            &                                       & Polish   & -2.20   & -1.73    & -2.10   & -0.43  & -1.47   & -0.77   & -1.87  & -2.93  & -3.90     & -4.27    & -4.77          \\
                      &                                       & Urdu     & -7.50   & -4.43    & -4.53   & -0.23  & -2.50   & -2.50   & -5.27  & -9.27  & -13.00    & -11.47   & -12.70         \\ \cmidrule(l){2-14} 
                      & \multirow{4}{*}{$\mathcal{L}_\text{held}$} & French   & -6.46   & -3.89    & -4.13   & -2.01  & -3.12   & -1.77   & -5.59  & -4.10  & -7.64     & -11.49   & -10.83         \\
                      &                                       & German   & -2.36   & -5.76    & -1.38   & -1.54  & -4.13   & -1.85   & -3.76  & -1.99  & -1.24     & -6.27    & -5.50          \\
                      &                                       & Persian  & -6.32   & -5.44    & -3.00   & -1.82  & -4.76   & -1.59   & -3.22  & -4.84  & -11.95    & -11.53   & -13.24         \\
                      &                                       & Russian  & -7.43   & -4.27    & -4.73   & 0.00   & -1.73   & -1.37   & -4.73  & -7.53  & -14.60    & -9.00    & -10.13         \\ \midrule
                      & \multirow{8}{*}{$\mathcal{L}_\text{id}$}   & English  & -13.17  & -10.27   & -5.60   & -2.10  & -9.47   & -9.33   & -5.00  & -10.27 & -4.70     & -13.07   & -13.97         \\
                      &                                       & Mandarin & -10.87  & -11.40   & -4.53   & -2.40  & -8.97   & -7.83   & -3.90  & -11.87 & -4.93     & -13.70   & -14.00         \\
                      &                                       & Amharic  & -11.70  & -11.90   & -11.30  & -5.60  & -14.60  & -15.13  & -7.70  & -13.43 & -9.00     & -19.43   & -19.17         \\
                      &                                       & Arabic   & -16.00  & -15.23   & -9.61   & -6.51  & -13.37  & -16.90  & -7.53  & -15.93 & -9.40     & -19.20   & -19.63         \\
                      &                                       & Bengali  & -16.87  & -15.80   & -8.30   & -4.50  & -13.40  & -14.87  & -4.43  & -15.90 & -9.70     & -23.43   & -22.13         \\
Kimi-Audio            &                                       & Italian  & -12.07  & -11.43   & -6.10   & -3.70  & -10.50  & -13.63  & -2.70  & -11.57 & -6.40     & -14.93   & -15.93         \\
                      &                                       & Polish   & -7.90   & -6.20    & -4.37   & -2.87  & -5.80   & -8.57   & -2.73  & -8.93  & -4.67     & -9.93    & -10.17         \\
                      &                                       & Urdu     & -15.43  & -14.13   & -7.97   & -2.93  & -10.37  & -16.17  & -4.77  & -15.90 & -6.67     & -19.73   & -18.97         \\ \cmidrule(l){2-14} 
                      & \multirow{4}{*}{$\mathcal{L}_\text{held}$} & French   & -8.51   & -7.19    & -5.56   & -4.58  & -7.50   & -11.22  & -2.50  & -12.19 & -6.53     & -15.00   & -16.60         \\
                      &                                       & German   & -9.94   & -12.57   & -2.48   & -2.80  & -7.91   & -12.28  & -3.97  & -13.65 & -6.05     & -15.07   & -17.73         \\
                      &                                       & Persian  & -21.22  & -19.68   & -8.85   & -1.99  & -15.13  & -20.62  & -5.36  & -21.27 & -11.01    & -25.23   & -28.48         \\
                      &                                       & Russian  & -11.53  & -10.73   & -6.40   & -5.13  & -9.40   & -10.43  & -2.03  & -12.97 & -7.07     & -15.00   & -14.43         \\ \midrule
                      & \multirow{8}{*}{$\mathcal{L}_\text{id}$}   & English  & -5.80   & -5.93    & 0.07    & -0.33  & -3.53   & -1.70   & -2.47  & -4.80  & -4.43     & -4.73    & -5.70          \\
                      &                                       & Mandarin & -3.30   & -3.13    & -0.53   & 0.03   & -1.23   & -1.03   & -1.27  & -2.83  & -3.07     & -2.43    & -3.67          \\
                      &                                       & Amharic  & -9.23   & -11.20   & -3.50   & -1.10  & -5.53   & -3.07   & -9.47  & -12.70 & -9.97     & -12.23   & -14.00         \\
                      &                                       & Arabic   & -2.27   & -1.87    & -0.33   & -0.33  & -1.30   & -0.67   & -0.70  & -1.73  & -1.37     & -1.57    & -1.80          \\
                      &                                       & Bengali  & -10.80  & -13.13   & -6.00   & -0.57  & -3.30   & -6.70   & -0.90  & -11.60 & -5.30     & -9.80    & -13.50         \\
MiniCPM               &                                       & Italian  & -6.20   & -5.80    & -1.43   & -1.80  & -4.27   & -1.60   & -4.63  & -6.33  & -5.70     & -6.00    & -6.63          \\
-o-4.5                &                                       & Polish   & -1.47   & -1.43    & -0.30   & -0.30  & -0.70   & -0.50   & -1.00  & -1.20  & -1.10     & -1.03    & -1.37          \\
                      &                                       & Urdu     & -9.07   & -11.30   & -4.53   & -0.17  & -3.20   & -5.60   & -0.97  & -8.60  & -4.17     & -7.13    & -11.67         \\ \cmidrule(l){2-14} 
                      & \multirow{4}{*}{$\mathcal{L}_\text{held}$} & French   & -9.72   & -12.01   & -4.79   & -1.39  & -3.09   & -6.74   & -4.37  & -11.49 & -6.11     & -10.03   & -13.16         \\
                      &                                       & German   & -5.63   & -6.37    & -1.07   & -1.04  & -3.73   & -2.95   & -2.47  & -4.98  & -3.60     & -4.66    & -5.77          \\
                      &                                       & Persian  & -9.50   & -11.60   & -5.10   & -0.07  & -4.07   & -4.50   & -2.90  & -10.96 & -7.23     & -9.93    & -11.77         \\
                      &                                       & Russian  & -6.60   & -7.90    & -2.70   & -0.47  & -2.53   & -3.47   & -1.57  & -5.73  & -4.47     & -5.60    & -7.33          \\ \midrule
                      & \multirow{8}{*}{$\mathcal{L}_\text{id}$}   & English  & -10.47  & -8.73    & -5.57   & -2.90  & -5.63   & -0.30   & 0.20   & -3.03  & 0.37      & -5.20    & -8.40          \\
                      &                                       & Mandarin & -11.57  & -10.03   & -5.00   & 0.43   & -6.93   & -2.77   & -0.77  & -1.23  & -2.07     & -4.87    & -10.40         \\
                      &                                       & Amharic  & -5.03   & -5.20    & -8.13   & -0.53  & -0.97   & -2.70   & -1.13  & -3.03  & 2.13      & -7.30    & -5.67          \\
                      &                                       & Arabic   & -7.97   & -6.70    & -5.87   & -2.20  & -1.50   & -1.43   & -1.97  & -0.57  & 2.53      & -6.03    & -6.07          \\
                      &                                       & Bengali  & -5.47   & -6.07    & -7.50   & 0.37   & -2.17   & -0.83   & 0.50   & -3.23  & 1.40      & -5.17    & -4.10          \\
Qwen2.5-              &                                       & Italian  & -3.97   & -2.77    & -6.60   & -3.37  & -3.47   & -3.07   & 2.27   & -2.73  & 0.27      & -5.53    & -5.43          \\
Omni-7B               &                                       & Polish   & -4.67   & -4.50    & -8.00   & 2.03   & -0.90   & -3.33   & -0.27  & 2.30   & 1.87      & -5.77    & -4.23          \\
                      &                                       & Urdu     & -7.60   & -6.23    & -6.97   & -0.47  & -3.27   & -4.53   & -1.10  & -4.23  & -1.10     & -7.43    & -9.13          \\ \cmidrule(l){2-14} 
                      & \multirow{4}{*}{$\mathcal{L}_\text{held}$} & French   & -7.43   & -5.66    & -5.80   & 0.00   & -2.57   & -3.99   & -0.52  & -3.99  & 1.63      & -6.49    & -5.59          \\
                      &                                       & German   & -5.66   & -5.72    & -7.83   & -0.58  & -3.05   & -4.55   & 1.88   & -5.15  & -2.51     & -6.34    & -10.75         \\
                      &                                       & Persian  & -5.41   & -5.12    & -4.41   & 0.56   & -1.72   & -3.56   & -0.19  & -3.73  & 2.26      & -6.02    & -7.32          \\
                      &                                       & Russian  & -8.60   & -7.13    & -6.03   & -2.23  & -6.17   & -3.17   & -0.03  & -2.47  & 0.20      & -6.50    & -7.77          \\ \bottomrule
\end{tabular}%
}
\caption{Per-language deactivation effects. More negative values indicate stronger causal necessity of deactivated neurons.
}
\label{tab:deact_results_full}
\end{table*}

\begin{table*}[h]
\centering
\resizebox{\textwidth}{!}{%
\begin{tabular}{@{}lllrrrrrrrrrrr@{}}
\toprule
\multirow{2}{*}{LALM} & \multirow{2}{*}{Group}                 & Lang.    & \multicolumn{8}{c}{Monolingual ESN Masks}                                 & \multicolumn{3}{c}{Fusion MLEN Masks} \\ \cmidrule(lr){4-11} \cmidrule(lr){12-14}
                      &                                       & Set      & English & Mandarin & Amharic & Arabic & Bengali & Italian & Polish & Urdu & Overlap    & Joint   & CR ($\lambda{=}0.3$)   \\ \midrule
                      & \multirow{8}{*}{$\mathcal{L}_\text{id}$}   & English  & 4.47    & 3.53     & 1.13    & 0.10   & 1.27    & 1.37    & 2.83   & 3.33 & 2.37       & 3.77    & 4.60           \\
                      &                                       & Mandarin & 4.20    & 2.57     & 1.93    & -0.03  & 1.67    & 1.23    & 2.37   & 3.43 & 2.63       & 3.87    & 4.50           \\
                      &                                       & Amharic  & 4.00    & 1.13     & 3.50    & 0.63   & 1.60    & 1.17    & 2.27   & 5.03 & 2.47       & 5.53    & 6.03           \\
                      &                                       & Arabic   & 5.10    & 1.70     & 2.43    & 1.40   & 1.60    & 0.77    & 5.43   & 4.50 & 3.30       & 5.83    & 5.40           \\
                      &                                       & Bengali  & 4.30    & 2.37     & 3.27    & 0.27   & 2.73    & 0.80    & 2.17   & 4.57 & 2.87       & 6.30    & 6.60           \\
Audio                 &                                       & Italian  & 1.70    & 1.80     & 0.57    & 0.67   & 1.27    & 0.63    & 1.97   & 2.03 & 1.27       & 3.17    & 3.10           \\
Flamingo 3            &                                       & Polish   & 3.00    & 1.40     & 1.27    & -0.03  & 0.63    & 0.47    & 3.13   & 2.40 & 1.37       & 3.53    & 3.30           \\
                      &                                       & Urdu     & 3.00    & 2.40     & 1.00    & 0.60   & 1.33    & 0.83    & 2.40   & 3.47 & 1.50       & 4.00    & 3.90           \\ \cmidrule(l){2-14} 
                      & \multirow{4}{*}{$\mathcal{L}_\text{held}$} & French   & 2.95    & 1.88     & 1.87    & 0.66   & 2.36    & 2.36    & 2.12   & 3.09 & 2.26       & 5.10    & 5.28           \\
                      &                                       & German   & 1.12    & 1.48     & 0.68    & -0.14  & 0.57    & 0.65    & 1.00   & 0.52 & -0.22      & 1.75    & 1.84           \\
                      &                                       & Persian  & 2.99    & 2.36     & 1.93    & 0.32   & 1.55    & 1.48    & 1.10   & 2.15 & 1.99       & 4.32    & 4.69           \\
                      &                                       & Russian  & 3.80    & 2.43     & 2.13    & 0.27   & 1.13    & 0.27    & 2.70   & 3.23 & 2.00       & 4.00    & 4.20           \\ \midrule
                      & \multirow{8}{*}{$\mathcal{L}_\text{id}$}   & English  & 5.70    & 4.83     & 2.60    & 0.80   & 4.10    & 4.23    & 2.10   & 4.70 & 2.70       & 5.43    & 5.63           \\
                      &                                       & Mandarin & 5.27    & 5.67     & 2.13    & 2.37   & 3.83    & 3.93    & 1.93   & 4.73 & 2.37       & 6.33    & 7.17           \\
                      &                                       & Amharic  & 6.30    & 6.57     & 5.37    & 4.07   & 6.53    & 6.90    & 3.30   & 8.27 & 5.30       & 9.40    & 8.70           \\
                      &                                       & Arabic   & 8.53    & 8.33     & 4.73    & 4.13   & 7.97    & 7.37    & 4.50   & 7.93 & 5.07       & 10.97   & 10.67          \\
                      &                                       & Bengali  & 5.73    & 4.87     & 4.63    & 2.13   & 5.13    & 6.13    & 1.67   & 6.20 & 3.97       & 8.07    & 8.03           \\
Kimi-Audio            &                                       & Italian  & 4.47    & 3.77     & 1.90    & 1.30   & 4.60    & 5.30    & 0.57   & 4.73 & 2.50       & 7.30    & 7.17           \\
                      &                                       & Polish   & 3.17    & 2.70     & 1.83    & 1.10   & 2.43    & 2.77    & 0.83   & 3.73 & 1.97       & 4.23    & 4.60           \\
                      &                                       & Urdu     & 3.73    & 4.77     & 2.57    & 1.03   & 3.80    & 4.03    & 2.60   & 4.37 & 2.40       & 6.23    & 6.07           \\ \cmidrule(l){2-14} 
                      & \multirow{4}{*}{$\mathcal{L}_\text{held}$} & French   & 4.51    & 2.88     & 1.74    & 1.81   & 2.71    & 3.19    & 0.87   & 4.34 & 2.92       & 5.03    & 5.56           \\
                      &                                       & German   & 2.48    & 3.59     & 1.18    & 2.48   & 2.52    & 3.06    & 1.88   & 3.83 & 2.92       & 4.35    & 4.34           \\
                      &                                       & Persian  & 6.74    & 6.43     & 3.55    & 1.43   & 7.08    & 5.71    & 2.76   & 6.08 & 4.11       & 9.03    & 8.13           \\
                      &                                       & Russian  & 5.07    & 5.23     & 3.23    & 2.13   & 4.30    & 4.23    & 1.30   & 6.33 & 3.63       & 6.73    & 7.33           \\ \midrule
                      & \multirow{8}{*}{$\mathcal{L}_\text{id}$}   & English  & 3.83    & 3.60     & 0.50    & -0.03  & 2.20    & 1.07    & 2.07   & 3.03 & 2.43       & 4.27    & 3.67           \\
                      &                                       & Mandarin & 2.40    & 3.33     & 0.33    & 0.23   & 1.93    & 1.37    & 2.00   & 2.83 & 1.60       & 2.90    & 3.77           \\
                      &                                       & Amharic  & 6.30    & 7.77     & 4.00    & 0.97   & 2.40    & 3.80    & 3.43   & 9.07 & 3.70       & 5.00    & 10.07          \\
                      &                                       & Arabic   & 2.20    & 3.23     & 1.13    & 0.13   & 1.60    & 1.27    & 2.73   & 3.93 & 1.83       & 3.37    & 4.07           \\
                      &                                       & Bengali  & 6.80    & 8.53     & 4.27    & -0.03  & 3.03    & 4.27    & 2.23   & 8.53 & 3.87       & 7.40    & 10.33          \\
MiniCPM               &                                       & Italian  & 4.63    & 5.30     & 1.43    & 0.93   & 2.93    & 1.97    & 5.77   & 5.50 & 4.33       & 4.83    & 5.90           \\
-o-4.5                &                                       & Polish   & 1.63    & 2.13     & 0.60    & 0.20   & 1.37    & 0.53    & 3.13   & 2.43 & 1.73       & 2.83    & 2.17           \\
                      &                                       & Urdu     & 5.67    & 7.30     & 3.57    & -0.30  & 1.37    & 3.43    & 1.10   & 6.03 & 1.20       & 3.73    & 7.77           \\ \cmidrule(l){2-14} 
                      & \multirow{4}{*}{$\mathcal{L}_\text{held}$} & French   & 5.10    & 5.45     & 3.75    & -0.17  & 1.08    & 3.47    & 1.74   & 6.60 & 2.78       & 2.81    & 7.08           \\
                      &                                       & German   & 4.65    & 5.65     & 1.74    & 0.48   & 2.45    & 2.32    & 3.28   & 4.68 & 2.65       & 3.84    & 5.60           \\
                      &                                       & Persian  & 5.39    & 6.07     & 2.87    & -0.37  & 2.63    & 2.03    & 1.87   & 5.93 & 3.73       & 4.97    & 8.16           \\
                      &                                       & Russian  & 3.63    & 4.77     & 2.23    & -0.30  & 1.10    & 2.80    & 0.80   & 3.77 & 1.80       & 3.90    & 5.07           \\ \midrule
                      & \multirow{8}{*}{$\mathcal{L}_\text{id}$}   & English  & 5.53    & 4.50     & 3.33    & 1.67   & 2.57    & 0.80    & 0.00   & 2.10 & 0.07       & 3.87    & 5.47           \\
                      &                                       & Mandarin & 4.83    & 4.03     & 3.50    & 1.07   & 2.33    & 2.33    & 0.30   & 0.33 & 1.70       & 2.83    & 4.27           \\
                      &                                       & Amharic  & 3.07    & 4.23     & 3.53    & 0.03   & 1.00    & 1.60    & 0.00   & 2.57 & -1.57      & 4.80    & 3.10           \\
                      &                                       & Arabic   & 4.23    & 3.83     & 2.23    & 0.57   & 1.43    & 1.03    & 2.20   & 0.60 & -0.80      & 4.27    & 3.30           \\
                      &                                       & Bengali  & 3.80    & 3.43     & 2.97    & -0.30  & 0.43    & 0.37    & -0.43  & 1.97 & -0.93      & 5.57    & 3.07           \\
Qwen2.5-              &                                       & Italian  & 2.67    & 1.70     & 4.47    & 0.33   & 1.77    & 0.93    & -0.30  & 0.80 & 0.10       & 2.83    & 1.67           \\
Omni-7B               &                                       & Polish   & 3.07    & 3.73     & 4.30    & -1.07  & -0.77   & 1.23    & -0.30  & 0.33 & -0.67      & 3.13    & 2.13           \\
                      &                                       & Urdu     & 4.33    & 4.00     & 2.47    & -0.20  & 1.50    & 0.77    & 0.47   & 2.93 & 0.33       & 3.67    & 4.10           \\ \cmidrule(l){2-14} 
                      & \multirow{4}{*}{$\mathcal{L}_\text{held}$} & French   & 3.75    & 2.64     & 3.51    & 0.31   & -1.67   & 2.12    & 0.28   & 2.26 & -1.22      & 2.71    & 2.19           \\
                      &                                       & German   & 2.62    & 4.40     & 3.89    & 0.73   & -1.11   & 2.04    & -1.98  & 3.05 & 1.46       & 2.61    & 4.04           \\
                      &                                       & Persian  & 2.33    & 4.68     & 1.87    & -0.03  & 2.25    & 1.72    & 1.02   & 1.87 & -0.20      & 5.02    & 3.09           \\
                      &                                       & Russian  & 3.87    & 4.17     & 2.77    & -0.40  & 1.47    & 1.77    & 0.63   & 0.93 & 1.13       & 3.97    & 5.27           \\ \bottomrule
\end{tabular}%
}

\caption{Per-language steering effects ($\alpha{=}0.5$). More positive values indicate stronger causal sufficiency of amplified neurons.
}
\label{tab:steer_results_full}
\end{table*}

\end{document}